%% file: main.tex
\documentclass[11pt, a4paper, logo, onecolumn, copyright]{googledeepmind} %

\usepackage{kantlipsum, lipsum}
\usepackage{dsfont}
\usepackage{xcolor}
\usepackage{setspace}
\usepackage[toc]{multitoc}

\usepackage{caption}
\usepackage{subcaption}

\usepackage{multirow}

\usepackage[sort&compress]{natbib}%

\correspondingauthor{mwulfmeier@google.com}

\author[]{Markus Wulfmeier}
\author[]{Arunkumar Byravan}
\author[]{Sarah Bechtle}
\author[]{Karol Hausman}
\author[]{Nicolas Heess}

\affil[]{Google DeepMind}
\title{Foundations for Transfer in Reinforcement Learning: A Taxonomy of Knowledge Modalities}

\input{abstract}

\begin{document}
\maketitle

\input{main_paper}

\section*{Acknowledgements}
\input{acks}

\bibliographystyle{plainnat}
\setlength{\bibsep}{4pt} 
\setlength{\bibhang}{0pt}
\bibliography{main}

\end{document}

%% file: abstract.tex
\begin{abstract}

Contemporary artificial intelligence systems exhibit rapidly growing abilities accompanied by the growth of required resources, expansive datasets and corresponding investments into computing infrastructure.
Although earlier successes predominantly focused on constrained settings, recent strides in fundamental research and applications aspire to create increasingly general systems.
This evolving landscape presents a dual panorama of opportunities and challenges in refining the generalisation and transfer of knowledge - the extraction from existing sources and adaptation as a comprehensive foundation for tackling new problems.
Within the domain of reinforcement learning (RL), the representation of knowledge manifests through various modalities, including dynamics and reward models, value functions, policies, and the original data. 
This taxonomy systematically targets these modalities and frames its discussion based on their inherent properties and alignment with different objectives and mechanisms for transfer. 
Where possible, we aim to provide coarse guidance delineating approaches which address requirements such as limiting environment interactions, maximising computational efficiency, and enhancing generalisation across varying axes of change.
Finally, we analyse reasons contributing to the prevalence or scarcity of specific forms of transfer, the inherent potential behind pushing these frontiers, and underscore the significance of transitioning from designed to learned transfer.

\end{abstract}

%% file: main_paper.tex
{
\setlength \parskip {0pt}
  \linespread{0.95}
  \small
  \hypersetup{linkcolor=black}
  \vspace{5mm}
  \rule{\textwidth}{0.5pt}
  \vspace{-5mm}
  \tableofcontents
}

\section{Introduction} \label{sec:intro}

\textit{'Ex nihilo nihil fit'}\footnote{Parmenides, in Aristotle's Physica - Latin version.
Its origin lies in philosophical discussions in ancient Greek cosmology as to the questions surrounding a world emerging out of nothing.} 
- Nothing comes from nothing. 
While this principle engenders ongoing discourse in philosophy \citep{chapman2015complexity}, in machine learning we have to take a pragmatic stance. Performance and accuracy of our models do not materialize spontaneously but rather rely on different sources of information to effectively model a system.
These sources of information are commonly either labor-intensive manual design, or vast quantities of data.
Given the nontrivial costs associated with both endeavors, the ability to reuse or transfer previously obtained, relevant information assumes crucial significance for efficient and effective learning.

The reuse and transfer of previously acquired knowledge is prevalent in natural learning in humans and other animals \citep{spelke1992origins}. 
Nature provides us with ample examples of rapid adaptation to variations in environments, tasks, objects, and even alterations to the embodiment induced by factors such as growth or injury \citep{Wolpert2001-or}. 
This process of transfer is not confined to an individual's lifetime but extends to an evolutionary scale, facilitated by diverse mechanisms, including the transmission of genetic material \citep{Fox1996-po}. Indeed, evolution itself has engendered multiple mechanisms for transfer and reuse of information \citep{michod2000darwinian}, underscoring the inherent relevance of transfer for living beings.

For artificial learning systems, recent triumphs in supervised~\citep{he2019rethinking,huh2016makes} or self-supervised learning~\citep{devlin2018bert,brown2020language,grill2020bootstrap} result from the transfer of large pre-trained models, often referred to as foundation models (FMs)~\citep{bommasani2021opportunities}, which can be adapted to new tasks requiring only relatively small amounts of additional data.
In reinforcement learning, where the agent actively shapes its own training data, the impact of transfer compounds. It affects two stages of the optimisation process: The initial model, which can be rendered closer to optimality, and the data distribution, which builds the foundation for all subsequent optimisation.
The coalescence of both paths amplifies the importance of effective transfer, especially in active learning settings like reinforcement learning, as opposed to passive scenarios where the dataset remains fixed.

Reinforcement learning (RL) provides a formalism and framework describing the emergence of complex, goal-oriented behaviours through interaction, offering a paradigm more akin to natural learning than the conventional supervised learning from fixed datasets~\citep{sutton2018reinforcement}.
Beyond this conceptual alignment, RL's practical merit, in contrast to supervised learning, lies in its expansive access to data. 
Less constrained by our ability to assemble suitable datasets, the system determines itself which data is most relevant. 
Improving this data collection process and define apt optimization criteria includes many challenges and represents an avenue for ongoing research, with the potential for scalability surpassing human-curated datasets in the long term.
In nature, this process of acquiring information is continuous - everything is built on top of previously acquired knowledge, creating a perpetually evolving set of abilities. Conversely, in many scenarios in which RL is currently being explored, knowledge is not carried over between learning problems; each solution is learned anew, in isolation and from scratch. 
Transfer learning has the potential to disrupt this pattern and offers a trajectory where once-acquired knowledge is repurposed and adapted. 
It not only holds promise for enhancing results but also expedites experimental iteration, thereby accelerating research cycles.
The underlying insights that knowledge can generalise or be adapted more effectively transcends machine learning \citep{taylor2009transfer}, finding a place in diverse fields such as psychology \citep{skinner1953some,woodworth1901influence} and biology \citep{Wolpert2001-or, michod2000darwinian}. 

Much like the diverse mechanisms employed by natural systems for the storage and adaptation of knowledge, our current reinforcement learning agents exhibit a spectrum of representations for knowledge, encompassing policies, value functions, dynamics models, and reward models — all derived from data as the original source of knowledge.
An agent's ability to transfer or generalise is inseparably connected to the manner in which knowledge is encapsulated and stored. 
This taxonomy structures its discourse around these \textit{knowledge modalities} (policies, value functions, dynamics models, reward models, and data), their properties and transfer mechanisms.
Leveraging these considerations, we classify and analyse various forms of transfer in RL.
Consider for intuition the impact of changes in tasks on different modalities, dynamics remain consistent while optimal policies change. Similarly, computational requirements differ strongly for learning and generating behaviour between the two modalities.
While a predominant focus within the field centers on the transfer of explicit representations of behavior, such as policies or state-action value functions \citep{kirk2021survey}, it is important to underscore how the remaining modalities can facilitate transfer, a more prevalent perspective before the recent surge of successes in deep learning techniques \citep{taylor2009transfer}. 

This taxonomy focuses on transfer from a sequential perspective, progressing from an existing knowledge source, whether derived from one or multiple RL experiments or fixed datasets, to improve an RL agent's abilities during subsequent experiments. 
We can divide this process into two phases: an initial preparation phase for extracting and preparing knowledge modalities, which is followed by an application phase wherein we deploy and potentially adapt the prepared modalities in a target domain. 
In the application phase, we will consider both zero-shot generalisation \citep{kirk2021survey} (the direction application of a trained model) as well as more involved processes that require further adaptation or optimisation.
Although framing our discussion in terms of sequential transfer, we further consider approaches which have been developed in alternative settings (e.g. multi-task and meta learning) but can be deployed for transfer learning. 
We highlight these connections where appropriate, offering a comprehensive exploration of transfer methodologies that transcend the application in a sequential setting.

Transfer plays several roles in the long-term development of RL research.
Practically, it serves as a catalyst enabling the solution of harder tasks. It creates a strong conceptual connection to natural learning and finally it constitutes an elementary step to create lifelong learning systems.
Going forward, it is necessary to reflect on previous 'bitter lessons' of research in artificial intelligence \citep{bitter}.
Historically, a predominant focus has been on transfer from smaller sets of domains or tasks \citep{taylor2009transfer, zhu2020transfer}, where the intricacies of the relationship between original sources and transfer targets assume a disproportionately influential role. Transfer methods have often been tailored to optimize performance for specific relationships, posing limitations on flexibility.
To enable increasingly flexible transfer, it is imperative to reduce the amount of manual design and pivot towards learning how to adapt from immense source training distributions, akin to prevalent practices in supervised learning and the domain of large pretrained models.
Following the perspective of sequential decision making and RL in this context enables us to scrutinize foundation models in relation to the knowledge modalities, as dynamics or rewards models, value functions or policies.

Our goal is to provide a comprehensive overview of existing research and a systematic categorisation of methods in terms of benefits and trade-offs such as overall performance, data-efficiency, and suitability for specific changes between source and target MDP. 
In particular, computational cost and ease of implementation have considerable impact on the usefulness of methods for transfer.
Although quantifying these properties poses challenges, our commitment is to engage in their qualitative discussion, offering useful insights for informed design choices.
This structure allows us to survey the field and steers the discourse towards untapped research opportunities, guiding the discussion towards the potential for improvements in transfer.
This taxonomy starts with a discussion of the reinforcement learning setting in Section \ref{sec:modalities_behavior}, delving into the roles and advantages of different knowledge modalities within this context.
We proceed with a definition of transfer, an examination of different metrics, and an exploration of related fields in Section \ref{sec:transfer}.
Section \ref{sec:transfer_prep_adapt} presents a high-level perspective on diverse transfer methods in a modality-agnostic manner, which is extended to transfer via the specific knowledge modalities in Section \ref{sec:modalities_transfer}. 
We finally conclude in Section \ref{sec:discussion} with a discussion of the interplay between transfer and various modalities. Here, we illuminate open avenues for research and opportunities as well as current and future challenges for transfer in RL at a time when our ambitions for transfer, datasets, and models grow exponentially.
While we aim to provide a broad coverage of the entire field and adjacent areas, the fast-paced nature of this domain inevitably results in some omissions of relevant work. 
Nonetheless, we trust that our work serves as a valuable guide, providing a comprehensive overview and insightful direction for future research endeavors for those striving to enhance generalization and transfer capabilities in reinforcement learning.

\section{Knowledge Modalities in Reinforcement Learning} \label{sec:modalities_behavior}
\subsection{Reinforcement Learning Formalism}

\label{sec:rl_formalism}
In reinforcement learning, an agent observes its environment, takes an action, and receives a reward after the state of the environment changes.
This can be formalised as a Markov Decision Process (MDP) consisting of the state space $\mathcal{S}$, the action space $\mathcal{A}$, and the transition probability $p(s_{t+1}|s_t,a_t)$ of reaching state $s_{t+1}$ from state $s_t$ when executing action $a_t$. 
The agent's behaviour is given in terms of a policy $\pi(a_t | s_t)$.  Often, not all aspects of the environment state can be observed. In the resulting case of a Partially Observable MDP, the agent only has access to observations $o_t \in \mathcal{O}$ that are partial mappings of the state $o_t=f(s_t)$. In line with the majority of recent work, we simplify the notation and denote agent inputs as $s_t$. We only briefly work with the extended notation in later sections when we explicitly discuss the mapping between state and observations.

The agent aims to maximise the sum of discounted future rewards, commonly referred to as returns. This is captured by the objective in Eq. \ref{eq:rl} below
\begin{equation}\label{eq:rl}
    J(\pi) = \mathbb{E}_{[\rho\left(s_0\right),p(s_{t+1}|s_t,a_t),\pi(a_t|s_t)]_{t=0...T}}\Big[\sum_{t=0}^{\infty} \gamma^{t} r_t \Big],
\end{equation}
where $\gamma$ is the discount factor, $r_t=r\left(s_{t}, a_{t}\right)$ is the reward and $\rho\left(s_0\right)$ is the initial state distribution.

For an agent to successfully act in its environment, we need to determine the agent's optimal behaviour $\pi^*(a_t | s_t)$ which maximises Equation \ref{eq:rl}.
In order to generate optimal behaviour, the agent needs to implicitly understand both the dynamics of a system as well as the task itself, as described by the reward function or additional observations. The general assumption in RL is that both the true environment dynamics and the true reward are not known in advance; instead they can only be estimated via interaction with the environment.

\begin{figure*}[t]
\centering
\includegraphics[width=0.7\textwidth]{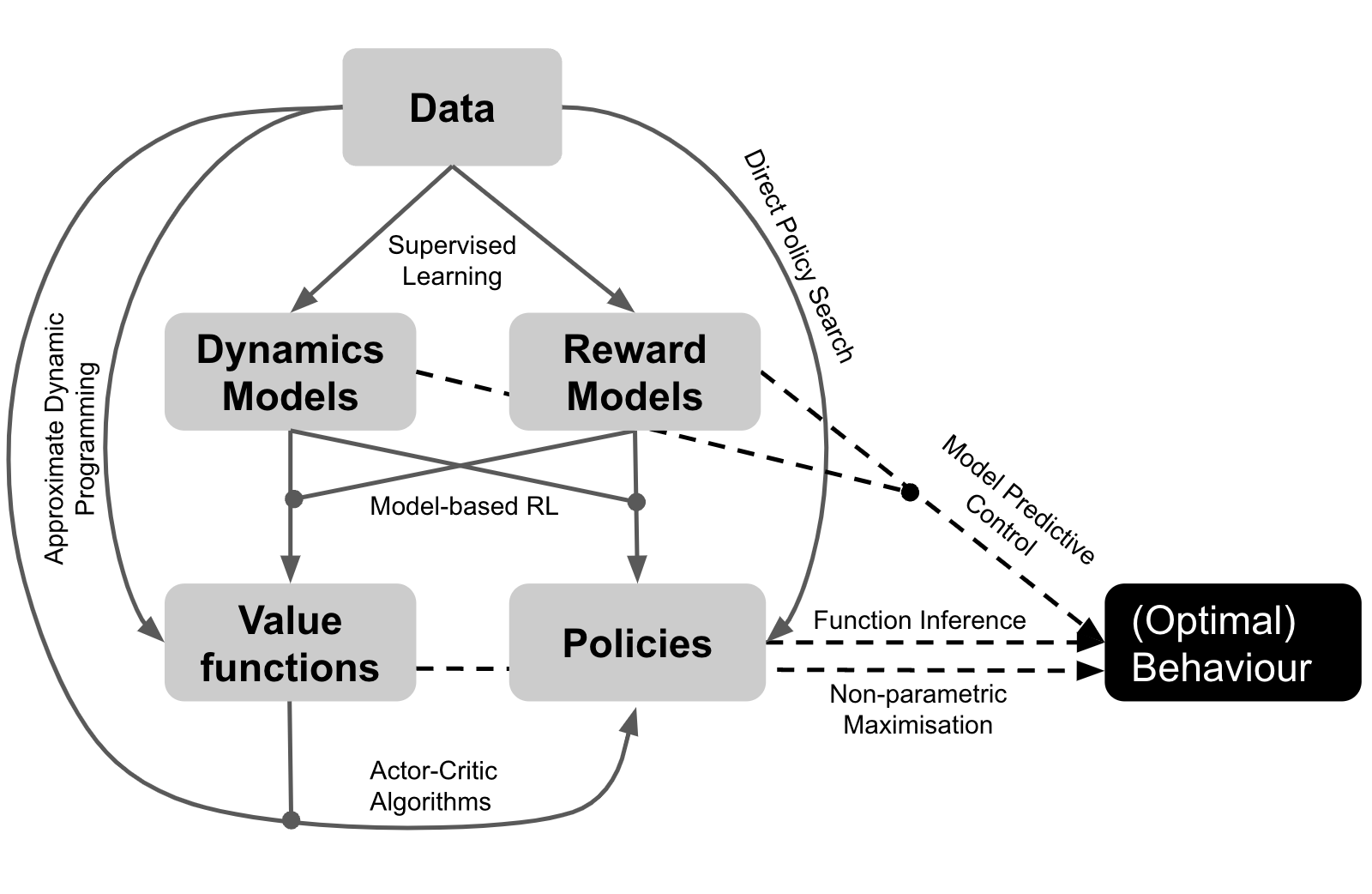}
\caption{The different knowledge modalities and most common connections among them. %
Optimal behaviour can be derived either from a single or multiple modalities. The detailed description of these connections is provided in Section \ref{sec:modalities_behavior}.}
\label{fig:modalities}
\end{figure*}

\subsection{Model-based \& Model-free Reinforcement Learning}

Optimal behaviour can be generated in different ways. In general, methods can be split into \textit{model-free} and \textit{model-based} approaches, respectively directly optimising behaviour or intermediately estimating explicit modalities for reward and dynamics.  

Model-free approaches directly optimise the agent's policy or the state-action value function. Often even direct policy learning is supported via estimating a state-action value function (also described as critic) or state value function via dynamic programming or Monte Carlo estimation. 
Alternatively, model-based methods fit a dynamics model to estimate the transition function and can additionally learn a reward model or use a defined ground truth reward. In a second step, they either apply model predictive control or learn the policy or value function by interacting with the learnt models. 
All approaches rely on data to extract these functions, either from interaction with an environment or an existing dataset.

\subsection{Knowledge Modalities}

We can estimate one or several of the following functions from existing data to represent specific types of knowledge.

\begin{itemize}
    \item \textbf{Policy} $\pi(a_t | s_t)$ - a mapping from states to actions directly describing the agent's behaviour.
    \item \textbf{State/State-Action Value Function} $V_\pi(s_t) / Q_\pi(s_t, a_t)$ - a function that maps from states (or state-action pairs) to the expected future return when acting under a particular policy $\pi$.
    \item \textbf{Dynamics Model} of the environment $p(s_{t+1}|s_t,a_t)$ - a function that predicts the next state given the current state and action.
    \item \textbf{Reward Model} $r(s_t, a_t)$  - a function predicting the reward provided by the environment for a specific state-action pair.
\end{itemize}

We will refer to these functions as ``Knowledge Modalities'' (KMs). Each of them characterises a specific aspect of the information extracted by RL agents from the MDP (such as transition dynamics, reward or (optimal) behaviour). 
Data, commonly the agent's experience, constitutes the underlying source of raw information which can be distilled into each of the modalities discussed above. By representing unprocessed information rather than a specific mapping, it provides an important mechanism for transfer and is therefore included as another knowledge modality. 
Each of these modalities provides a different angle to the transfer of information, has different properties with respect to generalisation and adaptation, and will be suited under different circumstances. 
Therefore, we structure this taxonomy of transfer learning through the lens of these knowledge modalities.
Figure \ref{fig:modalities} visualises the connections between the different knowledge modalities together with common mechanisms for conversion across them. 
We will discuss these high-level ideas in further detail in Sections \ref{sec:modalities_transfer} and \ref{sec:discussion}.

Going forward, we will denote the transition dataset used for learning
with $D$.
From transitions in $D$ it is possible to directly estimate models for
\begin{equation}
\begin{aligned}
    \{s_t,a_t\} \xrightarrow[]{} \pi_b(a_t|s_t) \\
    \{s_t,a_t,r_t\} \xrightarrow[]{} r(s_t,a_t) \\
    \{s_t,a_t,s_{t+1}\} \xrightarrow[]{} p(s_{t+1}|s_t,a_t).
\end{aligned}
\end{equation}
We use $\pi_b$ to denote the behavioural policy that generated the data and which, in general, does not have to be optimal for any task. All of these knowledge modalities are direct in the sense that it is possible to observe and learn directly from a subset of the data. 

In contrast, the state and state-action value functions (or critic), respectively $V^\pi$ and $Q^\pi$ (for any policy $\pi$), and the policy $\pi$ itself require further expensive data processing. 
There are exceptions such as when the data stems from an optimal expert and the behaviour policy is already the optimal, but in general these modalities have higher computational costs due to the aggregation of information across transitions.
The optimal policy $\pi^*$ maximises the expected future return described in Equation \ref{eq:rl} and the optimal value functions $V^{*}$ and $Q^{*}$ are simply the value functions for the optimal policy (i.e. $V^{*}=V^{\pi^*}, Q^{*}=Q^{\pi^*}$). 
In general, we can write the equations for the value functions as follows with the Monte Carlo estimates (determined over complete trajectories) in Equations \ref{eq:q_mc} and \ref{eq:v_mc} and the Temporal Difference estimates (determined over individual transitions) in Equations \ref{eq:q_td} and \ref{eq:v_td}: 
\begin{align}
        Q^{\pi}(s_t,a_t) &= r(s_t,a_t) + 
        \mathbb{E}_{p(s_{t+1}|s_t,a_t),\pi(a_{t+1}|s_{t+1})}\big[\sum_{t'=t+1}^T[\gamma^{t'-t} r(s_{t'},a_{t'})]\big]\label{eq:q_mc}\\
        &= r(s_t,a_t) + 
        \gamma \mathbb{E}_{p(s_{t+1}|s_t,a_t),\pi(a_{t+1}|s_{t+1})}\big[Q^{\pi}(s_{t+1},a_{t+1})\big]\label{eq:q_td}
\end{align}
and 
\begin{align}
    V^{\pi}(s_t) &=\mathbb{E}_{p(s_{t+1}|s_t,a_t),\pi(a_t|s_t)}\big[\sum_{t'=t}^{T}[\gamma^{t'-t} r(s_{t'},a_{t'})]\big]\label{eq:v_mc}\\
    &=\mathbb{E}_{p(s_{t+1}|s_t,a_t),\pi(a_t|s_t)}\big[r(s_t,a_t) +  \gamma V^{\pi}(s_{t+1})\big]\label{eq:v_td}
\end{align}

In the rest of this section, we will discuss the properties of these knowledge modalities and approaches to generating (optimal) behaviour from each of them. 
We will build upon these concepts in later sections to cover the transfer of different modalities.

\subsubsection{Data}
\label{sec:modalities_data}
Data includes the raw information from which the other modalities are distilled. %
It 
provides information about the transition dynamics, the agent's objective via the rewards, and the underlying behaviour policy which generated the dataset. Note, the latter is often not equal to the optimal policy.
Data can be used to learn a value function or policy using model-free offline RL algorithms. 
It further can be used to first estimate a dynamics model and reward function to perform model-predictive control or model-based RL.

Even incomplete data can be useful. When any aspect is missing, information about the remaining parts can be extracted. Imagine data with missing or sparse information about rewards. Here, we can still fit a model for the dynamics or the behaviour policy \citep{pomerleau1988alvinn}.
Given sufficient assumptions, we can further indirectly extract information about missing components. For example, if the rewards are unavailable but we assume the data to be expert trajectories, we can apply inverse reinforcement learning to train a reward function \citep{ng2000inverse}.
While first person experience in the form of states and actions is the most versatile form of data, state-only sequences or sequences of third-person observations include relevant information but may be more restricted in the inferences they allow and the processing that is needed to generate (optimal) behaviour. 
In summary, data enables all further modalities. From it, we can extract parametric functions either online during interaction with an environment or offline via supervised learning or offline RL.
However, data is limited as in order to generate behaviour, we generally need to first extract a function for any of the other modalities. %

\subsubsection{Dynamics Models}
\label{sec:modalities_dynamicsmodels}

Dynamics models capture the underlying transition dynamics of the agent embodiment and its environment. Model learning is usually formulated as (self-)supervised learning (estimating the conditional transition density $p(s_{t+1} | s_t,a_t)$ or deterministic transition function $s_{t+1}=f(s_t,a_t)$) \citep{sutton1991dyna}. %

The predominant way to use dynamics models to generate behaviour is via model-based planning within a model-predictive control loop \citep{garcia1989model, morari1999model, mayne2014model}. To generate a single action, the dynamics model has to be evaluated multiple times, which can be both computationally more expensive as well as accumulate errors across steps. Other ways to leverage models directly within the policy optimisation process can be broadly categorised as distilling knowledge from the model into a policy or value function. Examples include using the model to generate imagined transitions for model-free RL or value gradients for policy training. 
These approaches commonly require access to a reward function -- either learned or pre-specified.

\subsubsection{Reward Models}
\label{sec:modalities_rewardmodels}
Models of the reward may be learned via supervised learning of the probabilistic ($p(r_t | s_t, a_t)$) or the deterministic model $r_t = g(s_t, a_t)$. %
In some cases reward models can also be inferred from reward-free trajectories if we assume they are generated by task experts, for example, via inverse reinforcement learning~\citep{ng2000inverse}. %

Given access to both the dynamics and reward models, we can generate optimal behaviour as previously discussed. If only the reward model is available, information about the dynamics has to be additionally provided. For example, we can apply model-free offline RL with a dataset containing only state transitions in combination with a learned reward model. Generally, behaviour generation with reward models relates to dynamics models in terms of computational cost as well as accumulation of errors.

\subsubsection{Value Functions}
\label{sec:modalities_valuefunctions}
Unlike dynamics models or rewards which directly represent per-transition properties of the underlying environment or task, value functions make aggregated predictions about the future. 
In particular, a state-action value function or critic $Q^\pi(s,a)$ predicts the return obtained with a particular policy $\pi$, as a function of the current state-action pair (while state value functions $V^\pi(s)$ condition on the state alone). Consequently, it entangles information about the system dynamics, agent behaviour and objective function. Value functions can be optimised either via Monte Carlo (MC) methods \citep{michie1968boxes} or Temporal Difference (TD) learning \citep{sutton1990time}, the latter enables partial verification of the value function from individual transitions via the TD error.

Behaviour can be generated from state-action value functions either by performing non-parametric optimisation over sampled actions or by amortising this step by sampling actions from a parametric policy that is trained in parallel (e.g. in the actor-critic setting). 
Pure state value functions fulfil a different role in RL and are often constrained to actor-critic algorithms to accelerate or stabilise learning a policy. 
Several sub-classes of value functions provide additional flexibility and preferable properties for transfer and generalisation which are covered in more detail in Section \ref{sec:modalities_transfer}.

\subsubsection{Policies}
\label{sec:modalities_policies}

Policies represent behaviour as deterministic $a=\pi(s)$ or stochastic mappings $\pi(a|s)$ from states to actions and are directly trained to maximise the future returns.
This optimisation can be done either directly from perceived rewards \citep{williams1992simple} or via the intermediate step of learned state-action value functions \citep{WITTEN1977286} or dynamics models \citep{sutton1991dyna}.

Generating behaviour with policies is low-cost compared to other approaches such as on-the-fly optimization with MPC, as function outputs have to be computed only once. 
Sub-optimal policies can be further used as starting point for fast refinement by guiding exploration as well as reducing the learning problem. 
For example, we can use stochastic policies as a proposal distributions in model-based RL \citep{wang2019exploring, SilverAlphaGo} or prior to perform additional re-weighting or filtering (e.g. using a Q-function) for action selection in the model-free case \citep{galashov2020importance, barreto2018transfer}.

\subsection{The Role of Representations}
\label{sec:modalities_representations}
The transfer of representations has an important role across the research landscape to simplify and improve the training of policies, value, or other functions. %
In practice, representations are commonly integrated within RL agents through the use in one or several of the discussed modalities. For this reason, we are discussing relevant work directly within the context of each modality.
There are several useful resources for a separate discussion and explicit focus on different forms of representation learning \citep{schoel2019, bengio2013representation, finlayson2020learning}.

\section{Transfer Learning Background}%
\label{sec:transfer}
This section discusses general goals and metrics for transfer in RL and connects to related fields. 
We begin by discussing a broad definition for transfer to capture key aspects and structure the taxonomy, and create bridges to prior work \citep{taylor2009transfer,Lazaric2012-mv}, 

\subsection{Transfer Definition}
We formalise transfer in reinforcement learning as a process where knowledge extracted in a source MDP $\mathcal{M}_S$ (or a set or distribution thereof) - either directly via interaction or via an intermediate dataset - is used to improve performance and accelerate learning in a target MDPs $\mathcal{M}_T$ (or a set or distribution thereof). The objective function of transfer in RL can be intuitively expressed as: 
\begin{equation*}\label{eq:transfer}
    J^T(\pi_T,\mathbf{KM}_S) = \mathbf{E}_{\rho_T(\mathbf{s_0}), p_T(\mathbf{s'}|\mathbf{s},\mathbf{a}), \pi_T (\mathbf{a}|\mathbf{s}, \mathbf{KM}_S)} \sum_{t=0}^\infty \gamma^{t} r_T(\mathbf{s_{t}}, \mathbf{a_{t}}),
\end{equation*}
where subscript $S$ and $T$ correspond to the source and target MDPs respectively and $\mathbf{KM}_S$ corresponds to the knowledge modality being transferred from the source domain.
While we use Equation \ref{eq:transfer} to describe the \textit{single source MDP single target MDP} setting for simplicity, a distribution over MDPs is possible for both aspects.
Generally, this transfer setting assumes that the agent cannot collect further information from the source MDP  $\mathcal{M}_S$ once transferred and interacting with the target MDP $\mathcal{M}_T$. It can however take advantage of the transferred knowledge modalities.
If there is no additional training in $\mathcal{M}_T$, we evaluate generalisation or 0-shot performance and, if we continue training, we can evaluate adaptation performance via the metrics discussed in Section \ref{sec:transfer_metrics}.

We are using the term $\pi_T (\mathbf{a}|\mathbf{s}, \mathbf{KM}_S)$ not to denote an explicitly represented policy but from a more general perspective. Instead of inference of an explicit policy mapping from states to actions, it describes any optimisation mechanisms which generates this mapping. This includes the simple execution of a function for the policy, but also planning with a model and even learning a new mapping using information from $\mathbf{KM}_S$. It serves as common representation for all paths of transferring knowledge modalities discussed later in Sections \ref{sec:modalities_transfer}.

\subsection{Transfer Targets} %
We previously defined transfer in RL as a process that allows the agent to improve performance in the target MDP $\mathcal{M}_T$ based on prior knowledge extracted from a source MDP $\mathcal{M}_S$ via interaction or corresponding dataset. Different changes of the MDP can cause the need for transfer and we will refer to them as \emph{transfer targets}. 

Common targets are given by new tasks (described by new reward functions \citep{teh2017distral, kalashnikov2021mt}), system dynamics (e.g. when transferring from simulation to real hardware \citep{nagabandi2018learning, yu2017preparing}, changes in the mapping from states to observations (e.g. lighting or other visual changes \citep{bousmalis2018using, Wulfmeier2017AddressingAC}), or changes in the initial state distribution (e.g. in autonomous or reset-free RL~\citep{zhu2020ingredients, gupta2021reset}). 
One step further we can also consider changes in the state or action space (e.g. when transferring from general, non-robotic data for applications in robotics~\citep{chen2021learning, smith2019avid} or between different robotic embodiments~\citep{truong2021learning,zakka2022xirl}).
Overall, different transfer targets often favor different methods which we will further discuss in Section \ref{sec:modalities_transfer}. %

\subsection{Transfer Metrics}\label{sec:transfer_metrics}
While the principal objective of transfer learning is captured by the performance of the agent in the target MDP $\mathcal{M}_T$, we can use additional criteria to disentangle different facets of the impact of transfer. 
Extending further than pure unsupervised or supervised learning with existing, pre-defined datasets, the problem of transfer is particularly important in reinforcement learning. In addition to affecting the initial transferred function and therefore the starting point of the optimisation, transfer can affect the future training data distribution by guiding exploration. 
In order to cover these facets, there are multiple key metrics to describe the effectiveness of transfer in RL:

\begin{enumerate}
    \item initial performance of the source agent in the target MDP $\mathcal{M}_T$,
    \item amount of target data required to reach a certain performance in $\mathcal{M}_T$,
    \item asymptotic performance of the agent additionally trained on data from $\mathcal{M}_T$%
\end{enumerate}

\begin{figure}[t]
\centering
\includegraphics[width=0.5\textwidth]{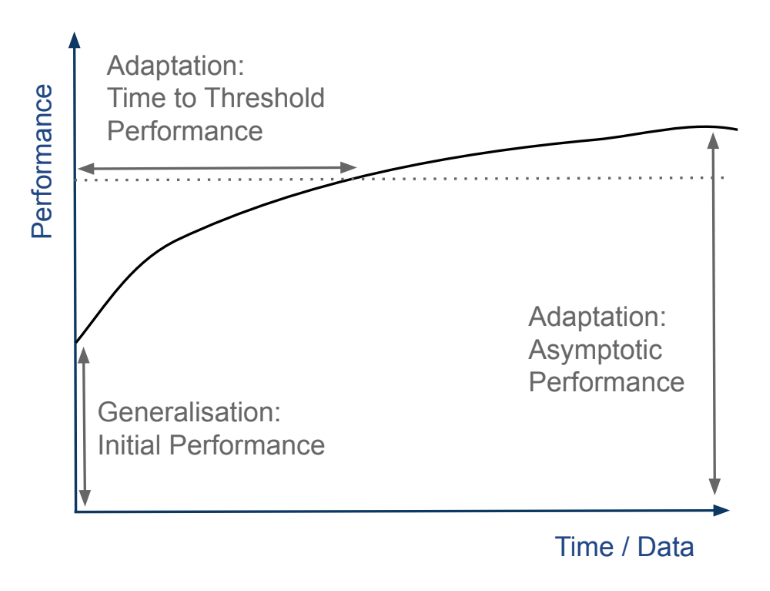}
\caption{The principal performance metrics for transfer in RL. Without further training we can measure pure generalisation via an agent's initial performance. Adaptation can be measured both via time to a specified threshold performance or the agent's asymptotic performance after convergence.}
\label{fig:transfer_metrics}
\end{figure}

The first metric displayed in Fig. \ref{fig:transfer_metrics} describes 0-shot performance or generalisation of a transferred agent without further training in the target domain.
The other two provide different perspectives on the effectiveness and efficiency of adaptation or how well an agent performs after additional training
\footnote{See \citet{taylor2009transfer} for further perspectives on related metrics .}.

As the metrics above focus on the data efficiency and performance, it is important to investigate and understand other relevant aspects of transfer. In particular computational and memory requirements critically affect cost and time required for experimentation. If a mechanism or modality requires vast computational resources, we are limited to specific hardware and in the extreme case might not be able to generate behaviour in real time during inference \citep{Schubert2023AGD, Brohan2023RT2VM}. 
Further metrics can include interpretability, naturalness, robustness, and safety of the transferred behaviour, as well as ease of implementation and use.
While these are important aspects for any sub-field of machine learning, we will focus the discussion on the most common, limited resources with performance, data-efficiency and computational cost.

\subsection{Connections to Multi-Task and Meta RL}
The problem formulation of meta reinforcement learning~\citep{wang2016learning, duan2016rl2, finn2017model,Schmidhuber1994OnLH} shares many commonalities with the transfer learning setting.
Indeed, meta learning can be considered as a special case of transfer learning: given access to a distribution of source MDPs, meta learning directly optimises an agent's ability to transfer efficiently (e.g. with regards to variants of the criteria listed above).
Transfer learning, in general, is less prescriptive and does not necessarily explicitly optimise for fast adaptation. %

A narrow view of multi-task RL, in contrast to sequential transfer, does not inherently distinguish between training and transfer. %
Multi-task RL assumes that all the tasks of interest are solved simultaneously, while the term transfer learning frequently implies a sequential scenario. Multi-task learning assumes access to the source MDP $\mathcal{M}_S$ throughout training, while transfer learning assumes that $\mathcal{M}_S$ cannot be accessed during transfer, but only during the preparation phase.
For a broader overview of multi-task learning we refer to~\citep{crawshaw2020multi,caruana1997multitask}.

These distinctions, while important, are orthogonal to the main categorisation of knowledge modalities of this taxonomy and for this survey, it is highly relevant to connect to the use of different modalities in multi-task and meta RL works as examples of mechanisms for transfer.

\subsection{Connections to Continual and Lifelong Learning}
Transfer in RL is intimately connected to the field of continual or lifelong learning~\citep{parisi2019continual, thrun1998lifelong, ring1994continual}. In fact, it can be considered as an elementary building block of a system that learns continually: transfer in RL allows an agent to quickly acquire a new behaviour, but it is not explicitly concerned with
many of the challenges associated with scenarios in which agents are required to repeatedly acquire new behaviours or skills (like growing knowledge representations, or catastrophic forgetting  \citep{goodfellow2013empirical,kirkpatrick2017overcoming}).
Thus, while there exist differences, we believe that many of the challenges encountered in transfer learning scenarios are also fundamental for continual learning scenarios and advances in the former will transfer to the latter.

\begin{figure*}[t]
\centering
\includegraphics[width=0.6\textwidth]{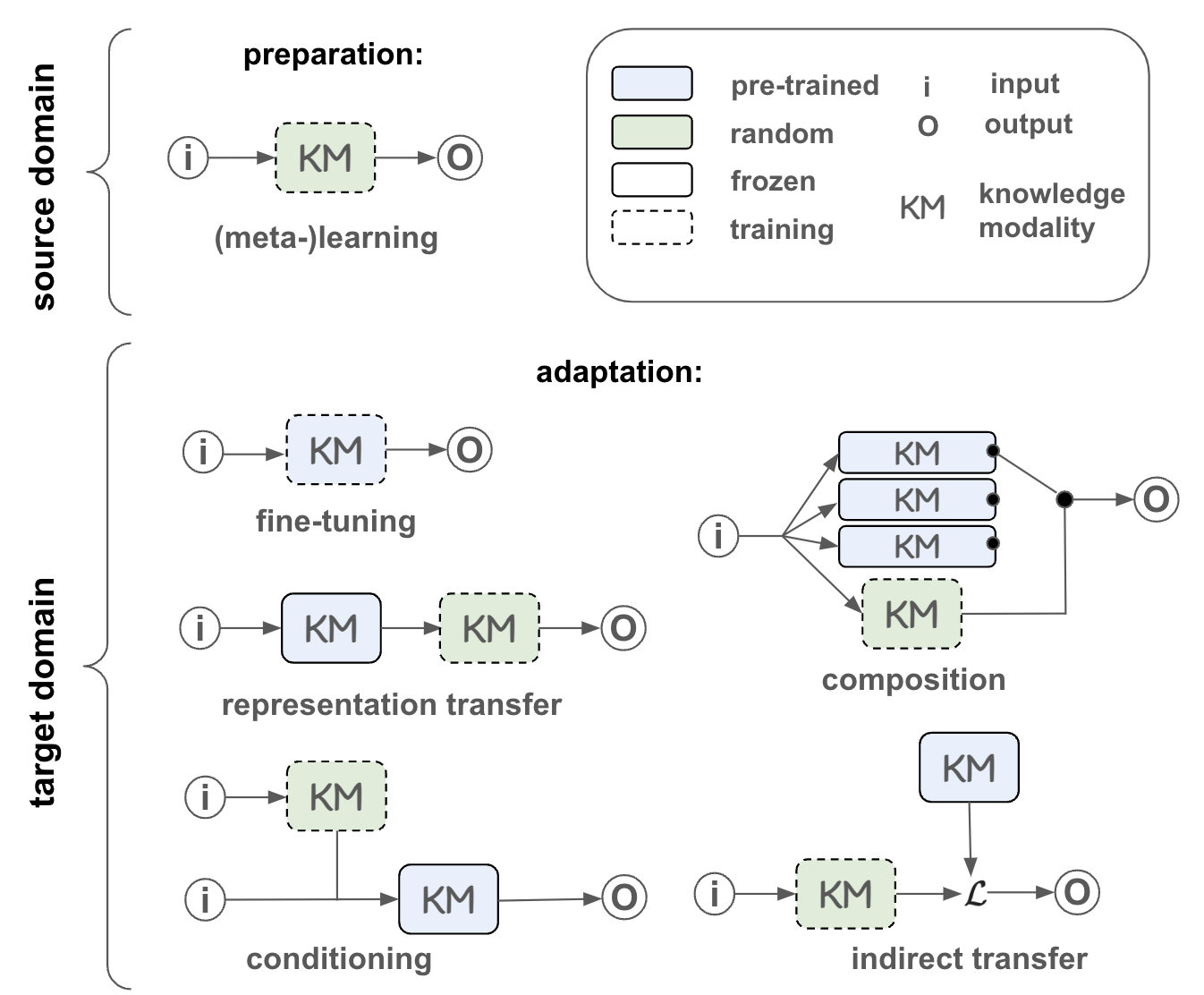}
\caption{Knowledge modalities in transfer. Top: preparation of KM involves learning from dataset or environment interaction which the potential application of meta learning. In the target domain, different transfer mechanisms exist to adapt to new tasks, we can coarsely categorise these as: fine-tuning, transfer of representations, composition, conditioning, and indirect transfer. We provide detailed descriptions regarding these categories for each KM in Section \ref{sec:modalities_transfer}.}
\label{fig:km_in_transfer}
\end{figure*}

\section{Transfer Mechanisms}
\label{sec:transfer_prep_adapt}
Following the discussion of transfer learning as a field and its connections to relevant areas, this section provides a broad overview of mechanisms relevant for transfer across all modalities. 
We can loosely split the transfer problem into two stages: preparation and application. 
Preparation denotes any agent interaction with the source MDP. Examples can include training a policy either with direct environment interaction or offline from data from the source MDP. Application refers to both initial 0-shot generalisation and further adaptation.
Figure \ref{fig:km_in_transfer} provides a schematic visualisation of different mechanisms with generic input output pairs to cover all modalities.

All modalities can be largely prepared in the same ways and we will cover the preparation in this section. %
The application, including mechanisms for adaptation, will then be discussed separately for each modality in 
Section \ref{sec:modalities_transfer}. 
We will add individual examples for explanatory purpose and expand on breadth and details in the following sections.

\subsection{Preparation}
A coarse categorisation of mechanisms for the preparation stage includes:
\begin{itemize}
    \item learning with fixed datasets (unsupervised/self-supervised, supervised learning)  
    \item reinforcement learning (with given rewards or unsupervised)
    \item meta learning (in combination with any of the above)
    \item hand-designed/engineered modalities (e.g. PID controllers)
\end{itemize}

Supervised and self-supervised/unsupervised learning enable us to extract policies, value functions, dynamics and reward models from existing datasets. 
For example, this could be training a policy encoder via input reconstruction \citep{lange2010deep, Wulfmeier2021RepresentationMI}, supervised learning of reward functions \citep{singh2019end, ebert2018visual}, or imitation learning and distillation of policies \citep{lange2010deep, ross2011reduction, tirumala2020behavior}.
With reinforcement learning, we can extract these modalities while training on related source MDPs given rewards \citep{byravan2020imagined, wulfmeier2020compositional} or when rewards are not given via agent-intrinsic objectives such as curiosity, diversity or empowerment \citep{Groth2021IsCA,Eysenbach2019DiversityIA,Gregor2017VariationalIC}.
In addition, learning to learn or meta learning \citep{thrun1998lifelong,schmidhuber1987srl} can be introduced into these steps to improve the efficiency of any later adaptation.
Finally, we can apply more classical engineering techniques to generate mappings such as a PID controller \citep{Rana2021BayesianCF} or other manually scripted behaviour \citep{torrey2007relational, Johannink2019ResidualRL} for the agent's policy. %

\subsection{Application: Generalisation and Adaptation}
When transferring the resulting modalities to the target MDP, we have multiple options for their application.
From a high-level perspective, we can identify two principal schemes. 
First, we can transfer and directly apply previously learned modalities to generate behaviour as discussed in Section \ref{sec:modalities_behavior}, and continue learning.
A common example is direct fine-tuning of a transferred policy.
Second, we can apply modalities indirectly to affect the optimisation of other modalities which we use to generate behaviour.
Staying with the example of policies, we can also regularise a new student policy towards a transferred expert.
In order to make use of transferred data, we are commonly required to indirectly integrate it into the training process for other modalities which can be applied to generate behaviour. There is however a small set of exceptions where data is a direct part of the behaviour generating mechanism \citep{Pari2021TheSE} which we will discuss in Section \ref{sec:modalities_transfer_data}. 

\paragraph{Direct Application of Modalities}
We will categorise the application of modalities along the following list of mechanisms and provide a very short description with more details and specific examples for each modality to follow in the next section. 
Many of these mechanisms are orthogonal and complementary and can be combined to improve performance or learning speed.
The main mechanisms are:

\begin{itemize}
    \item generalisation and 0-shot transfer 
    \item fine-tuning 
    \item representation transfer 
    \item hierarchy: conditioning
    \item hierarchy: composition
    \item meta learning
\end{itemize}

In many cases, pure \textbf{generalisation or 0-shot transfer} of learned functions has shown impressive results when the target MDP can be treated as close to the distribution of source MDPs \citep{kirk2021survey, finn2017deep, chapman1991input}.
A common example is the use of domain randomisation for transfer between simulation and real hardware for robotics \citep{tobin2017domain, sadeghi2016cad2rl}.
The generalisation setting focuses on the first metric from Section \ref{sec:transfer}: the initial zero-shot performance. 
Section \ref{sec:modalities_behavior} describes how the different modalities can be used to generate behaviour without further adaptation and discusses the trade-offs involved. 

However, there are many scenarios where 0-shot transfer does not suffice and different adaptation methods are required.
The simplest mechanism to consider is \textbf{fine-tuning} \citep{hinton2006reducing} which continues to train the pre-trained model. One of the key benefits of adapting via fine-tuning is its engineering simplicity. The only requirement is the reloading of a previous modality and the training algorithm can remain unchanged.

Instead of transferring complete instances of modalities and fine-tuning, \textbf{transferring representations} \citep{taylor2007representation} generally enables the combination of parts from randomly initialised and previously trained models. This mechanism enables transfer across less related domains as similarity is not required with respect to optimal behaviour but only the usefulness of representations for further optimisation \citep{Reid2022CanWH}.

The application of \textbf{hierarchical RL} provides another, modular perspective to the combination of initialised and transferred modalities. Commonly, a new learned modality can be used to \textbf{condition} the transferred one by providing inputs, e.g. by direct optimisation in an embedding space  \citep{peng2020learning}, %
or the space of conditioning functions \citep{heess2016learning, nachum2018data}.  %
textbf{Composition} provides another mechanism to reuse and adapt multiple modalities (e.g. multiple policies) in a hierarchical setting. We can learn to compose a solution from multiple objects rather than modulate the function of a single one via conditioning. 
Common examples include composition of multiple policies \citep{andreas2017modular,wulfmeier2020compositional} or value functions \citep{barreto2016successor, sutton1999between}.

Finally, \textbf{meta learning} targets explicit optimisation for fast adaptation and gnerally builds on other optimisation mechanisms which are themselves being rendered more efficient.  
General categories of meta learning include gradient-based approaches
\citep{finn2017model} or sequence modelling or in-context learning \citep{wang2016learning, duan2016rl2,openai2019solving}.
Meta learning stands out in comparison to other forms of adaptation as we cannot spontaneously decide to apply it during adaptation if it has not been used during preparation. Meta learning uses the preparation phase in order to discover how to quickly adapt.

\paragraph{Indirect Application of Modalities} 
Indirect application describes the use of modalities to guide optimisation rather than direct usage for behaviour generation. Specifically, we can differentiate two approaches:

\begin{itemize}
    \item transfer via (auxiliary) objectives
    \item transfer via data (e.g. by shaping exploration)
\end{itemize}

A common path for indirect transfer is via (auxiliary) optimisation \textbf{objectives}.
Examples include distillation of knowledge by training with regularisation towards previous policies \citep{tirumala2020behavior}, the use of data to define a reward function \citep{ho2016generative, merel2018neural, wulfmeier2015maximum}, optimisation targets for policies and Q-functions by executing planning with transferred models \citep{hafner2019dream}, or using transferred Q-functions directly as targets \citep{wang2020target}.

The other indirect way to use knowledge modalities is via \textbf{data} for the optimisation process. 
For example, we can use previous policies for data generation and learn off-policy from this experience \citep{Vezzani2022SkillSAS, Campos2021BeyondFT}. 
An agent's original experience (as data) is commonly used for indirect transfer via learning another modality. Often we directly reload previous data into the agent's replay buffer \citep{vecerik2017leveraging, nair2020accelerating} 
Special cases are represented by the direct use as a foundation for an engineered reward function \citep{peng2018deepmimic, Hasenclever2020} or a non-parametric policy \citep{Pari2021TheSE}.
While the distinction between transfer via the objective and via data is important, we will summarise both into the category of indirect transfer for all following sections for simplification.

\section{Knowledge Modalities in Transfer}\label{sec:modalities_transfer}

In this section, we will cover how different knowledge modalities have been used for transfer in RL. 
They generally represent and transfer knowledge in distinct ways, offering varying degrees of generalisation, flexibility and adaptability, and necessitating diverse computational efforts for adaptation and the generation of functional behaviors.
From data to models to policies, computational costs for inferring optimal behaviour are being reduced, albeit at the potential expense of loss of information.
The functions underlying policies, for instance, allow direct execution for determining new actions and generating trajectories, but their utility is intricately tied to both system dynamics and reward functions.

At the opposite end, data represents the raw source of all information from which dynamics models, policies, and rewards can be distilled.  However, the extraction of behavior from data demands heightened computational resources.
Beyond these fundamental constraints, we are going to explore how various forms of transfer apply across the entire spectrum, offering a comprehensive understanding shared patterns.

\subsection{Data}
\label{sec:modalities_transfer_data}
Raw data can provide unprocessed information about the environment dynamics, task, or (optimal) behaviour. As discussed in Section~\ref{sec:modalities_data}, different types of data carry these distinct types of information and require respective processing steps in order to derive optimal behaviour.
Beyond the focus on action-centric data for training with different reinforcement or imitation learning methods, the inclusion of domain-external data sources can be used for representation training. While unlocking access to expansive datasets, this approach is inherently limited in providing insights into specific aspects of the problem. Notably, leveraging large image datasets for the pre-training of early layers in deep networks exemplifies this paradigm \citep{lange2010deep, Xiao2022MaskedVP, Wulfmeier2021RepresentationMI, Dittadi2021TheRO}, as detailed in the representation transfer section for each respective modality.

With minor exceptions, domain-specific data is predominantly applied \textit{indirectly} to transfer knowledge by integration into the optimisation process, commonly either as addition to new data or via auxiliary objectives. Noteworthy exceptions involve the mechanisms for effective retrieval and processing of existing trajectory data during behavior generation, exemplified by studies such as \citep{Pari2021TheSE, Du2023BehaviorRF,goyal2022retrieval,humphreys2022large}. However, these instances constitute a minority within the spectrum of data usage.
Consequently, while the other modality sections broadly adhere to the categorization delineated in Section \ref{sec:transfer_prep_adapt}, this section adopts a more granular lens, delving into nuances surrounding the indirect transfer via data.

\paragraph{Transfer of Data or Intermediate Modalities}
The use of data can follow two general paths. We can integrate data directly into an online experiment, or first extract another knowledge modality offline which to apply in the online setting, the latter of which we will cover in the respective modalities' sections. In particular, transition data (including actions) can be used online in the context of off-policy \citep{hester2018deep,cabi2019scaling,vecerik2017leveraging,xie2020deep,tao2021repaint,Tirinzoni2018ImportanceWT, Walke2022DontSF,andrychowicz2017hindsight, ball2023efficient}, or imitation learning \citep{Ebert2022BridgeDB}. 
It can further shape the agent's behaviour via additional objectives \citep{ho2016generative,nair2018overcoming,abdolmaleki2021multi,Davchev2021WishYW}. 

The alternative application is via an (offline) preparation step, during which an intermediate knowledge modality - for instance a model of the system dynamics, a value function, or a policy - is computed (see for example \citep{levine2020offline,fujimoto2018offpolicy,nair2020accelerating,singh2020cog,ebert2018visual,song2022hybrid}). 
The resulting modality can then be transferred directly or indirectly to optimise behaviour in the target domain. Detailed cases are being discussed in the subsections of the respective modalities, while this section focuses on the transfer of data into the new agent to aid learning in the target MDP.

\paragraph{Adapting Datasets}
Data can provide information about system dynamics and tasks. If the system dynamics remain unchanged between source and target, data can be relabelled across different tasks and reward functions \citep{andrychowicz2017hindsight}.
This makes it possible to learn off-policy or offline for new objectives. This can target a number of discrete tasks \citep{riedmiller2018learning, laroche2017transfer} as well as a dense goal space \citep{Davchev2021WishYW}. It can be beneficial to either use data indiscriminately to learn about all tasks \citep{Lambert2022TheCO}, or to learn more effectively, e.g. by selecting specific states as goals that were visited as part of a trajectory \citep{andrychowicz2017hindsight}. Similarity-based data filtering methods have been further extended for image-based, high-dimensional data \citep{chebotar2021actionable,Du2023BehaviorRF}. 
Further mechanisms have been proposed for determining suited tasks for more effective data relabelling \citep{eysenbach2020rewriting,NEURIPS2020_57e5cb96,sharma2021persistent}.

When learning multiple tasks from the same data, a policy or value function requires additional conditioning inputs to distinguish between different tasks.
Different task descriptors have been used in this context with the simplest case being task IDs \citep{wulfmeier2020compositional}. 
Naturally, there exist better suited spaces for identifying and describing structure across tasks. For example, similarities between tasks should give rise to shorter distances between their descriptors. Examples are given by goal positions \citep{andrychowicz2017hindsight}, images \citep{chebotar2021actionable}, or language commands \citep{Lynch2020LanguageCI}.
Even without any task descriptors or rewards, offline RL can incorporate existing data. When these are unknown, the data still includes information about system dynamics \citep{singh2020cog}.

We can further combine various data sources such as agent embodiments or source tasks \citep{Brohan2022RT1RT, reed2022generalist, zhou2022forgetting}. However, there are challenges when integrating heterogeneous data sources. These issues range from different dynamics and task distributions which can be addressed via importance sampling or weighting techniques \citep{zhou2022forgetting,chebotar2021actionable} to varying observation and action types which require flexible agent architectures such as transformer models \citep{Vaswani2017AttentionIA, reed2022generalist, Brohan2022RT1RT}. This area is of particular interest when scaling data access for data-expensive domains and is further discussed under Section \ref{sec:discussion}.

\paragraph{Objectives for Action Matching} %
Beyond biasing the data distribution for learning, expert data may be used more directly to provide information about the optimal behaviour for a given task \citep{vecerik2017leveraging, Shiarlis2018TACOLT,ross2011reduction}. 
By optimising a policy to imitate expert actions, we can guide agents to overcome exploration challenges, such as given by sparse or noisy rewards. 
In an extreme case, when this is the only signal used to train the new policy, the approach reduces to behavioural cloning \citep{pomerleau1988alvinn}. In many cases, however, this learning signal is either combined with an RL loss \citep{nair2018overcoming}, or the data is further weighted, for instance by a reward-based signal \citep{wang2020critic,nair2020accelerating,jeong2020learning}.

\paragraph{Objectives for State Matching} %
Where expert actions are not available or not compatible with the target domain, different approaches can encourage a new policy to match the expert's state visitation distribution \citep{ ho2016generative, ziebart2008maximum, ng2000inverse}. 
While matching states instead of actions commonly comes with the cost of required environment interaction to determine the new agents trajectories, there are multiple benefits for this direction.
Behavioural cloning requires that the coverage of the transferred data is sufficient to learn robust controllers and otherwise can lead to compounding errors \citep{ross2011reduction}.
In contrast, matching the expert's state distribution often requires fewer expert trajectories at the cost of further environment interaction for the agent.
Finally, we can additionally match abstractions or projected features of transferred trajectories \citep{Torabi2019RecentAI}.

Objectives for matching states can be created by penalising an engineered distance function between pairs of student and expert states \citep{peng2018deepmimic,merel2018neural,Hasenclever2020,aytar2018playing} or via a learned distance function. 
Learned objectives can include inverse reinforcement learning \citep{ziebart2008maximum, Ratliff2006MaximumMP, Abbeel2004ApprenticeshipLV} where in particular adversarial settings have gained popularity which train a discriminator to distinguish between between agent and transferred trajectories \citep{ho2016generative,merel2017learning,wang2017robust,fu2017learning} as well as other forms of distribution matching \citep{Lee2019EfficientEV,Kim2018ImitationLV}. 
In these settings, the transferred knowledge can be either data, which is directly used in the optimisation procedure, or intermediate policies which are used to generate data. The latter is discussed in further detail in Section \ref{sec:modalities_transfer_policies}. 
Similar to matching actions, methods for matching states can also be combined with a regular reward-based learning signal \citep{zhu2018reinforcement, Peng2021AMPAM}.

Finally, we can further simplify the learning problem by providing information about suitable initial states for training the agent \citep{Popov2017DataefficientDR,merel2017learning,nair2018overcoming}. This application requires the ability to control the initial states of the agent, an ability not available in many real-world settings.

\paragraph{Benefits and Disadvantages}
Unlike the other modalities, raw data provides information about the original experience that an agent receives and is not affected by any potential loss of information or influence of the inductive bias that may result from fitting a function approximator for another modality. In that sense, data is highly useful as the uncompressed, unbiased representation of available information. 

However, this comes at the price of additional inference cost in translating the raw experience into an actionable representation (e.g. via model learning and MPC; or offline value iteration to obtain a policy). Data use can be limited by distribution shift (e.g. data may have been recorded in parts of the state space that are irrelevant for the transfer target). Finally, processing data offline without access to the underlying generating distribution (e.g.\ the (PO)MDP), can in some cases aggravate instabilities of the inference process (for instance, in the case of offline reinforcement learning \citep{levine2020offline}).

\subsection{Dynamics Models}
\label{sec:modalities_transfer_dynamics}

In general, models of the transition dynamics tend to be amenable to task transfer as they capture the dynamics of the underlying environment, albeit with a dependence on the relation between source and target data distributions.

\paragraph{Generalisation and 0-Shot} 
The most common method for transferring models is to transfer pre-trained models onto a target task, where the model can be applied in different ways. In particular, there is a long line of work on training action-conditional predictive models on datasets collected via random, pre-scripted interactions with the environment, followed by model-predictive control (MPC) on top of the frozen pre-trained model to generate actions on downstream tasks, using either a hand-specified or learned reward on the downstream task ~\citep{lenz2015deepmpc, finn2017deep, ebert2018visual, byravan2017se3, nematollahi2020hindsight}. %

Early examples cover low-dimensional state-space models %
\citep{lenz2015deepmpc,williams2017information} %
with a combination of offline model learning and MPC for online behaviour generation. MPC can further be leveraged for generating training data for model-training \citep{levine2016end, zhang2016learning}; demonstrating good generalisation to novel control tasks across manipulation, locomotion and aerial robots.

With the advent of convolutional architectures there is an influx of visual predictive models which directly model the dynamics of visual or depth observations, predominant amongst these is the "Visual Foresight" paradigm ~\citep{finn2017deep, ebert2018visual} which trains action-conditional visual predictive models in an unsupervised manner for robot manipulation. 
On different target tasks (defined via either pixel locations, goal images or goal classifiers), the transferred model is used together with MPC and designed or learned rewards. Assuming a broad training data distribution, this approach displays good performance even to held-out objects. Several extensions address deformable objects \citep{hoque2020visuospatial, hoque2021visuospatial}, tool use \citep{xie2019improvisation} as well as large scale robot data \citep{dasari2019robonet} and additional egocentric human videos~\citep{schmeckpeper2020learning}. 
This direction can further be combined with temporally extended sub-goals ~\citep{nair2019hierarchical, paxton2019visual, nair2020goal} and other forms of low-level skills ~\citep{wu2021example}. 

A concurrent line of work investigates structured variants of predictive models. These can involve models that take into account the dynamics of rigid bodies interacting with manipulators~\citep{byravan2017se3, nematollahi2020hindsight} or adding relational inductive biases to improve the generalisation and transfer ability of dynamics models through the use of Graph Neural Networks~\citep{wu2020comprehensive}. 
The latter approach has shown good results in learning models of simulated characters and robot arms that transfer to different morphologies ~\citep{sanchez2018graph} as well as for learning object-centric visual predictive models~\citep{veerapaneni2020entity} which can be transferred to predict visual dynamics of different scenes. 

\paragraph{Fine-Tuning}
While zero-shot transfer is a predominant paradigm in model-based RL, performance can be improved via fine-tuning the model directly on the target task \citep{byravan2020imagined, Byravan2021EvaluatingMP, ha2018world}. Fine-tuning can further improve models generated via task-agnostic exploration~\citep{sekar2020planning, bucher2020adversarial}, or when the model is pre-trained on existing offline datasets ~\citep{dasari2019robonet}. 
Finally, there is a small set of works which consider using model-free RL for fine-tuning~\citep{nagabandi2018neural}.

\paragraph{Representations} 
Model transfer is oftentimes combined with representation learning. Prominent examples include the use of state or observation prediction as objective to train temporally consistent representations. After training, we can transfer either the representation in isolation or together with the pre-trained dynamics model~\citep{zhang2019solar, ha2018world}. 

\paragraph{Hierarchy - Composition} Several papers consider different approaches to introduce additional structure into models to improve transfer. Composition structures the model to be built of multiple parts with increased flexibility for transfer to downstream tasks.

Examples are given by combining a global dynamics model which is valid for the entire state space with a task-specific local model~\citep{fu2016one}. This direction has been further extended to learn directly from images by learning latent representations and subsequently predictive models within this latent space~\citep{zhang2019solar, ha2018world}. 

Decomposed models have been used to transfer knowledge across different robot embodiments, enabling more effective transfer for solving similar tasks across different robots. A simple approach is to train two separate models for agent dynamics and environment dynamics; the latter environment model can be transferred efficiently to different robot embodiments~\citep{antoine2021mbrl} and, in addition, can build on data from multiple embodiments~\citep{ kang2021hierarchically}. 
Going one step further, we can train action-free latent representations from diverse video datasets followed by action-conditional model training within a model-based RL loop on the target task~\citep{seo2022reinforcement}. 

\paragraph{Hierarchy - Conditioning} 
Alternatively, conditioning provides a way for adaptation wherein a model takes in additional variables (often a latent embedding) that capture variations across tasks and dynamics. Approaches in this direction separate the problem of adapting the dynamics model into two parts: 1) identifying the current dynamics parameters or embeddings, and 2) making predictions conditioned on these parameters. The former is akin to the task of system identification~\citep{ljung1998system, nelles2001nonlinear, yu2017preparing}, and is often done through the estimation of latent parameters that can either be the underlying physical parameters of the system or a learned representation \citep{killian2017robust, saemundsson2018meta, lee2020context, seo2020trajectory, yen2020experience, kaushik2020fast, belkhale2021model, ball2021augmented, evans2022context}. 

The model's predictions are conditioned on these parameters. Here, different architectures have been explored including Hidden Parameter MDPs ~\citep{killian2017robust}, Gaussian Processes~\citep{saemundsson2018meta, kupcsik2017model} with a recent shift to focus on deep neural networks~\citep{lee2020context, seo2020trajectory, yen2020experience}. 
Different objectives have been proposed such as maximising the likelihood of forward predictions~\citep{saemundsson2018meta, belkhale2021model}, or the consistency of forward and inverse predictions~\citep{lee2020context}. %
On a transfer MDP, the conditioning variables are estimated via an inference procedure, the pre-trained global model is transferred and used together with MPC for action selection. 

Furthermore, these types of context-conditioned models can be used to derive intrinsic bonuses for exploration ~\citep{zhou2019environment}. Combinations of this approach with conditioned policies can enable enable fast policy adaptation to novel environments~\citep{wang2022model}; with further applications in model-based RL instead of pure MPC~\citep{lee2021improving}.

\paragraph{Meta Learning}
Similar to the other KMs, meta learning has been used to train models that quickly adapt to different tasks or dynamics. Approaches following gradient-based meta learning~\citep{finn2017model} are particularly common~\citep{nagabandi2018learning, nagabandi2018deep}. Further approaches leverage recurrent networks for sequence-based meta learning across changes in dynamics and tasks~\citep{nagabandi2018learning, weinstein2017structure}, and Neural Processes~\citep{galashov2019meta}.

\paragraph{Benefits and Disadvantages}
Since dynamics models mainly represent information about the system dynamics, they support transfer by not being tied to a particular reward function or behaviour. 
In principle, a model of the system dynamics could be combined with any reward function (or model thereof) to estimate value functions or perform policy optimisation. 
In practice, the generalisation and transfer capabilities of models are highly dependent on their training data and the inductive bias of the parametric function approximator. 
Furthermore, dynamics models require repeated inference resulting in higher computational requirements for behaviour generation (e.g. MPC, or model-based policy optimisation) which can further be associated with instabilities or compounding errors~\citep{clavera2018model, wang2019benchmarking}.

\subsection{Reward Models}
\label{sec:modalities_transfer_rewards}

Reward functions play a special role in RL, they commonly serve as our interface to design the task to be solved and, in part, the behaviour. 
However, reward functions are not always straightforward to define especially with complex inputs such as raw images or for achieving sufficiently complex behaviours \citep{singh2019end}. 
The challenge can be further split into measuring the state for determining success, designing a reward function for the behaviour or goal of interest, and designing a reward function from which it is easy to learn. These cases often lead to considerable effort in reward engineering \citep{lee2020learning}.
Additionally, auxiliary reward functions, which can be seen as agent-internal instead of coming from the environment \citep{singh2009rewards}, can be learnt to improve agents' performance in particular for hard exploration tasks \citep{Jaderberg2017ReinforcementLW,Wulfmeier2021RepresentationMI}.

Often, learned reward function are transferred together with dynamics models, which often can be estimated from the same data, and used in model-based RL \citep{hafner2019dream}; see additional details in Section \ref{sec:modalities_transfer_dynamics}. However, the application of transferred reward functions in model-free RL is possible as well \citep{Yoo2022LearningMT}.

\paragraph{Generalisation and 0-Shot}
Commonly, reward functions are directly transferred without additional adaptation and directly applied via model-based or model-free RL \citep{singh2019end, ebert2018visual,escontrela2023video}.
The most common source for transferring reward functions is supervised offline training based on existing datasets \citep{singh2019end, ebert2018visual} and added labels for human preferences \citep{christiano2016transfer,cabi2019scaling,lee2021pebble,ARL2015Daniel}.
If expert trajectories are given, either via state or state-action sequences, but no explicit reward labels are provided, we can still extract reward functions via Inverse Reinforcement Learning \citep{ziebart2008maximum, Abbeel2004ApprenticeshipLV, Ratliff2006MaximumMP, wulfmeier2015maximum, ho2016generative, Finn2016GuidedCL, Qi2022ImitatingFA, Fu2019FromLT}.

Different works train reward functions over distributions of tasks to improve generalisation to new tasks.
In these cases, each task is defined via a task descriptor in the same feature space as used during training, e.g. via language \citep{Fu2019FromLT, nair2021learning, cui2022can} or video \citep{chen2021learning}.
For example, it is possible to train a classifier for video similarity which is trained with a broad set of task videos both from human experts and robots \citep{chen2021learning}. In this way, the classifier can generalise as a reward source for future tasks when these are described via demonstration videos from the same distribution. Extending this direction, rewards from simpler tasks can enable effective reward shaping \citep{Ng1999PolicyIU} to accelerate the learning of harder tasks \citep{Konidaris2006AutonomousSK,Brys2015PolicyTU}.

\paragraph{Representations}
A further application is the transfer of pre-trained representations which are transformed into reward functions via manual rules, such as maximising or minimising a particular embedding dimension
\citep{grimm2019disentangled, Wulfmeier2021RepresentationMI, ebert2018visual}. These can be used as auxiliary tasks and for exploration in the target MDP. Recent work has also looked at leveraging pre-trained visual and textual representations from foundation models for generating language/image-conditioned reward functions for a variety of downstream tasks, often unrelated to the data these representations were trained on. While some perform zero-shot transfer of these representations~\citep{cui2022can, mahmoudieh2022zero}, others adapt representations on small amounts of labeled data from the target domain to learn reward functions~\citep{xiao2022robotic}. Representations can further be optimised specifically to enable learning good reward functions for a variety of downstream transfer tasks~\citep{ma2022vip}.

\paragraph{Meta Learning}
Transferring learned reward functions becomes particularly relevant when rewards in the target MDP are unavailable or expensive, but when regular transitions are easily obtained. 
External rewards from direct human feedback provide one example for this situation. Here, meta learning can be applied to optimise for fast adaptation of transferred reward functions with limited external reward feedback \citep{Agarwal2019LearningTG, Li2021MURALMU}.
For example, \citep{Li2021MURALMU} applies gradient-based meta learning \citep{finn2017model} to update a probabilistic reward classifier with small sets of task demonstrations. 
Bayesian Optimisation \citep{Agarwal2019LearningTG} can provide another alternative.
    
\paragraph{Benefits and Disadvantages}
Cheap and automated access to a ground truth reward function renders the transfer of a trained reward model irrelevant. However, reward generation transfer shines when target rewards are directly produced by human operators. Furthermore, efficient reward functions can be hard to manually design and implement in real-world scenarios: optimal behaviour may be implicitly defined (e.g. via human preferences \citep{ravichandar2020recent}), and even well-defined reward functions may require significant instrumentation (e.g. detecting successful states for a robotics task \citep{kalashnikov2021mt}). %

Reward models purely represent the task and commonly have the benefit of being easier to transfer compared to policies given dynamics changes between source and target MDPs \citep{Fu2019FromLT,Bahdanau2019LearningTU}.
At the same time, they require further environment interaction before being able to generate behaviour. Different approaches combine the transfer of both modalities for this reason (e.g. \citep{Qi2022ImitatingFA}).

\subsection{Value Functions}
\label{sec:modalities_transfer_values}

State and state-action value functions haven been successfully applied for transfer in RL, however they entangle information about system dynamics, policy and reward function as well as design choices, such as the horizon and discount factor. 
The fact that different components of the MDP are entangled in the representation of value functions can render it harder to use value functions for transfer. For this reason, historical focus lies on transfer with value functions lies in particular on state-action value function in the critic-only setting where the parameterised value function directly characterises the policy. 
In the following, we provide examples of different methods that overcome these challenges and use value functions for transfer.

\paragraph{Generalisation and 0-Shot}
To enable generalisation, value functions can be trained over distributions of MDPs with the potential for further task or goal conditioning to enable specialisation \citep{kaelbling1993learning}. 
Universal value functions~\citep{schaul2015universal} represent an example of this conditioning. After training, the value functions can generalise to held out tasks based on their task descriptors. Learning the process that provides the conditioning is covered in the hierarchy part of this section. Common forms of universal value functions are presented in \citep{gupta2019relay,nair2018visual} where the task is to reach goal states. 

\paragraph{Fine-Tuning}
Fine-tuning represents the most common method for adapting value functions. Early examples of successfully fine-tuning a value function are present in \citep{selfridge1985,asada1994vision}. In particular, there has been a range of recent works pre-training value functions offline for further online fine-tuning methods~\citep{nair2020accelerating, kostrikov2021offline, lu2022aw, kumar2020conservative, lee2022offline, chebotar2021actionable}.  Similarly, \citep{Lynch2019LearningLP} show how pre-training can be combined with imitation learning for fine-tuning during test time. 
Transfer between simulation and a physical platform represents another common setting~\citep{zhao2020sim}, where the initial pre-training phase is done in simulation and further fine-tuning is performed via online RL in the real world.
When it comes to fine-tuning itself, there are multiple architectural choices available (e.g. training all the weights vs certain layers)~\citep{julian2020never}.
Repeated adaptation of this type is particularly common in various works on RL-training curricula (e.g.~\citep{narvekar2020curriculum, florensa2017reverse}).

\paragraph{Representations}
Unsupervised pre-training of representations is an important topic to improve transfer in RL \citep{bengio2012deep,mesnil2012unsupervised}.
Although representation transfer mostly targets the addition of new final layers, more sophisticated schemes for reusing features and augmenting value functions have been considered~\citep{rusu2016progressive}.  
Going further than pre-training from individual images, representations can take into context temporal patterns by using videos \citep{ma2022vip}. Pre-training happens in this case via a self-supervised goal-conditioned value-function objective.
Further modalities can be included to benefit their specific properties as well. Examples are given by the combination of vision and language representations for downstream robotics tasks \citep{ma2023liv}.

\paragraph{Hierarchy - Conditioning}
Universal value function approximators \citep{schaul2015universal} (UVFAs) commonly are conditioned on a given description of the underlying MDP\citep{kulkarni2016deep, borsa2018universal, grimm2019disentangled} and trained over distributions of MDPs. 
After training, we can manually set the conditioning to the description of the new task. 
If a description is not available, we can instead optimise the conditioning. For example, we can build on Bayesian Optimisation~\citep{arnekvist2019vpe}.
The conditioning can be implicit, e.g. given access to observation histories, an agent can infer sufficient information about the relevant context~\citep{hausman2018learning}. Overall, the generalisation utility of task-conditional value functions depends on a suitable distribution of training tasks as well as on a suitable task descriptions. 
An appropriately conditioned UVFA can also provide a suitable initialisation for further adaptation.
A number of recent works describe how UVFAs can be pre-trained and then adapted to new tasks~\citep{kalashnikov2021mt, eysenbach2020rewriting, yu2021conservative}. %

Goal-conditioned value functions represent a special case of UVFAs where the task is to reach a goal state. While there are different reward functions that can be considered for such tasks (e.g. euclidean distance between the final and the goal state~\citep{gupta2019relay}, or the distance in the goal-embedding space~\citep{nair2018visual}), training of these value functions commonly uses the same mechanism: hindsight relabelling \citep{andrychowicz2017hindsight}. %
Interestingly, goal-conditioned value functions share similarities with model-based RL. Under certain conditions, if the event of reaching the desired goal is associated with a sparse reward of $1.0$, they can be considered as a model of the system dynamics predicting the probability of reaching that goal state from the present state and a taken action given a specific policy~\citep{chebotar2021actionable}.
They can be also used in lieu of successor representations, further discussed under compositional transfer below, as shown in~\citep{janner2020generative}, which allows them to be easily re-purposed to different tasks and reward functions.

While value functions (and other modalities) can be conditioned in various ways to specify tasks, more recently, natural language emerged as a particularly useful interface to facilitate generalisation. The expressiveness and compositionality of language as well as the recent advancements in natural language processing and understanding~\citep{brown2020language,devlin2018bert} render it a promising tool for transfer in reinforcement learning. 
Language has been used in different ways to support transfer, as an abstraction for hierarchical RL~\citep{jiang2019language}, as a general planner that decomposes long-horizon tasks~\citep{huang2022language,ahn2022can}, as a condition modality~\citep{andreas2017modular} or even as a pre-training mechanism~\citep{Reid2022CanWH}. For a thorough analysis of how natural language can inform value functions and RL agents, we refer the reader to the survey on this topic~\citep{luketina2019survey}.

\paragraph{Hierarchy - Composition}
Successor representations \citep{dayan1993improving} and successor features are another common mechanism that facilitates transfer to new tasks~\citep{barreto2016successor,barreto2018transfer,kulkarni2016deep}. 
In general, successor representations and generalised value functions \citep{sutton2011horde} can be considered as partial models which predict the expected future sum of particular features under a policy. Under the assumption that the reward function, describing the target task, is a linear function of the features, the value function can be shown to be linear in the expected discounted sum of these features.  
Maximising over the actions the value model enables one-step policy improvement (or generalised policy improvement \citep{barreto2016successor} if value functions for multiple policies can be evaluated in this manner). 
The appeal of successor representations comes from the fact that they decompose a value function into a reward predictor and a successor map~\citep{kulkarni2016deep}, which mitigates the value-function dependence on a specific reward function. This allows successor representations to be transferable to other, potentially new reward and tasks. 
These specialised value functions allow to compute the value (of a given policy) for any reward function of the above form.
In addition, successor features - extending successor representations past the tabular setting - can be combined to guide exploration and skill discovery for RL agents as shown with variational intrinsic control~\citep{hansen2019fast} and count-based exploration~\citep{machado2020count}.

There are further methods that prepare the value function during the pre-training phase to be particularly amenable to fast adaptation. One commonly-used example of this type of preparation are soft or entropy-regularised value functions~\citep{haarnoja2017reinforcement,haarnoja2018soft}. Soft value functions enable us to retain higher behavioural entropy ~\citep{kumar2020one}, which can facilitate later transfer of the obtained strategies to new tasks. 
The resulting policies can be effectively re-used during the adaptation phase as demonstrated in~\citep{haarnoja2018composable,van2019composing, hunt2019composing}. 
Other examples of creating compositional value functions learn value functions and policies in a way that allows the formulation of new tasks by using Boolean algebra in terms of the negation, disjunction and conjunction of a set of base tasks \citep{van2019composing,nangue2020boolean}.
Finally, the option framework has been applied to both policies, covered in the respective section, and value functions~\citep{singh1992transfer, barreto2019option}. Options enable the decomposition a set of sequential decision tasks and have been applied for transfer from offline datasets \citep{Shiarlis2018TACOLT} as well as related tasks \citep{barreto2019option}.

\paragraph{Meta Learning}

Context-conditioning of value-functions has been a common mechanism to enable effective meta learning. While there exists a variety of viable context types for conditioning~\citep{rakelly2019efficient, fakoor2019meta,laskin2022context}, they all showcase the ability to quickly explore and fine-tune to a new task during deployment.
In addition to context-conditioning, more recently optimisation-based meta learning has gained traction \citep{mitchell2021offline,sung2017learning,bechtle2021meta,houthooft2018evolved,zhou2020online}. 
In particular,~\citep{mitchell2021offline} introduces a offline meta RL setting where the RL agent does not need to interact with the environment and relies on a broad stored dataset, even at meta testing time.

\paragraph{Indirect Transfer}

Instead of directly adapting the online value function, temporal difference learning enables to shape the optimisation objective by using the transferred value function as a target value function on the new task \citep{wang2020target}.
Further regularisation towards expert behaviour can be provided by adding an auxiliary objective such as minimising the KL distance \citep{Agarwal2022ReincarnatingRL}. This method is often used in addition to fine-tuning, to prevent strong divergence from the transferred behaviour \citep{tirumala2020behavior}.

Value functions can further be applied for exploration and data generation in various ways. For example, we can use variational inference to estimate the posterior distribution of the optimal value function for a new reward to improve exploration~\citep{tirinzoni2018transfer}. %
While naive context-based exploration can be effective, it is possible to further explicitly adapt the agent's exploration parameters based on the current experience~\citep{schaul2019adapting}.

\paragraph{Benefits and Disadvantages}

Traditionally value functions can be deemed challenging for transfer because of their strong coupling of information about behaviour policy, environment dynamics and rewards. In addition, they can be more restrictive %
due to explicit dependence on other MDP-specific parameters such as the discount factor and episode length. 
While computing the optimal action requires less effort given a state-action value function than a dynamics model, our standard value functions are bound to make predictions about one reward function.
However, different extensions can provide partial models such as universal value functions, goal-conditioned value functions and successor representations that partially break this coupling and facilitate transfer to new tasks. 
Value-function transfer further allows for RL-based behaviour stitching, where different parts of the state-action space can be connected via the principles of dynamic programming to create chained behaviours as shown in~\citep{singh2020cog}.

\subsection{Policies}
\label{sec:modalities_transfer_policies}

Generating behaviour from policies comes at the lowest computational cost in comparison to all other modalities as we are only required to perform evaluation of the function representing the policy. However, in return, policies are highly specific with respect to both system dynamics as well as task or reward and large changes in either between source and target setting can require further adaptation.

\paragraph{Generalisation and 0-Shot}
Generalisation or 0-shot transfer of policies demonstrably works well in a range of cases \citep{kirk2021survey, openai2019solving,Brohan2022RT1RT}. 
Commonly, these situations can be categorised into aligning source and target domains either by broadening the distribution of source MDPs or otherwise reducing the differences between both domains. The latter point can be addressed via system identification or additional inductive biases \citep{stone2021distracting,muller2018driving,salter2019attention,Zhou2022mj}.

Training over distributions of MDPs has proven helpful for transfer between simulations and real systems via domain randomisation of both visual and physical aspects \citep{Molchanov2019SimtoMultiRealTO, muller2018driving, openai2019solving, tobin2017domain, sadeghi2016cad2rl,bousmalis2018using}. 
Similarly, we can sample training tasks from a distribution \citep{Oh2017ZeroShotTG, Hill2020HumanIW, Cobbe2020LeveragingPG,schaul2015universal} to support transfer over task changes.
If we evaluate such a trained agent on new tasks from the same distribution, effective generalisation is possible. In particular, when we provide the agent with additional conditioning task descriptors, defining the specific task or goal and enabling the agent to adapt with respect to this task space \citep{schaul2015universal}.

When training over a distribution of configurations, our agents will either need to learn conservative solutions which apply to all configurations or rely on a conditioning that enables the identification of a particular configuration.
History conditioning such as fully connected models with stacked observations \citep{Kumar2021RMARM}, LSTMs \citep{openai2019solving, wang2016learning,Humplik2019MetaRL} and transformer-like models \citep{duan2017one} have been used to enable this identification or adaptation as part of the neural network inference process by conditioning on the history of inputs. Learning separate models to control the conditioning is discussed later under the corresponding section.

\paragraph{Fine-Tuning}
In general, policies representing the optimal behaviour for a specific MDP may not transfer well 0-shot to other settings if reward or transition dynamics are sufficiently different.
Challenges for generalisation can exist even within the same MDP due to distributional shifts \citep{Ghosh2021WhyGI}.
However, when target and source MDP are sufficiently similar, the transferred policy can serve as a good starting point for further learning.
One of the most common approach for adaptation is the continuation of regular training and fine-tuning of pre-trained parameters \citep{hinton2006reducing}.
Recent examples for fine-tuning include policies pre-trained with supervised learning \citep{Mirhoseini2020ChipPW,cruz2017pre, davchev2022residual} as well as RL \citep{Schwarzer2021PretrainingRF,Mutti2020APG,held2017probabilistically,tao2021repaint,team2021creating}.

\paragraph{Representations}
Representation transfer is often applied to policies. For example, fine-tuning can be combined with the introduction of a small set of new parameters such as one or multiple added final layers in a neural network \citep{parisotto2015actor}. 
A frequent direction is the pre-training of representations via unsupervised or self-supervised objectives such as input reconstruction (e.g. \citep{lange2010deep, Xiao2022MaskedVP, Wulfmeier2021RepresentationMI, Dittadi2021TheRO,Nair2022R3MAU}).
Expanding past the familiar approaches, we have involved variations of representation transfer.
For example, instead of freezing the earlier layers, we can transfer both an encoder and a policy decoder but adapt only the encoder via a self-supervised objective \citep{Hansen2021SelfSupervisedPA}. 
Another common approach is the combination of transferred policy and new learned policy in the agents action space such as via the addition of predicted actions and modelling a residual instead of the absolute action \citep{Silver2018ResidualPL,Johannink2019ResidualRL}. Finally, recent approaches leverage pre-trained visual and textual representations from foundation models~\citep{radford2021learning, jia2021scaling} to train generalisable policies that improve transfer on a variety of downstream tasks~\citep{shridhar2022cliport, khandelwal2022simple, lynch2022interactive}.

\paragraph{Hierarchy - Conditioning}

Hierarchy represents a regular setting for the transfer of policies, reloading one or multiple low-level policies and learning a new high-level controller for adaptation.
This setting often can be regarded through the lens of distributions over behaviours that approximately cover the optimal behaviour for multiple tasks and can be specialised and filtered by a new higher-level controller. 

Related to generalisation with policies, a simplified perspective on conditioning in RL \citep{kaelbling1993learning,sutton2011horde} treats goals or task descriptors naively as another observation for our agents. If this input in the target aligns with closely to their values in the source MDP, we can obtain better generalisation.
When a single value in this goal space is insufficient, we can optimise another high-level controller to provide inputs to solve the new problem. 

A number of cases can be distinguished. In the simplest setting, we can adapt by training an unconditional high-level controller which does not take into account the agent's observations. 
The problem becomes one of inferring a single embedding, a vector which is additionally provided to the transferred policy as input. 
Examples are given by applying different optimisers such as CMA-ES \citep{Yu2019PolicyTW,yu2019sim,peng2020learning} and Bayesian Optimisation \citep{ arnekvist2019vpe} in the embedding space to maximise performance in the target MDP. 
Different variations of the approach exist such as the optimisation of an embedding in a semantically meaningful space including physical parameters \citep{yu2017preparing} or to improve data efficiency by adding a probing policy to generate trajectories which are most informative \citep{Yang2020SingleEP}. %
Similarly, we can accelerate adaptation via meta learning and apply CMA-ES also during preparation \citep{yu2020learning}.
The overall approach is limited to inferring constant values such as those representing physical parameters of a system, which renders it suited for transfer between simulation and physical systems \citep{yu2019sim,peng2020learning}.

We can additionally condition these high-level controllers on agent observations \citep{heess2016learning, merel2018neural, levy2017learning} but  this additional flexibility can create the requirement for more data than optimising unconditional embeddings.
Commonly, a semantically meaningful space as interface between high- and low-level components, such as moving an locomotion agent into specific directions on a ground plane, can add human interpretability and enable the design for separate rewards for low-level behaviours \citep{levy2017learning,levy2018hierarchical,nachum2018data,nachum2019multi}.
However, instead of defining goals in an interpretable part of the agent's state or observation space, we can also define goals in a learned embedding space \citep{dayan1993feudal,vezhnevets2017feudal}, which can overcome the engineering challenge of designing a task-suited goal space. Across both cases, we commonly define additional reward functions for training the low-level behaviours to fulfil high-level commands.

A final common method to learned interfaces is to treat these as probabilistic latent variables.
Here, a whole range of probabilistic models can be applied including variational autoencoders \citep{Lynch2019LearningLP, merel2018neural,Pertsch2020AcceleratingRL,Pertsch2021DemonstrationGuidedRL, tirumala2020behavior,Rao2021LearningTM, hausman2018learning} or normalising flow-based models \citep{haarnoja2018latent,singh2020parrot}.

\paragraph{Hierarchy - Composition}
Instead of transferring a single low-level controller or skill, a common approach is to transfer sets of skills providing modular behaviours which can be compositionally combined to solve new tasks.
Similar to goal-conditioned policies, modular approaches can generalise 0-shot if the task description is compatible with the transferred policies such that e.g. a new task described by a sequence of sub-tasks can be directly solved by applying the corresponding skills in the described order \citep{Shiarlis2018TACOLT,andreas2017modular}.

Most commonly however, the partial specification given by a set of skills requires further optimisation to identify the optimal behaviour.
This usually amounts to optimising a high-level controller, for instance via model-free RL \citep{daniel2012hierarchical,wulfmeier2020compositional, Peng2019MCPLC}.
A simple case is a bandit formulation for determining the low-level policy choice \citep{Li2018AnOO}. To further reduce the effort of further learning a high-level controller, in some cases we can define manual heuristics \citep{Haarnoja2023LearningAS} or a decision making process based on properties of the low-level skills \citep{Goyal2019ReinforcementLW}. 
When this is impossible and we turn towards training a high-level controller via reinforcement learning, different probabilistic models can be build. This includes mixture \citep{daniel2012hierarchical,wulfmeier2020compositional} or product \citep{Peng2019MCPLC} policy distributions with the high-level controller describing the weighting. 

Extensions of this perspective use transition policies to simplify switching between transferred skills without overlapping state visitation distributions \citep{Lee2018ComposingCS,Lee2021AdversarialSC}. %
Instead of a new explicit high-level controller, it is possible in the actor-critic setting to use the new critic as an implicit controller by comparing or ranking the values of actions from different low-level components \citep{Xie2018LearningWT,Kurenkov2019ACTeachAB}.
To accelerate learning in this setting, different approaches additionally transfer their source critics \citep{hunt2019composing, galashov2020importance}.

Most previously described approaches focus on the hierarchical setting to shape the agent's per-timestep distribution over actions, thereby enabling us to guide exploration.
Depending on the source of pre-trained skills, the agent's search space can be further efficiently reduced via temporal abstraction, following the same  skill for multiple time steps.

The options framework \citep{sutton1999between,pmlr-v32-brunskill14,konidaris2007} is a common approach to use temporal abstraction, where each low-level behaviour, each option, additionally includes a termination condition such that the high-level controller only needs to chose new behaviours when the previous one is completed.
More recent work applies options to neural networks \citep{wulfmeier2020dataefficient, bacon2017option, andreas2017modular}.
Alternatively, we can benefit from  fixed temporal abstraction by following policies for a defined number of steps \citep{Li2020SubpolicyAF, heess2016learning}.

\paragraph{Meta Learning}
Meta learning is often applied to policies with the goal of optimising for effective adaptation.
Common meta learning strategies for policies include sequence-based \citep{wang2016learning,duan2016rl2,mishra2017simple} %
\citep{Humplik2019MetaRL,Zintgraf2020VariBADAV},
optimisation-based 
\citep{finn2017model, Nichol2018OnFM},
and metric-based methods 
\citep{Lengyel2007HippocampalCT}.%
Furthermore, meta learning has been applied to improve policies for exploration from supervised \citep{Gupta2018MetaReinforcementLO, Zintgraf2021ExplorationIA} as well as unsupervised objectives \citep{Gupta2018UnsupervisedMF}.
These applications have been further extended to applying the preparation phase to offline datasets with adaptation being performed during online learning  \citep{Pong2021OfflineML,Dorfman2020OfflineMR}.

\paragraph{Indirect Transfer}

Instead of direct use, we can further integrate the prepared source policy into the training objective for the new agent. A regular application is via the addition of a regularisation term between the action distributions of both agents
\citep{teh2017distral,tirumala2020behavior,ross2011reduction, Czarnecki2019DistillingPD, Schmitt2018KickstartingDR,rusu2015policy,parisotto2015actor, galashov2019information}.
In this case, it can be often necessary to decay the influence of the source policy over the course of learning to avoid learning sub-optimal solutions when the source itself is sub-optimal \citep{Haarnoja2023LearningAS}.

It is possible to regularise an agent's embedding space instead of the final action space. Options here include embeddings for observations and actions for individual timesteps \citep{Gupta2017LearningIF} or extended trajectories \citep{merel2018neural, Bohez2022ImitateAR}.
We can treat this embedding space as an action space for a high-level policy such that methods for hierarchical transfer of low-level behaviours and regularisation in embedding spaces can be conveniently combined \citep{liu2021motor, tirumala2020behavior}.  

A last form of using transferred policies is to shape the optimisation via proposal distributions for model-based RL \citep{Byravan2021EvaluatingMP} or as proposals to be specialised an reweighted via task-specific value functions \citep{galashov2020importance}.

Instead of aligning distributions over actions, we can optimise alignment over visited states or trajectories.
A common approach is the use of adversarial training frameworks to generate rewards based on the discrimination between states visited by agent or transferred expert
\citep{Zhang2020fGAILLF,Wulfmeier2017MutualAT,Stadie2017ThirdPersonIL,kostrikov2019imitation}
By training the discriminator on existing datasets, this method furthermore builds the foundation for transfer of expert knowledge via data instead of policies in Section \ref{sec:modalities_transfer_data}.
Finally, it is possible to apply policies for exploration and to transfer knowledge via the generated data \citep{riedmiller2018learning,Campos2021BeyondFT, torrey2007relational,nachum2019why, Vezzani2022SkillSAS,Smith2023LearningAA}. In the simplest case, the resulting data can be added to the agent's replay buffer for off-policy RL.

\paragraph{Benefits and Disadvantages}
As explicit behaviour representations, policies are well-suited both for fine-tuning and other adaptions as well as shaping the agent's data distribution when used for exploration.
At the same time, they entangle information about the dynamics and rewards and therefore can create the requirement for adaptation as soon as either changes. This presents a stronger theoretical constraint for policies. However, in practice a different reward function often means a different data distribution, which also limits the practical transferability of other modalities such as dynamics models due to challenges in generalisation. 

It is worth considering the differences between policies and state-action value functions with regards to their utility for transfer.  Both are closely related and given a Q-function the computational expense of obtaining the optimal action is comparably low. However, regular policies contain strictly less information. In particular, they only contain information about the relative value of actions but not about their absolute value and are absent of information about the state value \citep{sutton2018reinforcement, peters2008natural}. Whether this additional information is useful in a transfer setting depends on the context. For instance, a simple transformation of the reward may change the absolute value of state-action tuples but may leave the relative value of actions unchanged. Here, a policy would be better suited for transfer and lead to high 0-shot generalisation performance. On the other hand, the state value can be difficult to estimate and prior information from a related task can dramatically speed up estimation in the target task \citep{galashov2020importance}.

\section{Discussions and Conclusions} \label{sec:discussion}

The field of transfer learning has demonstrated considerable growth in recent years. Research in RL is targeting tasks of increasing complexity requiring increasing amounts of experience and computation. This situation renders the ability to reuse knowledge ever more valuable.

A lot has changed since \citet{taylor2009transfer} discussed about thirty methods in their overview of transfer learning in RL, but several key ideas remain unchanged.
Although the research landscape has become more diverse, a strong focus on a subset of knowledge modalities for transfer prevails. While this focus remains relevant in the coming years, we will discuss in Section \ref{sec:discussion:modalities_overview} how recent developments are likely to further diversify transfer learning. 

Overall, the field has made considerable progress across all modalities. Novel ideas have emerged, and new mechanisms for transfer have been introduced and old one extended (Section \ref{sec:discussion:mechanisms}). The increasing scale of RL problems has brought forward new questions which we will discuss in Section \ref{sec:discussion:trends}.
However, prevalent challenges remain for research in transfer learning, in particular around measuring progress. These will be discussed in Section \ref{sec:discussion:benchmarks}.
We will use this final part to discuss each of these aspects as well as future opportunities.

\subsection{Connections and Trade-offs}
After providing in detail surveys of work on each modalities and each transfer mechanism, this section summarises main principles underlying overall progress in the field.

\subsubsection{Comparing Knowledge Modalities}\label{sec:discussion:modalities_overview}
In the previous sections, we describe the transfer via each KM in detail with a broad set of examples. We will use this section to summarise their connections, benefits, and disadvantages. 
We can think of the information represented by the KMs (data, dynamics, rewards, values and policies) as lying on a spectrum that goes from raw and immediately observable and verifiable (such as dynamics) to more processed and requiring additional derivation (such as optimal behaviour). %
As an example, a model of the system dynamics makes prediction about the next state given current state and action. This prediction can be verified against individual transitions in the observed data. Optimal policies, in contrast, make predictions about optimal action choices which, in general, cannot be inferred from a single transition (unless we assume that the current data-generating behaviour is already optimal). The predictions of value functions can be partially verified in the dynamic programming setting via temporal difference error values on individual transitions.

The most processed KMs, policies and value functions, constitute the foundation of much of early work in reinforcement learning \citep{watkins1992q,williams1992simple,sutton2018reinforcement}. Hence, they also form the core of work on transfer learning \citep{taylor2009transfer}. Value functions and policies in particular have the benefit of computationally low cost for behaviour generation in comparison to model-based approaches \citep{Yarats2022MasteringVC} and often simpler infrastructure.
At the same time, they entangle information about dynamics and tasks. Due to this entanglement, changes in either can require adaptation and negatively affect initial generalisation.

On the other hand, reward models and dynamics models can be directly inferred from transition data. %
Because of the disentangled representation of different types of information, these can remain valid if only a part of the system changes. In these cases, the ground-truth reward or dynamics model could be transferred with high generalisation accuracy. 
If the underlying system dynamics change, a dynamics model needs to be re-learned or adapted. However, the reward model can still be valid. On the other hand, if only the task and corresponding reward change, the dynamics can still be valid. In principle, given the latter only the reward model needs to be adapted. In practice, learned models need to generalise outside the distribution that they were trained on. Changing the task generally changes the optimal policy and therefore the visited state-action distribution. This often leads to increased inaccuracy for learned dynamics models \citep{Byravan2021EvaluatingMP}.
Improvements in the generalisation of learned models are likely to increase the importance of transferring modalities which do not entangle information between dynamics and task, i.e. dynamics and reward models. These can include inductive biases for improved generalisation such as broader pre-training datasets \citep{reed2022generalist, Brohan2022RT1RT, Schubert2023AGD} as well as stronger physical inductive biases \citep{Greydanus2019HamiltonianNN}.

\begin{figure}[t]
\centering
\begin{subfigure}[t]{0.48\textwidth}
\includegraphics[width=0.98\textwidth]{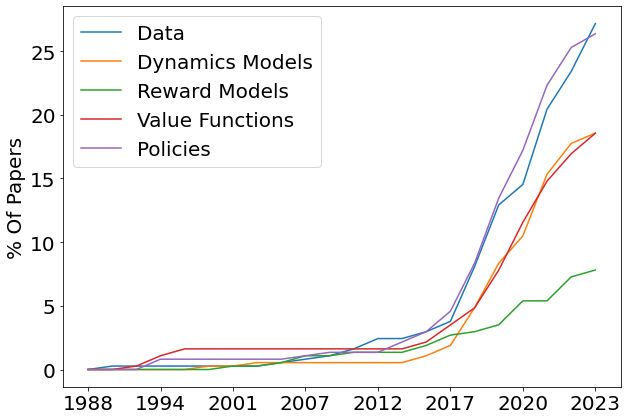}
\caption{Approximate research publications aggregated over the years to the best of our knowledge. The plot visualises the focus on different KM in transfer with a clear focus on data and policy transfer but strong increase across modalities in the last decade.}
\label{fig:stats_km_years}
\end{subfigure}
\hfill
\begin{subfigure}[t]{0.48\textwidth}
\centering
\includegraphics[width=0.94\textwidth]{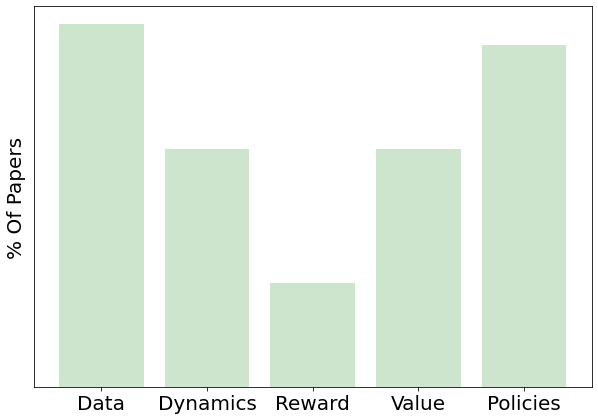}
\caption{The plot shows the percentage of papers as a function of KM as presented in this taxonomy. These numbers to be seen as approximation due to the intractability of collecting all related work. %
}
\label{fig:stats_km}
\end{subfigure}
\caption{Publication statistics for transfer in reinforcement learning.}
\end{figure}

Figure \ref{fig:stats_km} shows the distributions of works presented in this taxonomy as a function of KM as well as transfer mechanisms. Most of the work focuses of transferring policies and indirect transfer of collected data. %
While transfer of dynamics or reward models has been less common in the past due to an increased focus on model-free learning \citep{taylor2009transfer} the number of publications is growing in recent years as described in Sections \ref{sec:modalities_transfer_dynamics} and \ref{sec:modalities_transfer_rewards} and shown in Figure \ref{fig:stats_km_years}. This dynamic is slowly changing in recent times building on growing interest in model-based RL (see Section \ref{sec:modalities_dynamicsmodels}). Many mechanisms that have been thoroughly investigated for policies or value functions, such as compositional transfer \citep{sutton1999between, barreto2016successor, wulfmeier2020dataefficient}, are now becoming targets for model-based research \citep{antoine2021mbrl, kang2021hierarchically, seo2022reinforcement}.

We expect to see increased focus on comparing the effectiveness of different modalities for transfer and hybrid solutions combining multiple perspectives.
As an early step, \citet{Woczyk2022DisentanglingTI} investigate the impact of transferring combinations of policy, value function, and data.
Particular efforts have been made to connect perspectives between dynamics models and value functions \citep{Pong2018TemporalDM,chebotar2021actionable}. The resulting goal-conditioned value functions can be relatively agnostic to task engineering. %
They effectively turn into multi-step probabilistic dynamics models predicting the probability of reaching a given state. By using broad distributions of source MDPs they can provide powerful foundations for transfer with minimal engineering effort.
Combinations of model and policy transfer can combine the benefits of both \citep{Salter2022MO2MO,Walker2023InvestigatingTR}. Furthermore, they can enable adapting policies without access to rewards in the target MDP by sharing an embedding between model and policy, updating purely based on model accuracy during transfer \citep{Ball2021AugmentedWM, yao2018}. 

The relative ranking of computational costs between modalities less dependent of use in transfer or from scratch. 
The cost to learn dynamics and reward models is lower compared to value functions and policies as the learning targets can be directly accessed. Optimal value functions and policies have higher computational cost during learning, their targets cannot directly be observed from data but have to be computed over trajectories. 
Once trained, the computational requirements for obtaining optimal behaviour reverse. 
Dynamics and reward models are regularly used for planning and require rolling out trajectories, increasing computation and repeating model evaluation. Value functions still require an explicit maximisation process to evaluate which action has highest expected value. 
In particular in discrete, low-dimensional action spaces, we can cheaply determine all action's values to generate behaviour. In continuous action spaces, other mechanisms such as sampling-based maximisation are required. 
Finally, policies commonly provide us with the fastest way of obtaining action simply by performing model inference, with the potential of additional sampling from a parameterised distribution. In summary, the KMs trade-off the computational cost for learning with the cost for optimal behaviour generation.

Data, as unprocessed source of information, provides the most general KM for transfer but can also incur high costs. These include additional computation and a delayed impact on the distribution of additional target experience. For instance, hierarchical control using pre-trained policies can generate high performance faster while the transfer via data can lead to slower learning but higher asymptotic performance \citep{Vezzani2022SkillSAS}. For this reason, both modalities can be effectively combined when data is generated by RL agents \citep{Vezzani2022SkillSAS} or includes human demonstrations \citep{zhou2019watch} for the same task as well as task-agnostic data \citep{kumar2022pre}.
Data transfer, due to the added time and computational requirement to extract behaviour, remains particularly relevant in domains without strong requirements on the time-effectiveness of learning \citep{Smith2023LearningAA}.  

While we focus on the main knowledge modalities for RL, which directly relate to components of the MDP and the RL agent, other forms of information can be transferred as well. 
In particular, information related to the optimisation procedure can accelerate learning. Examples are given by the application of evolution-based learning to optimise parameters of the loss function \citep{houthooft2018evolved} or meta learning the loss for Q-Learning \citep{Xu2018MetaGradientRL, bechtle2021meta}.
     
\subsubsection{Comparing Transfer Mechanisms}\label{sec:discussion:mechanisms}

Improving our understanding of the techniques applied for transfer is critical as even the most commonly used approaches are not fully understood \citep{white2017unifying,mahadevan1996average,pitis2019rethinking}. 
Intuitively, direct use of modalities can create higher initial performance (see Section \ref{sec:transfer}) than their indirect use since it affects the initial behaviour in comparison to randomly initialised functions \citep{tirumala2020behavior, wulfmeier2020compositional}. At the same time, it can lead to a loss of flexibility of the training process \citep{Nikishin2022ThePB,igl2020transient} as we will discuss over the next paragraphs. 

Fine-tuning remains one of the dominant forms of transfer, it simply starts training with a pre-trained model instead of a randomly initialised one. This approach has the benefit of not requiring additional change to existing algorithms other than different initial parameters.
However, recent studies show that starting the optimisation of initial randomised models instead of pre-trained ones can lead to higher asymptotic performance given high distance between source and target domains both in supervised \citep{ash2020warm} and reinforcement learning \citep{igl2020transient, Nikishin2022ThePB} with neural networks.
In reinforcement learning, this behaviour can have multiple reasons.
Saturated non-linearities and higher parameter weights can become harder to adapt than the initial model and affect the straightforward applications of fine-tuning in supervised learning \citep{ash2020warm}. In RL the initial models can further affect exploration \citep{Campos2021BeyondFT, Vezzani2022SkillSAS}. 
In the supervised case, recent work additionally shows that fine-tuning for out-of-distribution adaptation can lead to strong performance reduction in comparison to representation transfer via the addition of further linear layers \citep{Kumar2022FineTuningCD}.
Generally speaking, the performance of transfer in RL strongly depends on the models we use to represent knowledge. The recent focus on deep neural networks affects what type of transfer works well and further understanding and improvements regarding the general training of these models will in return improve and shape the landscape of transfer learning.

Direct mechanisms for transfer enable the agent to rely on the transferred knowledge from the start of the experiment and generally provide the best way for high initial performance. Therefore, direct mechanisms are the default methods when the key focus lies on generalisation. Indirect transfer both via auxiliary objectives or via data can provide additional flexibility by not constraining any aspects of the agent - such as model choice. 
In this case, auxiliary objectives will often bias final solution which can be detrimental if the applied knowledge modality is highly sub-optimal. Solutions for annealing of additional objectives exist \citep{Haarnoja2023LearningAS} but introduce additional complexity and hyperparameters.
Transfer via data generally relies on being able to generate a useful training signal from this data. For example, transferring data can introduce challenges if the reward signal in the used data is not rich enough \citep{singh2020cog}, but adds further flexibility in comparison to changing the objective \citep{Vezzani2022SkillSAS}.

\subsection{Benchmarks for Transfer Learning}
\label{sec:discussion:benchmarks}
Supervised and unsupervised learning, particularly for vision and language, have familiar, standard benchmarks and datasets which have supported scientific communication and measurement of progress \citep{Russakovsky2015ImageNetLS, Sun_2020_CVPR, Lin2014MicrosoftCC,Cordts2016TheCD, merity2016pointer, wang2019glue}. 
Moreover, efforts have been expanded to establish benchmarks for assessing transfer in the context of fixed datasets \citep{Venkateswara2017DeepHN, Kornblith2019DoBI}.

When studying transfer learning in RL, we encounter a pivotal challenge: a comparably lower degree of standardisation of benchmarks, datasets, tasks and evaluation protocols. This situation has considerably improved but its core persists since the early days of transfer for reinforcement learning \citep{taylor2009transfer}.
Designing a benchmark for RL is of increased complexity with a breadth of possible evaluation criteria (see Section \ref{sec:transfer}). In contrast to the fixed datasets employed in other learning paradigms, RL demands dynamic environment interaction for evaluation, introducing further difficulties. It poses particular challenges in the case of non-simulatable domains where platforms or their access have to be shared \citep{Bauer2021RealRC}.
Moreover, the requirement for environment interaction creates additional computational costs, particularly when given intricate simulations \citep{degrave2022magnetic}.

Correspondingly, in the reinforcement learning literature in comparison to computer vision, there are relatively fewer large-scale domains and widespread benchmarks. Notable exceptions predominantly manifest in smaller-scale tasks \citep{tassa2018control,openaigym} that lend themselves to effective solutions without necessitating prior knowledge.
Visual complexity is increased in a set of extensions towards more complex 3D vision-based environments \citep{Beattie2016DeepMindL, habitat19iccv,Guss2019MineRLAL}. These benchmarks have the potential to be further expanded for transfer learning as tasks become intractable from scratch or require escalating computational resources. 
Physical benchmarks can separate us from the constraints of simulations but provide additional challenges \citep{Bauer2021RealRC} and therefore represent a minority of existing benchmarks. %

The reinforcement learning literature, in its current state, provides a collection of benchmarks related to specific facets of transfer learning. In the ensuing discussion, we aim to provide a succinct yet high-level overview with practical applications in mind. In addition to discussing existing benchmarks, our emphasis extends to RL environments suited for transformation into benchmarks in the coming years.
In particular, the recent consensus on shared from-scratch RL benchmarks enabled the development of benchmarks explicitly targeting transfer in RL with stronger community support \citep{matthews2022skillhack,muller2021procedural,bellemare2013arcade}.
Two particular fields have gained notable attention: Research on meta learning \citep{yu2020meta,Cobbe2020LeveragingPG} and offline RL \citep{Gulcehre_2020,Fu_2020}. These fields have attracted strong communities dedicated to benchmarking transfer, with a focal point on optimising adaptation efficiency as well as facilitating transfer from offline datasets.  

For the robotics domain, several manipulation benchmarks enable transfer across tasks. 
Benchmarks such as Metaworld \citep{yu2020meta}, Robodesk \citep{kannan2021robodesk}, RLBench \citep{james2019rlbench} and KitchenShift \citep{xing2021kitchenshift} maintain a consistent embodiment while introducing variations in tasks. 
Kitchenshift also tests for transfer with respect to the environment, including transformation in the state and observation spaces. 
Notably, KitchenShift goes a step further, evaluating transfer concerning the environment itself, including transformations in both state and observation spaces.
In a parallel vein, Robosuite \citep{robosuite2020} and ManiSkill \citep{mu2021maniskill} extend their evaluations to assess transfer within different embodiments and environments. CausalWorld \citep{ahmed2021causalworld} takes a distinctive approach, benchmarking causal reasoning and providing a diverse array of procedurally generated experimental setups to facilitate testing for transfer to novel environments.

Beyond the domain of manipulation, various benchmarks cater to more general control tasks, often leveraging the DeepMind Control Suite (DMC) \citep{tassa2018control}. These benchmarks, emphasising generalisation and transfer, frequently incorporate evaluations of transfer to disturbances in observations. 
Examples include the Distracting Control Suite \citep{stone2021distracting}, DMC-GB \citep{hansen2021stabilizing}, and DMC Remastered \citep{grigsby2020measuring}, all examining transfer concerning alterations in sensory input, particularly changes in the backgrounds of control task environments.
Benchmarks like Noisy MuJoCo \citep{zhao2019investigating} and GenAsses \citep{packer2018assessing} perturb the environment's physics parameters to assess transfer capabilities in response to changing dynamics. Collectively, these benchmarks provide a robust setting for systematically exploring transfer learning and leave space for extension to fully integrated benchmarking across all axes.

The investigation of transfer for reinforcement learning transcends traditional domains like robotics and control, extending to encompass general reasoning problems and game scenarios. ProcGen \citep{Cobbe2020LeveragingPG} and Crafter \citep{hafner2021benchmarking} are procedurally generated environments placing an agent in 2D game-like environments. Both of these benchmarks supply randomised content and allow for diverse variations in the environment. 
CARL \citep{BenEim2021a} offers additional perspectives by facilitating extensions of multiple RL environments for transfer to distinct tasks. Construction \citep{bapst2019structured} represents a benchmark for testing the agent's ability to tackle demanding physical construction tasks, constructing structures from blocks and assessing transfer to novel challenges.
On the meta RL side, Alchemy \citep{wang2021alchemy} introduces a game-like environment that probes the agent's high-level reasoning capacities within the benchmark's environment. 
Complementary to these benchmarks are recent advancements in powerful simulators, including Habitat \citep{habitat19iccv, szot2021habitat}, Omniverse \citep{omniverse}, and iGibson \citep{shen2021igibson}. These simulators offer a foundation for generating varied simulated environments for transfer-related benchmarks. In Table \ref{table:benchmarks}, we provide a simplified overview of various benchmarks.

\begin{table*}[ht]
\centering
\begin{tabular}{|l|clc|l|clc}
\hline
\multicolumn{1}{|c|}{\textbf{\begin{tabular}[c]{@{}c@{}}Benchmark\\ Environment\end{tabular}}} & \multicolumn{3}{c|}{\textbf{Transfer Focus}}                 & \textbf{\begin{tabular}[c]{@{}l@{}}Benchmark\\ Environment\end{tabular}} & \multicolumn{3}{c|}{\textbf{Transfer Focus}}                 \\ \hline
Metaworld                                                                                      & \multicolumn{2}{c}{$r_t$}               & adaptation               & DMC-GB                                                                   & \multicolumn{2}{c}{$o_t$}               & generalisation           \\ \cline{1-1} \cline{5-5}
Robodesk                                                                                       & \multicolumn{2}{c}{$r_t$}               & adaptation               & Noisy Mujoco                                                             & \multicolumn{2}{c}{$o_t$}               & generalisation           \\ \cline{1-1} \cline{5-5}
RLBench                                                                                        & \multicolumn{2}{c}{$r_t$}               & adaptation               & GenAsses                                                                 & \multicolumn{2}{c}{$o_t$, $s_t$}        & generalisation           \\ \cline{1-1} \cline{5-5}
KitchenShift                                                                                   & \multicolumn{2}{c}{$o_t$, $r_t$}               & generalisation           & Alchemy                                                                  & \multicolumn{2}{c}{$o_t$, $r_t$}        & adaptation               \\ \cline{1-1} \cline{5-5}
RoboSuite                                                                                      & \multicolumn{2}{c}{$o_t$, $s_t$, $r_t$} & generalisation           & CARL                                                                     & \multicolumn{2}{c}{$o_t$, $s_t$, $r_t$} & generalisation           \\ \cline{1-1} \cline{5-5}
ManiSkill                                                                                      & \multicolumn{2}{c}{$o_t$, $r_t$}        & generalisation           & Construction                                                             & \multicolumn{2}{c}{$r_t$}               & adaptation               \\ \cline{1-1} \cline{5-5}
CausalWorld                                                                                    & \multicolumn{2}{c}{$o_t$, $r_t$g}       & adaptation               & Procgen                                                                  & \multicolumn{2}{c}{$o_t$, $r_t$}        & generalisation           \\ \cline{1-1} \cline{5-5}
Distracting DMC                                                                                & \multicolumn{2}{c}{$o_t$}               & generalisation           & Crafter                                                                  & \multicolumn{2}{c}{$o_t$, $r_t$}        & generalisation           \\ \cline{1-1} \cline{5-5}
\end{tabular}
\caption{Comparison of different benchmarks with respect to their transfer requirements. Apart from the remainder of the taxonomy, here we explicitly discuss changes with respect to state space $s$ and observation space $o$. While some of these benchmarks were not specifically designed for transfer they can be used to benchmark transfer. The second column shows the key changes for each benchmark.}\label{table:benchmarks}
\end{table*}

\subsection{Ongoing and New Trends}\label{sec:discussion:trends}
While transfer learning has shown remarkable successes over the past few years, there are still several challenges that limit broad applicability. Particularly, transfer still often focuses on individual tasks and domains or smaller sets thereof, limiting the scale of change that can effectively be targeted \citep{yu2020meta}. 
In order to truly scale transfer via generalisation and adaption, we are required to scale our foundations, i.e. the richness of our training distributions presented as datasets and environments.  
The 'bitter lesson' \citep{bitter} of research in artificial intelligence generalises to transfer learning. We are implored to investigate methods which will ultimately scale in their ability to transfer, to generalise and adapt, with the underlying data distributions.

We believe that several ongoing and new trends in the machine learning community can help to accelerate research along this path. They include: 1) modular and distributed learning to enable sharing and open collaborations, 2) in-context and meta learning to explicitly scale adaptation abilities with larger knowledge sources, 3) the related improvements in large pre-trained models as foundation for integrating various sources of knowledge about the world, tasks, and controlled agent embodiments, and 4) truly large-scale sources for datasets to support training these models. %
These aspects are discussed in further detail in the following section as we believe that they have the potential to critically shape the future of the research landscape.

\subsubsection{Modularity \& Distributed Learning}
Similar to progress in computer science and software development, machine and reinforcement learning will benefit from improved sharing of progress across teams and building on collective work. Distributed updates and expansion of trained models and datasets has the potential to replicate the benefits of open collaborations to ML and RL.
Initial examples for ML cover the collective generation of datasets \citep{Scao2022BLOOMA1}, models \citep{Touvron2023Llama2O, MosaicML2023Introducing} and benchmarks \citep{Srivastava2022BeyondTI}. Increasing consensus on shared tasks and data formats enables this progress.
Following this direction in RL presents additional challenges due to lower agreement on common platforms, tasks, and large-scale benchmarks and increasing complexity of evaluation as discussed in Section \ref{sec:discussion:benchmarks}.

Initial efforts aiming at sharing knowledge across teams take place with focus on transferring complete, monolithic objects - sharing whole datasets \citep{Fu_2020, Gulcehre_2020} or trained modalities \citep{downloaddt,rl-zoo3}. 
Modularly transferring knowledge modalities for later combination opens up a range of opportunities and further challenges.
While obtaining and combining data from different sources becomes more common in ML (with examples given in natural language processing by creating supersets combining individual datasets \citep{Gao2020ThePA}), the process gains in complexity for merging trained models \citep{Lawson2023MergingDT, tan2022renaissance}.
Transferring learnt functions or models remains critical though as data generally requires additional computation and therefore financial cost \citep{Chowdhery2022PaLMSL} as further discussed in Section \ref{sec:modalities_behavior}

Developments in software engineering have led to version control and related tools for sharing and integrating functions represented as manually-written code.
We require their equals for trained machine learning models. 
ML research with fixed datasets has already progressed in this direction and methods have been developed for distributed and federated forms of transfer learning with similarity to software development \citep{raffelcall}. 
Recent work enables small, distributed changes to massive neural networks including the adaptation of small subsets of all parameters \citep{guo2020parameter}, additional adapter modules \citep{houlsby2019parameter,sharma2023lossless}, changing representations with backwards compatibility on previous predictions \citep{shen2020towards}
or the creation of knowledge bases for external retrieval \citep{Guu2020REALMRL}. 
Mechanisms have been developed for different aspect of version control such as merging of combining multiple models \citep{Matena2021MergingMW}. Investigating related mechanisms for RL agents provides fertile ground for future research in particular with respect to teams without access to the critical large-scale compute required to train massive policies \citep{VinyalsStarCraft, reed2022generalist, openai2019dota}. 
An early example is given by \citet{Gehring2021HierarchicalSF} who provide individual trained skills as policies for download and combination by other researchers.

\subsubsection{In-Context \& Meta Learning}
The field of meta learning has significantly grown over the last decades. It originates in the context of learning to learn 
\citep{thrun1998lifelong,schmidhuber1987srl} and computational demands led to initially slower growth \citep{taylor2009transfer}. 
Recent years have seen in uptake in the field from learning representations that can quickly adapt \citep{finn2017model, mendonca2019guided, Gupta2018MetaReinforcementLO} to learning the optimisation mechanism itself \citep{duan2016rl2,l2l} to reward functions and objectives \citep{houthooft2018evolved,bechtle2021meta,sung2017learning,xu2020meta}. 

In particular, sequence or model-based approaches to meta learning are gaining in importance by effectively leveraging truly massive datasets and models for improved generalisation. 
This development shifts the field towards more implicit forms of meta learning which do not require additional changes in the learning algorithm but rely on model architecture and data format. From the perspective of language modelling, this perspective is closely related to in-context learning where model parameters are not explicitly optimised. Instead adaptation builds on the model inference step itself \citep{Radford2019LanguageMA,brown2020language}. 
Initial investigations in the space of large language models (LLMs) aim to increase our understanding of the types of optimisation possible via ICL and its connection to adaptation via gradient descent fine-tuning \citep{dai2022can,min2021metaicl}. In particular, connections have been drawn to to linear functions, two-layer neural networks, and decision trees
\citep{garg2022can,von2023transformers}; and gradient descent, ridge and least-squares regression \citep{akyurek2022learning}. We are hopeful that these insights will inform further advances in RL.

As model architectures, transformers have gained increased relevance in the offline setting \citep{nguyen2022transformer,laskin2022context,lee2022multi,Chen2021DecisionTR}, training on large offline datasets, as well as the online setting \citep{parisotto2020stabilizing,Team2023HumanTimescaleAI,zheng2022online} training over distributions of MDPs.
Training online strongly relies on improvement via RL mechanisms \citep{Team2023HumanTimescaleAI,parisotto2020stabilizing}. In the offline setting, we can distinguish between training with offline RL \citep{yuan2022robust,pong2022offline,Chen2021DecisionTR} or pure imitation such as behavioural cloning \citep{reed2022generalist,bousmalis2023robocat}. 
The additionally availability of rewards enables offline meta RL to disentangle the impact of reward and dynamics changes in order to improve transfer across tasks \citep{yuan2022robust}. This separation further allows to meta train for adaptation with additional unsupervised online data (i.e. without rewards) to bridge the distribution shift between the offline and online settings \citep{pong2022offline}. %

Recent theoretical work describes how generalisation capabilities to new tasks are strongly related to the distance between training and test distributions for gradient-based meta RL \citep{fallah2021generalization}. 
This demonstrates that generalisation to tasks further away from the training distribution remains one of the key challenges of current meta learning and therefore transfer. 
Going forward, a critical role for meta learning will lie in the improved training and understanding of large pre-trained models also described as foundation models and analogous investigations for the sequence or model-based side of meta learning are highly valuable.

\subsubsection{Large Pre-trained \& Foundation Models}\label{sec:foundation_models}
In line with the previous discussion of broader pre-training distributions, transfer (and specifically generalisation) via foundation models gains increased focus. This includes large language models or vision language models pre-trained in the self- or unsupervised setting via generative modelling of large data corpora \citep{bommasani2021opportunities}. Without further change in parameters, these models benefit from in-context learning in the inference step of the model with strong 0-shot generalisation performance to a wide range of tasks \citep{Radford2019LanguageMA}.

Partially inspired by this success in supervised learning, goals within the RL research community have extended towards more ambitious transfer settings. Interest in broader distributions over target tasks \citep{bousmalis2023robocat} and transfer across agent embodiments \citep{Brohan2022RT1RT,Brohan2023RT2VM} are leading to an increasingly crucial role for foundation models.
In RL, the use foundation models can take different shapes in relation to the discussed knowledge modalities \citep{lee2022multi,reed2022generalist,Micheli2022TransformersAS,Chen2021DecisionTR, bousmalis2023robocat, Schubert2023AGD, di2023towards, driess2023palm} and represent system dynamics, task or rewards, values and behaviour.
This diversity opens a gamut of research questions along the discussed differences and relations between modalities. For a survey of FMs in general decision making we refer the interested reader to \citep{Yang2023FoundationMF}.

RL can benefit from FMs in two principal ways: 1. via the direct and indirect use of already pre-trained models, largely utilising the knowledge extracted from the original (mainly action-free) datasets; or 2. via the use of the underlying untrained models when training on large RL specific datasets (mainly including actions). 
The first path provides a way of injecting general-purpose world knowledge conveyed via language and vision at different levels of abstraction integrated with RL-specific data for grounding in the agents environment. %
We start seeing various applications of FMs integrated as pre-trained modules in this setting to strengthen generalisation with respect to semantic and visual shifts among others \citep{Brohan2023RT2VM}.
Recent examples include the use of large language models as robot planners to decompose long-horizon tasks~\citep{huang2022language,ahn2022can,di2023towards} into smaller, actionable steps additionally verified by robotics affordances. 
The application of foundation models for transfer go beyond planning, where they can be used as feedback mechanisms~\citep{huang2022inner} or where their language aspects can be used as a connecting medium between various separately pre-trained modalities~\citep{zeng2022socratic,Alayrac2022FlamingoAV}. This ability of combining different foundation models provides an critical alternative to approaches fine-tuning a single multi-modal model as a way to improve transfer learning \citep{Brohan2023RT2VM, driess2023palm}.

Foundation models have also been used for imbuing general purpose object knowledge through the use of pre-trained representations~\citep{radford2021learning} as well as for enabling understanding of the scene semantics (e.g. for visual navigation  \citep{shafiullah2022clip, huang2022visual}). Oftentimes, these models are relatively embodiment and environment agnostic as they are trained on large corpora covering distributions of settings. Similarly, large models have been used for producing general purpose language-conditioned reward functions with similar benefits~\citep{fan2022minedojo}. Finally, they can be applied for data augmentation in settings with small amounts of in-domain data~\citep{xiao2022robotic, yu2023scaling}. %
The other path for FMs to accelerate progress in transfer learning uses the underlying models but trains completely with specific data including actions \citep{bousmalis2023robocat, reed2022generalist, lee2022multi, Chen2021DecisionTR, Brohan2022RT1RT, Schubert2023AGD}. Models with the ability to fully exploit massive RL-specific datasets cover a wide distribution over tasks, appearance, embodiments and domains to simplify transfer with respect to these factors. This perspective is currently less common due to the high cost of collection large demonstration datasets \citep{Brohan2023RT2VM} but further work towards automating data collection \citep{bousmalis2023robocat} and sharing data across teams \citep{Padalkar2023OpenXR} have the potential for fast growth.

Leveraging the knowledge compressed into foundation models, such as by direct and indirect transfer, can help to tackle a wide variety of transfer settings, overcoming the limitations of present transfer methods. 
But in order to train models of increasing scale, corresponding growth is also required on the data side. 
Due to their large parameter count, foundation models impose additional challenges and costs for transfer - in particular adaptation. Fine-tuning all parameters of the model often becomes too expensive. Therefore, computationally efficient adaptation gains critical importance. In particular for LLMs, research has led to advancements in optimising small subsets or additional sets of parameters rather than training the whole model \citep{Hu2021LoRALA}.
LLMs commonly rely on prompts, i.e. text input, to describe tasks similar to additional conditioning information described in section \ref{sec:transfer_prep_adapt}. 
In addition to manual optimisation of prompts as conditioning \citep{Kojima2022LargeLM}, a set of techniques has emerged for automatically optimising them \citep{Lester2021ThePO,Liu2021GPTUT}.
Furthermore, new parameters and modules can be added to FMs to simplify adaptation \citep{li2021prefix,Hu2021LoRALA,houlsby2019parameter,Alayrac2022FlamingoAV} and trained models can be effectively merged to reduce inference and memory costs \citep{ainsworth2022git,Stoica2023ZipItMM}.

\subsubsection{Large-Scale Data Sources}\label{sec:knowledge_sources}
Current work in reinforcement learning often builds on transferring from a small set of tasks or experiments in contrast to recent research in natural language and computer vision. 
Transfer across more distant tasks and domains is often limited, both for artificial \citep{yu2020meta} and natural intelligence \citep{SALA201755}. Therefore, a broad source training distribution over relevant dimensions for transfer is critical. Online training and interaction in parallel with a very wide set of environments brings substantial challenges. However, these can be at least partially mitigated by focusing on integrating information into large datasets and on training models offline.
When sufficiently related datasets are given, and do not require additional effort for their creation, our metrics shift further towards pure measurement of target data instead if combined target and source \citep{taylor2009transfer}.

As our ability grows to train larger and more capable models, sufficient domain-specific data is not often the standard setting. Here, transfer from additional, unsupervised datasets can further affect scaling laws \citep{Hernandez2021ScalingLF}. 
Continued investigation is needed for RL because of elementary differences to learning from fixed datasets caused by entangling data generation and learning. Initial examples analyse transfer from broad, unstructured datasets to improve learning in RL \citep{Levine2021UnderstandingTW,reed2022generalist, xiao2022robotic}. While there are few current cases, the question of how to transfer from existing, potentially domain-agnostic datasets and more effective sharing of data and other modalities across projects invariably increases in relevance \citep{Padalkar2023OpenXR}.

Unlike language and vision, decision making data is highly limited \citep{bommasani2021opportunities}. 
In this context, the wider availability of large non RL-specific data sources could render them a precondition for future advancements in transfer.  
Image and video datasets present such a source, both to directly shape RL training via objectives to optimise policies \citep{bahl2022human} and create reward functions \citep{peng2018sfv, aytar2018playing} and for general representation pre-training \citep{pmlr-v205-radosavovic23a}.
For example, datasets created from YouTube videos have been applied for highly effective pre-training, leading to stronger results than the more common use of ImageNet data  \citep{sivakumar2022robotic,Xiao2022MaskedVP,pmlr-v205-radosavovic23a}.
Similarly, the use of unlabelled massive datasets such as video datasets to shape reward functions \citep{peng2018sfv} and the augmentation of existing offline data \citep{Lambert2022TheCO} is increasing.
Another current, principal data source is text, which can be integrated in various ways. Examples of direct transfer from this different, and intuitively less-related domain are given by pre-training RL agents \citep{Reid2022CanWH} or co-training agents with text to improve generalisation \citep{reed2022generalist,Brohan2023RT2VM}.
Even when intuitively we cannot see shared structure between these domains, it might for example help by improving the initialisation of neural networks similar to manual improvements in initialisation methods \citep{glorot2010understanding}.
Another form of leveraging knowledge extracted from text data is exemplified by modularly combining a pre-trained language model with low-level motor skills to produce robot behaviour that is semantically grounded by the language model \citep{ahn2022can,singh2023progprompt}. Here, the language models take over the role of a high-level controller or planner which connects to the underlying controller via manually defined interfaces.

In tandem with external sources of datasets, the process of data generation assumes a pivotal role within. This facet offers further avenues to generate large RL-based datasets to bootstrap new experiments \citep{Agarwal2022ReincarnatingRL}.
While data generation on physical hardware can incur substantial costs, RL-generated data from related domains \citep{Levine2021UnderstandingTW, Ebert2022BridgeDB}, particularly in large-scale simulated environments, has gained prominence \citep{zhao2020sim,Hofer2020PerspectivesOS,degrave2022magnetic, Haarnoja2023LearningAS}. 
The recent strides in research on robotic legged locomotion, particularly with quadrupedal robots, exemplify progress harnessing rapid simulations to generate intricate reactive behaviors ~\citep{lee2020learning, hwangbo2019learning}.
Likewise, contemporary work leverages simulated visual environments to showcase impressive sim2real transfer across various manipulation tasks \citep{chen2022system,lee2021beyond} including complex, long-horizon pick-and-place ~\citep{yokoyama2023adaptive, gervet2023navigating}.
The application domain extends to high-frequency control for agile quadrotor flight, as demonstrated by recent work largely transferring from simulation \citep{kaufmann2023champion}. These multifaceted examples underscore the role of data generation in advancing RL research.

\subsection{Concluding Remarks}
Transfer in reinforcement learning has seen critical advancements over the last decade.
With increasing interest in general, large-scale agents, it has only grown in importance, propelling it to the forefront of research endeavors.
A paradigm shift is evident in the transition from relatively confined source domains to more expansive foundations, extracting knowledge from massive datasets and wide-ranging pre-training distributions.
This development naturally aligns with the recent emergence of foundation models and further trends facilitating efficient learning in modular and distributed settings. 

Our taxonomy provides a systematic framework for comprehending knowledge transfer via its modalities and mechanisms. By discussing the trade-offs and interconnections inherent in this taxonomy, we aim to forge a stronger foundation for future insights and advancements.
As we look ahead to the coming decade, the landscape of reinforcement learning research is poised to grapple with new challenges propelled by increasingly ambitious goals and the corresponding required resources. We believe that transfer learning will assume a pivotal role in tackling these challenges.

%% file: acks.tex
The authors would like to thank the colleagues and friends whose invaluable contributions have enriched this work (in alphabetical order) Andre Barreto, Michael Bloesch, Yevgen Chebothar, Todor Davchev, Jessica Hamrick, Leonard Hasenclever, Dushyant Rao, Karl Tuyls, and Theophane Weber.
Their engaging discussions, meticulous reviews, and constructive feedback during the earlier drafts have significantly shaped its current version.

Additionally, our appreciation extends to colleagues who have been instrumental in our long-standing pursuit of improving transfer for reinforcement learning (in alphabetical order) Abbas Abdolmaleki, Steven Bohez, Philemon Brakel, Roland Hafner, Martin Riedmiller, Tobias Springenberg, Oleg Sushkov, Dhruva Tirumala, and Giulia Vezzani. Their collaborative spirit and dedication have been pivotal to the progress of our work over the years.

%% file: main.bbl
\begin{thebibliography}{486}
\providecommand{\natexlab}[1]{#1}
\providecommand{\url}[1]{\texttt{#1}}
\expandafter\ifx\csname urlstyle\endcsname\relax
  \providecommand{\doi}[1]{doi: #1}\else
  \providecommand{\doi}{doi: \begingroup \urlstyle{rm}\Url}\fi

\bibitem[Abbeel and Ng(2004)]{Abbeel2004ApprenticeshipLV}
P.~Abbeel and A.~Ng.
\newblock Apprenticeship learning via inverse reinforcement learning.
\newblock \emph{Proceedings of the twenty-first international conference on
  Machine learning}, 2004.

\bibitem[Abdolmaleki et~al.(2021)Abdolmaleki, Huang, Vezzani, Shahriari,
  Springenberg, Mishra, TB, Byravan, Bousmalis, Gyorgy,
  et~al.]{abdolmaleki2021multi}
Abbas Abdolmaleki, Sandy~H Huang, Giulia Vezzani, Bobak Shahriari, Jost~Tobias
  Springenberg, Shruti Mishra, Dhruva TB, Arunkumar Byravan, Konstantinos
  Bousmalis, Andras Gyorgy, et~al.
\newblock On multi-objective policy optimization as a tool for reinforcement
  learning.
\newblock \emph{arXiv preprint arXiv:2106.08199}, 2021.

\bibitem[{Adaptive Agent Team} et~al.(2023){Adaptive Agent Team}, Bauer,
  Baumli, Baveja, Behbahani, Bhoopchand, Bradley-Schmieg, Chang, Clay,
  Collister, Dasagi, Gonzalez, Gregor, Hughes, Kashem, Loks-Thompson, Openshaw,
  Parker-Holder, Pathak, Nieves, Rakicevic, Rockt{\"a}schel, Schroecker,
  Sygnowski, Tuyls, York, Zacherl, and Zhang]{Team2023HumanTimescaleAI}
{Adaptive Agent Team}, Jakob Bauer, Kate Baumli, Satinder Baveja, Feryal M.~P.
  Behbahani, Avishkar Bhoopchand, Nathalie Bradley-Schmieg, Michael~B. Chang,
  Natalie Clay, Adrian Collister, Vibhavari Dasagi, Lucy Gonzalez, Karol
  Gregor, Edward Hughes, Sheleem Kashem, Maria Loks-Thompson, Hannah Openshaw,
  Jack Parker-Holder, Shreyaan Pathak, Nicolas~Perez Nieves, Nemanja Rakicevic,
  Tim Rockt{\"a}schel, Yannick Schroecker, Jakub Sygnowski, Karl Tuyls, Sarah
  York, Alexander Zacherl, and Lei~M. Zhang.
\newblock Human-timescale adaptation in an open-ended task space.
\newblock \emph{ArXiv}, abs/2301.07608, 2023.

\bibitem[Agarwal et~al.(2019)Agarwal, Liang, Schuurmans, and
  Norouzi]{Agarwal2019LearningTG}
Rishabh Agarwal, Chen Liang, Dale Schuurmans, and Mohammad Norouzi.
\newblock Learning to generalize from sparse and underspecified rewards.
\newblock \emph{ArXiv}, abs/1902.07198, 2019.

\bibitem[Agarwal et~al.(2022)Agarwal, Schwarzer, Castro, Courville, and
  Bellemare]{Agarwal2022ReincarnatingRL}
Rishabh Agarwal, Max Schwarzer, Pablo~Samuel Castro, Aaron~C Courville, and
  Marc Bellemare.
\newblock Reincarnating reinforcement learning: Reusing prior computation to
  accelerate progress.
\newblock \emph{Advances in Neural Information Processing Systems},
  35:\penalty0 28955--28971, 2022.

\bibitem[Ahmed et~al.(2021)Ahmed, Tr{\"a}uble, Goyal, Neitz, W{\"u}thrich,
  Bengio, Sch{\"o}lkopf, and Bauer]{ahmed2021causalworld}
Ossama Ahmed, Frederik Tr{\"a}uble, Anirudh Goyal, Alexander Neitz, Manuel
  W{\"u}thrich, Yoshua Bengio, Bernhard Sch{\"o}lkopf, and Stefan Bauer.
\newblock Causalworld: A robotic manipulation benchmark for causal structure
  and transfer learning.
\newblock In \emph{International Conference on Learning Representations}, 2021.

\bibitem[Ahn et~al.(2022)Ahn, Brohan, Brown, Chebotar, Cortes, David, Finn,
  Gopalakrishnan, Hausman, Herzog, et~al.]{ahn2022can}
Michael Ahn, Anthony Brohan, Noah Brown, Yevgen Chebotar, Omar Cortes, Byron
  David, Chelsea Finn, Keerthana Gopalakrishnan, Karol Hausman, Alex Herzog,
  et~al.
\newblock Do as i can, not as i say: Grounding language in robotic affordances.
\newblock \emph{arXiv preprint arXiv:2204.01691}, 2022.

\bibitem[Ainsworth et~al.(2022)Ainsworth, Hayase, and
  Srinivasa]{ainsworth2022git}
Samuel~K Ainsworth, Jonathan Hayase, and Siddhartha Srinivasa.
\newblock Git re-basin: Merging models modulo permutation symmetries.
\newblock \emph{arXiv preprint arXiv:2209.04836}, 2022.

\bibitem[Aky{\"u}rek et~al.(2022)Aky{\"u}rek, Schuurmans, Andreas, Ma, and
  Zhou]{akyurek2022learning}
Ekin Aky{\"u}rek, Dale Schuurmans, Jacob Andreas, Tengyu Ma, and Denny Zhou.
\newblock What learning algorithm is in-context learning? investigations with
  linear models.
\newblock \emph{arXiv preprint arXiv:2211.15661}, 2022.

\bibitem[Alayrac et~al.(2022)Alayrac, Donahue, Luc, Miech, Barr, Hasson, Lenc,
  Mensch, Millican, Reynolds, Ring, Rutherford, Cabi, Han, Gong, Samangooei,
  Monteiro, Menick, Borgeaud, Brock, Nematzadeh, Sharifzadeh, Binkowski,
  Barreira, Vinyals, Zisserman, and Simonyan]{Alayrac2022FlamingoAV}
Jean-Baptiste Alayrac, Jeff Donahue, Pauline Luc, Antoine Miech, Iain Barr,
  Yana Hasson, Karel Lenc, Arthur Mensch, Katie Millican, Malcolm Reynolds,
  Roman Ring, Eliza Rutherford, Serkan Cabi, Tengda Han, Zhitao Gong, Sina
  Samangooei, Marianne Monteiro, Jacob Menick, Sebastian Borgeaud, Andy Brock,
  Aida Nematzadeh, Sahand Sharifzadeh, Mikolaj Binkowski, Ricardo Barreira,
  Oriol Vinyals, Andrew Zisserman, and Karen Simonyan.
\newblock Flamingo: a visual language model for few-shot learning.
\newblock \emph{ArXiv}, abs/2204.14198, 2022.

\bibitem[Andreas et~al.(2017)Andreas, Klein, and Levine]{andreas2017modular}
Jacob Andreas, Dan Klein, and Sergey Levine.
\newblock Modular multitask reinforcement learning with policy sketches.
\newblock In \emph{International Conference on Machine Learning}, pages
  166--175. PMLR, 2017.

\bibitem[Andrychowicz et~al.(2016)Andrychowicz, Denil, Colmenarejo, Hoffman,
  Pfau, Schaul, and de~Freitas]{l2l}
Marcin Andrychowicz, Misha Denil, Sergio~Gomez Colmenarejo, Matthew~W. Hoffman,
  David Pfau, Tom Schaul, and Nando de~Freitas.
\newblock Learning to learn by gradient descent by gradient descent.
\newblock In \emph{NeurIPS}, pages 3981--3989, 2016.

\bibitem[Andrychowicz et~al.(2017)Andrychowicz, Wolski, Ray, Schneider, Fong,
  Welinder, McGrew, Tobin, Abbeel, and Zaremba]{andrychowicz2017hindsight}
Marcin Andrychowicz, Filip Wolski, Alex Ray, Jonas Schneider, Rachel Fong,
  Peter Welinder, Bob McGrew, Josh Tobin, Pieter Abbeel, and Wojciech Zaremba.
\newblock Hindsight experience replay.
\newblock \emph{arXiv preprint arXiv:1707.01495}, 2017.

\bibitem[Arnekvist et~al.(2019)Arnekvist, Kragic, and Stork]{arnekvist2019vpe}
Isac Arnekvist, Danica Kragic, and Johannes~A Stork.
\newblock Vpe: Variational policy embedding for transfer reinforcement
  learning.
\newblock In \emph{2019 International Conference on Robotics and Automation
  (ICRA)}, pages 36--42. IEEE, 2019.

\bibitem[Asada et~al.(1994)Asada, Noda, Tawaratsumida, and
  Hosoda]{asada1994vision}
Minoru Asada, Shoichi Noda, Sukoya Tawaratsumida, and Koh Hosoda.
\newblock Vision-based behavior acquisition for a shooting robot by using a
  reinforcement learning.
\newblock In \emph{Proc. of IAPR/IEEE Workshop on Visual Behaviors}, pages
  112--118, 1994.

\bibitem[Ash and Adams(2020)]{ash2020warm}
Jordan Ash and Ryan~P Adams.
\newblock On warm-starting neural network training.
\newblock \emph{Advances in Neural Information Processing Systems}, 33, 2020.

\bibitem[Aytar et~al.(2018)Aytar, Pfaff, Budden, Paine, Wang, and
  De~Freitas]{aytar2018playing}
Yusuf Aytar, Tobias Pfaff, David Budden, Thomas Paine, Ziyu Wang, and Nando
  De~Freitas.
\newblock Playing hard exploration games by watching youtube.
\newblock \emph{Advances in neural information processing systems}, 31, 2018.

\bibitem[Bacon et~al.(2017)Bacon, Harb, and Precup]{bacon2017option}
Pierre-Luc Bacon, Jean Harb, and Doina Precup.
\newblock The option-critic architecture.
\newblock In \emph{Proceedings of the AAAI conference on artificial
  intelligence}, volume~31, 2017.

\bibitem[Bahdanau et~al.(2019)Bahdanau, Hill, Leike, Hughes, Hosseini, Kohli,
  and Grefenstette]{Bahdanau2019LearningTU}
Dzmitry Bahdanau, Felix Hill, Jan Leike, Edward Hughes, Seyedarian Hosseini,
  Pushmeet Kohli, and Edward Grefenstette.
\newblock Learning to understand goal specifications by modelling reward.
\newblock In \emph{ICLR}, 2019.

\bibitem[Bahl et~al.(2022)Bahl, Gupta, and Pathak]{bahl2022human}
Shikhar Bahl, Abhinav Gupta, and Deepak Pathak.
\newblock Human-to-robot imitation in the wild.
\newblock \emph{arXiv preprint arXiv:2207.09450}, 2022.

\bibitem[Ball et~al.(2021{\natexlab{a}})Ball, Lu, Parker-Holder, and
  Roberts]{ball2021augmented}
Philip~J Ball, Cong Lu, Jack Parker-Holder, and Stephen Roberts.
\newblock Augmented world models facilitate zero-shot dynamics generalization
  from a single offline environment.
\newblock In \emph{International Conference on Machine Learning}, pages
  619--629. PMLR, 2021{\natexlab{a}}.

\bibitem[Ball et~al.(2021{\natexlab{b}})Ball, Lu, Parker-Holder, and
  Roberts]{Ball2021AugmentedWM}
Philip~J. Ball, Cong Lu, Jack Parker-Holder, and Stephen~J. Roberts.
\newblock Augmented world models facilitate zero-shot dynamics generalization
  from a single offline environment.
\newblock \emph{ArXiv}, abs/2104.05632, 2021{\natexlab{b}}.

\bibitem[Ball et~al.(2023)Ball, Smith, Kostrikov, and
  Levine]{ball2023efficient}
Philip~J Ball, Laura Smith, Ilya Kostrikov, and Sergey Levine.
\newblock Efficient online reinforcement learning with offline data.
\newblock \emph{arXiv preprint arXiv:2302.02948}, 2023.

\bibitem[Bapst et~al.(2019)Bapst, Sanchez-Gonzalez, Doersch, Stachenfeld,
  Kohli, Battaglia, and Hamrick]{bapst2019structured}
Victor Bapst, Alvaro Sanchez-Gonzalez, Carl Doersch, Kimberly Stachenfeld,
  Pushmeet Kohli, Peter Battaglia, and Jessica Hamrick.
\newblock Structured agents for physical construction.
\newblock In \emph{International conference on machine learning}, pages
  464--474. PMLR, 2019.

\bibitem[Barreto et~al.(2016)Barreto, Dabney, Munos, Hunt, Schaul, Van~Hasselt,
  and Silver]{barreto2016successor}
Andr{\'e} Barreto, Will Dabney, R{\'e}mi Munos, Jonathan~J Hunt, Tom Schaul,
  Hado Van~Hasselt, and David Silver.
\newblock Successor features for transfer in reinforcement learning.
\newblock \emph{arXiv preprint arXiv:1606.05312}, 2016.

\bibitem[Barreto et~al.(2018)Barreto, Borsa, Quan, Schaul, Silver, Hessel,
  Mankowitz, Zidek, and Munos]{barreto2018transfer}
Andre Barreto, Diana Borsa, John Quan, Tom Schaul, David Silver, Matteo Hessel,
  Daniel Mankowitz, Augustin Zidek, and Remi Munos.
\newblock Transfer in deep reinforcement learning using successor features and
  generalised policy improvement.
\newblock In \emph{International Conference on Machine Learning}, pages
  501--510. PMLR, 2018.

\bibitem[Barreto et~al.(2019)Barreto, Borsa, Hou, Comanici, Ayg{\"u}n, Hamel,
  Toyama, Mourad, Silver, Precup, et~al.]{barreto2019option}
Andr{\'e} Barreto, Diana Borsa, Shaobo Hou, Gheorghe Comanici, Eser Ayg{\"u}n,
  Philippe Hamel, Daniel Toyama, Shibl Mourad, David Silver, Doina Precup,
  et~al.
\newblock The option keyboard: Combining skills in reinforcement learning.
\newblock \emph{Advances in Neural Information Processing Systems}, 32, 2019.

\bibitem[Bauer et~al.(2021)Bauer, Widmaier, Wuthrich, Buchholz, Stark, Goyal,
  Steinbrenner, Akpo, Joshi, Berenz, Agrawal, Funk, Jesus, Peters, Watson,
  Chen, Srinivasan, Zhang, Zhang, Walter, Madan, Schaff, Maeda, Yoneda, Yarats,
  Allshire, Gordon, Bhattacharjee, Srinivasa, Garg, Sikchi, Wang, Yao, Yang,
  McCarthy, S{\'a}nchez, Wang, Bulens, McGuinness, O'Connor, Redmond, and
  Scholkopf]{Bauer2021RealRC}
Stefan Bauer, Felix Widmaier, Manuel Wuthrich, Annika Buchholz, Sebastian
  Stark, Anirudh Goyal, Thomas Steinbrenner, Joel~Bessekon Akpo, Shruti Joshi,
  Vincent Berenz, Vaibhav Agrawal, Niklas Funk, Julen Urain~De Jesus, Jan
  Peters, Joe Watson, Claire Chen, Krishna~Parasuram Srinivasan, Junwu Zhang,
  Jeffrey Zhang, Matthew Walter, Rishabh Madan, Charles~B. Schaff, Takahiro
  Maeda, Takuma Yoneda, Denis Yarats, Arthur Allshire, Ethan~K. Gordon,
  Tapomayukh Bhattacharjee, Siddhartha~S. Srinivasa, Animesh Garg, Harshit~S.
  Sikchi, Jilong Wang, Qingfeng Yao, Shuyu Yang, Robert McCarthy,
  Francisco~Rold{\'a}n S{\'a}nchez, Qiang Wang, David~Cordova Bulens, Kevin
  McGuinness, Noel~E. O'Connor, Stephen~James Redmond, and Bernhard Scholkopf.
\newblock Real robot challenge: A robotics competition in the cloud.
\newblock In \emph{NeurIPS}, 2021.

\bibitem[Beattie et~al.(2016)Beattie, Leibo, Teplyashin, Ward, Wainwright,
  K{\"u}ttler, Lefrancq, Green, Vald{\'e}s, Sadik, Schrittwieser, Anderson,
  York, Cant, Cain, Bolton, Gaffney, King, Hassabis, Legg, and
  Petersen]{Beattie2016DeepMindL}
Charlie Beattie, Joel~Z. Leibo, Denis Teplyashin, Tom Ward, Marcus Wainwright,
  Heinrich K{\"u}ttler, Andrew Lefrancq, Simon Green, V{\'i}ctor Vald{\'e}s,
  Amir Sadik, Julian Schrittwieser, Keith Anderson, Sarah York, Max Cant, Adam
  Cain, Adrian Bolton, Stephen Gaffney, Helen King, Demis Hassabis, Shane Legg,
  and Stig Petersen.
\newblock Deepmind lab.
\newblock \emph{ArXiv}, abs/1612.03801, 2016.

\bibitem[Bechtle et~al.(2021)Bechtle, Molchanov, Chebotar, Grefenstette,
  Righetti, Sukhatme, and Meier]{bechtle2021meta}
Sarah Bechtle, Artem Molchanov, Yevgen Chebotar, Edward Grefenstette, Ludovic
  Righetti, Gaurav Sukhatme, and Franziska Meier.
\newblock Meta learning via learned loss.
\newblock In \emph{2020 25th International Conference on Pattern Recognition
  (ICPR)}, pages 4161--4168. IEEE, 2021.

\bibitem[Belkhale et~al.(2021)Belkhale, Li, Kahn, McAllister, Calandra, and
  Levine]{belkhale2021model}
Suneel Belkhale, Rachel Li, Gregory Kahn, Rowan McAllister, Roberto Calandra,
  and Sergey Levine.
\newblock Model-based meta-reinforcement learning for flight with suspended
  payloads.
\newblock \emph{IEEE Robotics and Automation Letters}, 6\penalty0 (2):\penalty0
  1471--1478, 2021.

\bibitem[Bellemare et~al.(2013)Bellemare, Naddaf, Veness, and
  Bowling]{bellemare2013arcade}
Marc~G Bellemare, Yavar Naddaf, Joel Veness, and Michael Bowling.
\newblock The arcade learning environment: An evaluation platform for general
  agents.
\newblock \emph{Journal of Artificial Intelligence Research}, 47:\penalty0
  253--279, 2013.

\bibitem[Bengio(2012)]{bengio2012deep}
Yoshua Bengio.
\newblock Deep learning of representations for unsupervised and transfer
  learning.
\newblock In \emph{Proceedings of ICML workshop on unsupervised and transfer
  learning}, pages 17--36. JMLR Workshop and Conference Proceedings, 2012.

\bibitem[Bengio et~al.(2013)Bengio, Courville, and
  Vincent]{bengio2013representation}
Yoshua Bengio, Aaron Courville, and Pascal Vincent.
\newblock Representation learning: A review and new perspectives.
\newblock \emph{IEEE transactions on pattern analysis and machine
  intelligence}, 35\penalty0 (8):\penalty0 1798--1828, 2013.

\bibitem[Benjamins et~al.(2021)Benjamins, Eimer, Schubert, Biedenkapp,
  Rosenhahn, Hutter, and Lindauer]{BenEim2021a}
Carolin Benjamins, Theresa Eimer, Frederik Schubert, André Biedenkapp, Bodo
  Rosenhahn, Frank Hutter, and Marius Lindauer.
\newblock Carl: A benchmark for contextual and adaptive reinforcement learning.
\newblock In \emph{NeurIPS 2021 Workshop on Ecological Theory of Reinforcement
  Learning}, December 2021.

\bibitem[Bohez et~al.(2022)Bohez, Tunyasuvunakool, Brakel, Sadeghi,
  Hasenclever, Tassa, Parisotto, Humplik, Haarnoja, Hafner, Wulfmeier, Neunert,
  Moran, Siegel, Huber, Romano, Batchelor, Casarini, Merel, Hadsell, and
  Heess]{Bohez2022ImitateAR}
Steven Bohez, Saran Tunyasuvunakool, Philemon Brakel, Fereshteh Sadeghi,
  Leonard Hasenclever, Yuval Tassa, Emilio Parisotto, Jan Humplik, Tuomas
  Haarnoja, Roland Hafner, Markus Wulfmeier, Michael Neunert, Ben Moran, Noah
  Siegel, Andrea Huber, Francesco Romano, Nathan Batchelor, Federico Casarini,
  Josh Merel, Raia Hadsell, and Nicolas Manfred~Otto Heess.
\newblock Imitate and repurpose: Learning reusable robot movement skills from
  human and animal behaviors.
\newblock \emph{ArXiv}, abs/2203.17138, 2022.

\bibitem[Bommasani et~al.(2021)Bommasani, Hudson, Adeli, Altman, Arora, von
  Arx, Bernstein, Bohg, Bosselut, Brunskill,
  et~al.]{bommasani2021opportunities}
Rishi Bommasani, Drew~A Hudson, Ehsan Adeli, Russ Altman, Simran Arora, Sydney
  von Arx, Michael~S Bernstein, Jeannette Bohg, Antoine Bosselut, Emma
  Brunskill, et~al.
\newblock On the opportunities and risks of foundation models.
\newblock \emph{arXiv preprint arXiv:2108.07258}, 2021.

\bibitem[Borsa et~al.(2018)Borsa, Barreto, Quan, Mankowitz, Munos, van Hasselt,
  Silver, and Schaul]{borsa2018universal}
Diana Borsa, Andr{\'e} Barreto, John Quan, Daniel Mankowitz, R{\'e}mi Munos,
  Hado van Hasselt, David Silver, and Tom Schaul.
\newblock Universal successor features approximators.
\newblock \emph{arXiv preprint arXiv:1812.07626}, 2018.

\bibitem[Bousmalis et~al.(2018)Bousmalis, Irpan, Wohlhart, Bai, Kelcey,
  Kalakrishnan, Downs, Ibarz, Pastor, Konolige, et~al.]{bousmalis2018using}
Konstantinos Bousmalis, Alex Irpan, Paul Wohlhart, Yunfei Bai, Matthew Kelcey,
  Mrinal Kalakrishnan, Laura Downs, Julian Ibarz, Peter Pastor, Kurt Konolige,
  et~al.
\newblock Using simulation and domain adaptation to improve efficiency of deep
  robotic grasping.
\newblock In \emph{2018 IEEE international conference on robotics and
  automation (ICRA)}, pages 4243--4250. IEEE, 2018.

\bibitem[Bousmalis et~al.(2023)Bousmalis, Vezzani, Rao, Devin, Lee, Bauza,
  Davchev, Zhou, Gupta, Raju, Laurens, Fantacci, Dalibard, Zambelli, Martins,
  Pevceviciute, Blokzijl, Denil, Batchelor, Lampe, Parisotto, Żołna, Reed,
  Colmenarejo, Scholz, Abdolmaleki, Groth, Regli, Sushkov, Rothörl, Chen,
  Aytar, Barker, Ortiz, Riedmiller, Springenberg, Hadsell, Nori, and
  Heess]{bousmalis2023robocat}
Konstantinos Bousmalis, Giulia Vezzani, Dushyant Rao, Coline Devin, Alex~X.
  Lee, Maria Bauza, Todor Davchev, Yuxiang Zhou, Agrim Gupta, Akhil Raju,
  Antoine Laurens, Claudio Fantacci, Valentin Dalibard, Martina Zambelli,
  Murilo Martins, Rugile Pevceviciute, Michiel Blokzijl, Misha Denil, Nathan
  Batchelor, Thomas Lampe, Emilio Parisotto, Konrad Żołna, Scott Reed,
  Sergio~Gómez Colmenarejo, Jon Scholz, Abbas Abdolmaleki, Oliver Groth,
  Jean-Baptiste Regli, Oleg Sushkov, Tom Rothörl, José~Enrique Chen, Yusuf
  Aytar, Dave Barker, Joy Ortiz, Martin Riedmiller, Jost~Tobias Springenberg,
  Raia Hadsell, Francesco Nori, and Nicolas Heess.
\newblock Robocat: A self-improving foundation agent for robotic manipulation,
  2023.

\bibitem[Brockman et~al.(2016)Brockman, Cheung, Pettersson, Schneider,
  Schulman, Tang, and Zaremba]{openaigym}
Greg Brockman, Vicki Cheung, Ludwig Pettersson, Jonas Schneider, John Schulman,
  Jie Tang, and Wojciech Zaremba.
\newblock Openai gym, 2016.

\bibitem[Brohan et~al.(2022)Brohan, Brown, Carbajal, Chebotar, Dabis, Finn,
  Gopalakrishnan, Hausman, Herzog, Hsu, Ibarz, Ichter, Irpan, Jackson,
  Jesmonth, Joshi, Julian, Kalashnikov, Kuang, Leal, Lee, Levine, Lu, Malla,
  Manjunath, Mordatch, Nachum, Parada, Peralta, Perez, Pertsch, Quiambao, Rao,
  Ryoo, Salazar, Sanketi, Sayed, Singh, Sontakke, Stone, Tan, Tran, Vanhoucke,
  Vega, Vuong, Xia, Xiao, Xu, Xu, Yu, and Zitkovich]{Brohan2022RT1RT}
Anthony Brohan, Noah Brown, Justice Carbajal, Yevgen Chebotar, Joseph Dabis,
  Chelsea Finn, Keerthana Gopalakrishnan, Karol Hausman, Alexander Herzog,
  Jasmine Hsu, Julian Ibarz, Brian Ichter, Alex Irpan, Tomas Jackson, Sally
  Jesmonth, Nikhil~J. Joshi, Ryan~C. Julian, Dmitry Kalashnikov, Yuheng Kuang,
  Isabel Leal, Kuang-Huei Lee, Sergey Levine, Yao Lu, Utsav Malla, Deeksha
  Manjunath, Igor Mordatch, Ofir Nachum, Carolina Parada, Jodilyn Peralta,
  Emily Perez, Karl Pertsch, Jornell Quiambao, Kanishka Rao, Michael~S. Ryoo,
  Grecia Salazar, Pannag~R. Sanketi, Kevin Sayed, Jaspiar Singh, Sumedh~Anand
  Sontakke, Austin Stone, Clayton Tan, Huong Tran, Vincent Vanhoucke, Steve
  Vega, Quan~Ho Vuong, F.~Xia, Ted Xiao, Peng Xu, Sichun Xu, Tianhe Yu, and
  Brianna Zitkovich.
\newblock Rt-1: Robotics transformer for real-world control at scale.
\newblock \emph{ArXiv}, abs/2212.06817, 2022.

\bibitem[Brohan et~al.(2023)Brohan, Brown, Carbajal, Chebotar, Choromanski,
  Ding, Driess, Finn, Florence, Fu, Arenas, Gopalakrishnan, Han, Hausman,
  Herzog, Hsu, Ichter, Irpan, Joshi, Julian, Kalashnikov, Kuang, Leal, Levine,
  Michalewski, Mordatch, Pertsch, Rao, Reymann, Ryoo, Salazar, Sanketi,
  Sermanet, Singh, Singh, Soricut, Tran, Vanhoucke, Vuong, Wahid, Welker,
  Wohlhart, Xiao, Yu, and Zitkovich]{Brohan2023RT2VM}
Anthony Brohan, Noah Brown, Justice Carbajal, Yevgen Chebotar, Krzysztof
  Choromanski, Tianli Ding, Danny Driess, Chelsea Finn, Peter~R. Florence,
  Chuyuan Fu, Montse~Gonzalez Arenas, Keerthana Gopalakrishnan, Kehang Han,
  Karol Hausman, Alexander Herzog, Jasmine Hsu, Brian Ichter, Alex Irpan,
  Nikhil~J. Joshi, Ryan~C. Julian, Dmitry Kalashnikov, Yuheng Kuang, Isabel
  Leal, Sergey Levine, Henryk Michalewski, Igor Mordatch, Karl Pertsch,
  Kanishka Rao, Krista Reymann, Michael~S. Ryoo, Grecia Salazar, Pannag~R.
  Sanketi, Pierre Sermanet, Jaspiar Singh, Anika Singh, Radu Soricut, Huong
  Tran, Vincent Vanhoucke, Quan~Ho Vuong, Ayzaan Wahid, Stefan Welker, Paul
  Wohlhart, Ted Xiao, Tianhe Yu, and Brianna Zitkovich.
\newblock Rt-2: Vision-language-action models transfer web knowledge to robotic
  control.
\newblock \emph{ArXiv}, abs/2307.15818, 2023.
\newblock URL \url{https://api.semanticscholar.org/CorpusID:260293142}.

\bibitem[Brown et~al.(2020)Brown, Mann, Ryder, Subbiah, Kaplan, Dhariwal,
  Neelakantan, Shyam, Sastry, Askell, et~al.]{brown2020language}
Tom Brown, Benjamin Mann, Nick Ryder, Melanie Subbiah, Jared~D Kaplan, Prafulla
  Dhariwal, Arvind Neelakantan, Pranav Shyam, Girish Sastry, Amanda Askell,
  et~al.
\newblock Language models are few-shot learners.
\newblock \emph{Advances in neural information processing systems},
  33:\penalty0 1877--1901, 2020.

\bibitem[Brunskill and Li(2014)]{pmlr-v32-brunskill14}
Emma Brunskill and Lihong Li.
\newblock Pac-inspired option discovery in lifelong reinforcement learning.
\newblock In Eric~P. Xing and Tony Jebara, editors, \emph{Proceedings of the
  31st International Conference on Machine Learning}, volume~32 of
  \emph{Proceedings of Machine Learning Research}, pages 316--324, Bejing,
  China, 22--24 Jun 2014. PMLR.
\newblock URL \url{https://proceedings.mlr.press/v32/brunskill14.html}.

\bibitem[Brys et~al.(2015)Brys, Harutyunyan, Taylor, and
  Now{\'e}]{Brys2015PolicyTU}
T.~Brys, A.~Harutyunyan, Matthew~E. Taylor, and A.~Now{\'e}.
\newblock Policy transfer using reward shaping.
\newblock In \emph{AAMAS}, 2015.

\bibitem[Bucher et~al.(2020)Bucher, Schmeckpeper, Matni, and
  Daniilidis]{bucher2020adversarial}
Bernadette Bucher, Karl Schmeckpeper, Nikolai Matni, and Kostas Daniilidis.
\newblock An adversarial objective for scalable exploration.
\newblock \emph{arXiv preprint arXiv:2003.06082}, 2020.

\bibitem[Byravan et~al.(2017)Byravan, Leeb, Meier, and Fox]{byravan2017se3}
Arunkumar Byravan, Felix Leeb, Franziska Meier, and Dieter Fox.
\newblock Se3-pose-nets: Structured deep dynamics models for visuomotor
  planning and control.
\newblock \emph{arXiv preprint arXiv:1710.00489}, 2017.

\bibitem[Byravan et~al.(2020)Byravan, Springenberg, Abdolmaleki, Hafner,
  Neunert, Lampe, Siegel, Heess, and Riedmiller]{byravan2020imagined}
Arunkumar Byravan, Jost~Tobias Springenberg, Abbas Abdolmaleki, Roland Hafner,
  Michael Neunert, Thomas Lampe, Noah Siegel, Nicolas Heess, and Martin
  Riedmiller.
\newblock Imagined value gradients: Model-based policy optimization with
  tranferable latent dynamics models.
\newblock In \emph{Conference on Robot Learning}, pages 566--589. PMLR, 2020.

\bibitem[Byravan et~al.(2021)Byravan, Hasenclever, Trochim, Mirza, Ialongo,
  Tassa, Springenberg, Abdolmaleki, Heess, Merel, and
  Riedmiller]{Byravan2021EvaluatingMP}
Arunkumar Byravan, Leonard Hasenclever, Piotr Trochim, M.~Berk Mirza,
  Alessandro~Davide Ialongo, Yuval Tassa, Jost~Tobias Springenberg, Abbas
  Abdolmaleki, Nicolas Manfred~Otto Heess, Josh Merel, and Martin~A.
  Riedmiller.
\newblock Evaluating model-based planning and planner amortization for
  continuous control.
\newblock \emph{ArXiv}, abs/2110.03363, 2021.

\bibitem[Cabi et~al.(2019)Cabi, Colmenarejo, Novikov, Konyushkova, Reed, Jeong,
  Zolna, Aytar, Budden, Vecerik, et~al.]{cabi2019scaling}
Serkan Cabi, Sergio~G{\'o}mez Colmenarejo, Alexander Novikov, Ksenia
  Konyushkova, Scott Reed, Rae Jeong, Konrad Zolna, Yusuf Aytar, David Budden,
  Mel Vecerik, et~al.
\newblock Scaling data-driven robotics with reward sketching and batch
  reinforcement learning.
\newblock \emph{arXiv preprint arXiv:1909.12200}, 2019.

\bibitem[Campos et~al.(2021)Campos, Sprechmann, Hansen, Barreto, Kapturowski,
  Vitvitskyi, Badia, and Blundell]{Campos2021BeyondFT}
V{\'\i}ctor Campos, Pablo Sprechmann, Steven Hansen, Andre Barreto, Steven
  Kapturowski, Alex Vitvitskyi, Adria~Puigdomenech Badia, and Charles Blundell.
\newblock Beyond fine-tuning: Transferring behavior in reinforcement learning.
\newblock \emph{arXiv preprint arXiv:2102.13515}, 2021.

\bibitem[Caruana(1997)]{caruana1997multitask}
Rich Caruana.
\newblock Multitask learning.
\newblock \emph{Machine learning}, 28\penalty0 (1):\penalty0 41--75, 1997.

\bibitem[Chapman and Kaelbling(1991)]{chapman1991input}
David Chapman and Leslie~Pack Kaelbling.
\newblock Input generalization in delayed reinforcement learning: An algorithm
  and performance comparisons.
\newblock In \emph{Ijcai}, volume~91, pages 726--731, 1991.

\bibitem[Chapman(2015)]{chapman2015complexity}
Kelly Chapman.
\newblock \emph{Complexity and Creative Capacity: Rethinking knowledge
  transfer, adaptive management and wicked environmental problems}.
\newblock Routledge, 2015.

\bibitem[Chebotar et~al.(2021)Chebotar, Hausman, Lu, Xiao, Kalashnikov, Varley,
  Irpan, Eysenbach, Julian, Finn, et~al.]{chebotar2021actionable}
Yevgen Chebotar, Karol Hausman, Yao Lu, Ted Xiao, Dmitry Kalashnikov, Jake
  Varley, Alex Irpan, Benjamin Eysenbach, Ryan Julian, Chelsea Finn, et~al.
\newblock Actionable models: Unsupervised offline reinforcement learning of
  robotic skills.
\newblock \emph{arXiv preprint arXiv:2104.07749}, 2021.

\bibitem[Chen et~al.(2021{\natexlab{a}})Chen, Nair, and Finn]{chen2021learning}
Annie~S Chen, Suraj Nair, and Chelsea Finn.
\newblock Learning generalizable robotic reward functions from" in-the-wild"
  human videos.
\newblock \emph{arXiv preprint arXiv:2103.16817}, 2021{\natexlab{a}}.

\bibitem[Chen et~al.(2021{\natexlab{b}})Chen, Lu, Rajeswaran, Lee, Grover,
  Laskin, Abbeel, Srinivas, and Mordatch]{Chen2021DecisionTR}
Lili Chen, Kevin Lu, Aravind Rajeswaran, Kimin Lee, Aditya Grover, Michael
  Laskin, P.~Abbeel, A.~Srinivas, and Igor Mordatch.
\newblock Decision transformer: Reinforcement learning via sequence modeling.
\newblock In \emph{Neural Information Processing Systems}, 2021{\natexlab{b}}.

\bibitem[Chen et~al.(2022)Chen, Xu, and Agrawal]{chen2022system}
Tao Chen, Jie Xu, and Pulkit Agrawal.
\newblock A system for general in-hand object re-orientation.
\newblock In \emph{Conference on Robot Learning}, pages 297--307. PMLR, 2022.

\bibitem[Chowdhery et~al.(2022)Chowdhery, Narang, Devlin, Bosma, Mishra,
  Roberts, Barham, Chung, Sutton, Gehrmann, Schuh, Shi, Tsvyashchenko, Maynez,
  Rao, Barnes, Tay, Shazeer, Prabhakaran, Reif, Du, Hutchinson, Pope, Bradbury,
  Austin, Isard, Gur-Ari, Yin, Duke, Levskaya, Ghemawat, Dev, Michalewski,
  Garc{\'i}a, Misra, Robinson, Fedus, Zhou, Ippolito, Luan, Lim, Zoph,
  Spiridonov, Sepassi, Dohan, Agrawal, Omernick, Dai, Pillai, Pellat,
  Lewkowycz, Moreira, Child, Polozov, Lee, Zhou, Wang, Saeta, D{\'i}az, Firat,
  Catasta, Wei, Meier-Hellstern, Eck, Dean, Petrov, and
  Fiedel]{Chowdhery2022PaLMSL}
Aakanksha Chowdhery, Sharan Narang, Jacob Devlin, Maarten Bosma, Gaurav Mishra,
  Adam Roberts, Paul Barham, Hyung~Won Chung, Charles Sutton, Sebastian
  Gehrmann, Parker Schuh, Kensen Shi, Sasha Tsvyashchenko, Joshua Maynez,
  Abhishek Rao, Parker Barnes, Yi~Tay, Noam~M. Shazeer, Vinodkumar Prabhakaran,
  Emily Reif, Nan Du, Benton~C. Hutchinson, Reiner Pope, James Bradbury, Jacob
  Austin, Michael Isard, Guy Gur-Ari, Pengcheng Yin, Toju Duke, Anselm
  Levskaya, Sanjay Ghemawat, Sunipa Dev, Henryk Michalewski, Xavier Garc{\'i}a,
  Vedant Misra, Kevin Robinson, Liam Fedus, Denny Zhou, Daphne Ippolito, David
  Luan, Hyeontaek Lim, Barret Zoph, Alexander Spiridonov, Ryan Sepassi, David
  Dohan, Shivani Agrawal, Mark Omernick, Andrew~M. Dai,
  Thanumalayan~Sankaranarayana Pillai, Marie Pellat, Aitor Lewkowycz, Erica
  Moreira, Rewon Child, Oleksandr Polozov, Katherine Lee, Zongwei Zhou, Xuezhi
  Wang, Brennan Saeta, Mark D{\'i}az, Orhan Firat, Michele Catasta, Jason Wei,
  Kathleen~S. Meier-Hellstern, Douglas Eck, Jeff Dean, Slav Petrov, and Noah
  Fiedel.
\newblock Palm: Scaling language modeling with pathways.
\newblock \emph{ArXiv}, abs/2204.02311, 2022.

\bibitem[Christiano et~al.(2016)Christiano, Shah, Mordatch, Schneider,
  Blackwell, Tobin, Abbeel, and Zaremba]{christiano2016transfer}
Paul Christiano, Zain Shah, Igor Mordatch, Jonas Schneider, Trevor Blackwell,
  Joshua Tobin, Pieter Abbeel, and Wojciech Zaremba.
\newblock Transfer from simulation to real world through learning deep inverse
  dynamics model, 2016.

\bibitem[Clavera et~al.(2018)Clavera, Rothfuss, Schulman, Fujita, Asfour, and
  Abbeel]{clavera2018model}
Ignasi Clavera, Jonas Rothfuss, John Schulman, Yasuhiro Fujita, Tamim Asfour,
  and Pieter Abbeel.
\newblock Model-based reinforcement learning via meta-policy optimization.
\newblock In \emph{Conference on Robot Learning}, pages 617--629. PMLR, 2018.

\bibitem[Cobbe et~al.(2020)Cobbe, Hesse, Hilton, and
  Schulman]{Cobbe2020LeveragingPG}
Karl Cobbe, Christopher Hesse, Jacob Hilton, and John Schulman.
\newblock Leveraging procedural generation to benchmark reinforcement learning.
\newblock \emph{ArXiv}, abs/1912.01588, 2020.

\bibitem[Cordts et~al.(2016)Cordts, Omran, Ramos, Rehfeld, Enzweiler, Benenson,
  Franke, Roth, and Schiele]{Cordts2016TheCD}
Marius Cordts, Mohamed Omran, Sebastian Ramos, Timo Rehfeld, Markus Enzweiler,
  Rodrigo Benenson, Uwe Franke, Stefan Roth, and Bernt Schiele.
\newblock The cityscapes dataset for semantic urban scene understanding.
\newblock \emph{2016 IEEE Conference on Computer Vision and Pattern Recognition
  (CVPR)}, pages 3213--3223, 2016.

\bibitem[Crawshaw(2020)]{crawshaw2020multi}
Michael Crawshaw.
\newblock Multi-task learning with deep neural networks: A survey.
\newblock \emph{arXiv preprint arXiv:2009.09796}, 2020.

\bibitem[Cruz~Jr et~al.(2017)Cruz~Jr, Du, and Taylor]{cruz2017pre}
Gabriel~V Cruz~Jr, Yunshu Du, and Matthew~E Taylor.
\newblock Pre-training neural networks with human demonstrations for deep
  reinforcement learning.
\newblock \emph{arXiv preprint arXiv:1709.04083}, 2017.

\bibitem[Cui et~al.(2022)Cui, Niekum, Gupta, Kumar, and Rajeswaran]{cui2022can}
Yuchen Cui, Scott Niekum, Abhinav Gupta, Vikash Kumar, and Aravind Rajeswaran.
\newblock Can foundation models perform zero-shot task specification for robot
  manipulation?
\newblock In \emph{Learning for Dynamics and Control Conference}, pages
  893--905. PMLR, 2022.

\bibitem[Czarnecki et~al.(2019)Czarnecki, Pascanu, Osindero, Jayakumar,
  Swirszcz, and Jaderberg]{Czarnecki2019DistillingPD}
Wojciech~M. Czarnecki, Razvan Pascanu, Simon Osindero, Siddhant~M. Jayakumar,
  Grzegorz Swirszcz, and Max Jaderberg.
\newblock Distilling policy distillation.
\newblock \emph{ArXiv}, abs/1902.02186, 2019.

\bibitem[Dai et~al.(2022)Dai, Sun, Dong, Hao, Sui, and Wei]{dai2022can}
Damai Dai, Yutao Sun, Li~Dong, Yaru Hao, Zhifang Sui, and Furu Wei.
\newblock Why can gpt learn in-context? language models secretly perform
  gradient descent as meta optimizers.
\newblock \emph{arXiv preprint arXiv:2212.10559}, 2022.

\bibitem[Daniel et~al.(2012)Daniel, Neumann, and
  Peters]{daniel2012hierarchical}
Christian Daniel, Gerhard Neumann, and Jan Peters.
\newblock Hierarchical relative entropy policy search.
\newblock In \emph{Artificial Intelligence and Statistics}, pages 273--281.
  PMLR, 2012.

\bibitem[Daniel et~al.(2015)Daniel, Kroemer, Viering, Metz, and
  Peters]{ARL2015Daniel}
Christian Daniel, Oliver Kroemer, Malte Viering, Jan Metz, and Jan Peters.
\newblock Active reward learning with a novel acquisition function.
\newblock \emph{Autonomous Robots}, 39\penalty0 (3):\penalty0 389 -- 405,
  October 2015.

\bibitem[Dasari et~al.(2019)Dasari, Ebert, Tian, Nair, Bucher, Schmeckpeper,
  Singh, Levine, and Finn]{dasari2019robonet}
Sudeep Dasari, Frederik Ebert, Stephen Tian, Suraj Nair, Bernadette Bucher,
  Karl Schmeckpeper, Siddharth Singh, Sergey Levine, and Chelsea Finn.
\newblock Robonet: Large-scale multi-robot learning.
\newblock \emph{arXiv preprint arXiv:1910.11215}, 2019.

\bibitem[Davchev et~al.(2021)Davchev, Sushkov, Regli, Schaal, Aytar, Wulfmeier,
  and Scholz]{Davchev2021WishYW}
Todor Davchev, Oleg~O. Sushkov, Jean-Baptiste Regli, Stefan Schaal, Yusuf
  Aytar, Markus Wulfmeier, and Jonathan Scholz.
\newblock Wish you were here: Hindsight goal selection for long-horizon
  dexterous manipulation.
\newblock \emph{ArXiv}, abs/2112.00597, 2021.

\bibitem[Davchev et~al.(2022)Davchev, Luck, Burke, Meier, Schaal, and
  Ramamoorthy]{davchev2022residual}
Todor Davchev, Kevin~Sebastian Luck, Michael Burke, Franziska Meier, Stefan
  Schaal, and Subramanian Ramamoorthy.
\newblock Residual learning from demonstration: Adapting dmps for contact-rich
  manipulation.
\newblock \emph{IEEE Robotics and Automation Letters}, 7\penalty0 (2):\penalty0
  4488--4495, 2022.

\bibitem[Dayan(1993)]{dayan1993improving}
Peter Dayan.
\newblock Improving generalization for temporal difference learning: The
  successor representation.
\newblock \emph{Neural Computation}, 5\penalty0 (4):\penalty0 613--624, 1993.

\bibitem[Dayan and Hinton(1993)]{dayan1993feudal}
Peter Dayan and Geoffrey~E Hinton.
\newblock Feudal reinforcement learning.
\newblock In \emph{Advances in Neural Information Processing Systems}, 1993.

\bibitem[Degrave et~al.(2022)Degrave, Felici, Buchli, Neunert, Tracey,
  Carpanese, Ewalds, Hafner, Abdolmaleki, de~Las~Casas,
  et~al.]{degrave2022magnetic}
Jonas Degrave, Federico Felici, Jonas Buchli, Michael Neunert, Brendan Tracey,
  Francesco Carpanese, Timo Ewalds, Roland Hafner, Abbas Abdolmaleki, Diego
  de~Las~Casas, et~al.
\newblock Magnetic control of tokamak plasmas through deep reinforcement
  learning.
\newblock \emph{Nature}, 602\penalty0 (7897):\penalty0 414--419, 2022.

\bibitem[Devlin et~al.(2018)Devlin, Chang, Lee, and Toutanova]{devlin2018bert}
Jacob Devlin, Ming-Wei Chang, Kenton Lee, and Kristina Toutanova.
\newblock Bert: Pre-training of deep bidirectional transformers for language
  understanding.
\newblock \emph{arXiv preprint arXiv:1810.04805}, 2018.

\bibitem[Di~Palo et~al.(2023)Di~Palo, Byravan, Hasenclever, Wulfmeier, Heess,
  and Riedmiller]{di2023towards}
Norman Di~Palo, Arunkumar Byravan, Leonard Hasenclever, Markus Wulfmeier,
  Nicolas Heess, and Martin Riedmiller.
\newblock Towards a unified agent with foundation models.
\newblock In \emph{Workshop on Reincarnating Reinforcement Learning at ICLR
  2023}, 2023.

\bibitem[Dittadi et~al.(2021)Dittadi, Tr{\"a}uble, W{\"u}thrich, Widmaier,
  Gehler, Winther, Locatello, Bachem, Sch{\"o}lkopf, and
  Bauer]{Dittadi2021TheRO}
Andrea Dittadi, Frederik Tr{\"a}uble, Manuel W{\"u}thrich, Felix Widmaier,
  Peter Gehler, Ole Winther, Francesco Locatello, Olivier Bachem, Bernhard
  Sch{\"o}lkopf, and Stefan Bauer.
\newblock The role of pretrained representations for the ood generalization of
  reinforcement learning agents.
\newblock \emph{arXiv preprint arXiv:2107.05686}, 2021.

\bibitem[Dorfman and Tamar(2020)]{Dorfman2020OfflineMR}
Ron Dorfman and Aviv Tamar.
\newblock Offline meta reinforcement learning.
\newblock \emph{ArXiv}, abs/2008.02598, 2020.

\bibitem[Driess et~al.(2023)Driess, Xia, Sajjadi, Lynch, Chowdhery, Ichter,
  Wahid, Tompson, Vuong, Yu, et~al.]{driess2023palm}
Danny Driess, Fei Xia, Mehdi~SM Sajjadi, Corey Lynch, Aakanksha Chowdhery,
  Brian Ichter, Ayzaan Wahid, Jonathan Tompson, Quan Vuong, Tianhe Yu, et~al.
\newblock Palm-e: An embodied multimodal language model.
\newblock \emph{arXiv preprint arXiv:2303.03378}, 2023.

\bibitem[Du et~al.(2023)Du, Nair, Sadigh, and Finn]{Du2023BehaviorRF}
Maximilian Du, Suraj Nair, Dorsa Sadigh, and Chelsea Finn.
\newblock Behavior retrieval: Few-shot imitation learning by querying unlabeled
  datasets.
\newblock \emph{ArXiv}, abs/2304.08742, 2023.

\bibitem[Duan et~al.(2016)Duan, Schulman, Chen, Bartlett, Sutskever, and
  Abbeel]{duan2016rl2}
Yan Duan, John Schulman, Xi~Chen, Peter~L. Bartlett, Ilya Sutskever, and Pieter
  Abbeel.
\newblock Rl$^2$: Fast reinforcement learning via slow reinforcement learning,
  2016.

\bibitem[Duan et~al.(2017)Duan, Andrychowicz, Stadie, Ho, Schneider, Sutskever,
  Abbeel, and Zaremba]{duan2017one}
Yan Duan, Marcin Andrychowicz, Bradly~C Stadie, Jonathan Ho, Jonas Schneider,
  Ilya Sutskever, Pieter Abbeel, and Wojciech Zaremba.
\newblock One-shot imitation learning.
\newblock \emph{arXiv preprint arXiv:1703.07326}, 2017.

\bibitem[Ebert et~al.(2018)Ebert, Finn, Dasari, Xie, Lee, and
  Levine]{ebert2018visual}
Frederik Ebert, Chelsea Finn, Sudeep Dasari, Annie Xie, Alex Lee, and Sergey
  Levine.
\newblock Visual foresight: Model-based deep reinforcement learning for
  vision-based robotic control.
\newblock \emph{arXiv preprint arXiv:1812.00568}, 2018.

\bibitem[Ebert et~al.(2022)Ebert, Yang, Schmeckpeper, Bucher, Georgakis,
  Daniilidis, Finn, and Levine]{Ebert2022BridgeDB}
Frederik Ebert, Yanlai Yang, Karl Schmeckpeper, Bernadette Bucher, Georgios
  Georgakis, Kostas Daniilidis, Chelsea Finn, and Sergey Levine.
\newblock Bridge data: Boosting generalization of robotic skills with
  cross-domain datasets.
\newblock \emph{ArXiv}, abs/2109.13396, 2022.

\bibitem[Escontrela et~al.(2023)Escontrela, Adeniji, Yan, Jain, Peng, Goldberg,
  Lee, Hafner, and Abbeel]{escontrela2023video}
Alejandro Escontrela, Ademi Adeniji, Wilson Yan, Ajay Jain, Xue~Bin Peng, Ken
  Goldberg, Youngwoon Lee, Danijar Hafner, and Pieter Abbeel.
\newblock Video prediction models as rewards for reinforcement learning.
\newblock \emph{arXiv preprint arXiv:2305.14343}, 2023.

\bibitem[Evans et~al.(2022)Evans, Thankaraj, and Pinto]{evans2022context}
Ben Evans, Abitha Thankaraj, and Lerrel Pinto.
\newblock Context is everything: Implicit identification for dynamics
  adaptation.
\newblock \emph{arXiv preprint arXiv:2203.05549}, 2022.

\bibitem[Eysenbach et~al.(2020)Eysenbach, Geng, Levine, and
  Salakhutdinov]{eysenbach2020rewriting}
Ben Eysenbach, Xinyang Geng, Sergey Levine, and Russ~R Salakhutdinov.
\newblock Rewriting history with inverse rl: Hindsight inference for policy
  improvement.
\newblock \emph{Advances in neural information processing systems},
  33:\penalty0 14783--14795, 2020.

\bibitem[Eysenbach et~al.(2019)Eysenbach, Gupta, Ibarz, and
  Levine]{Eysenbach2019DiversityIA}
Benjamin Eysenbach, Abhishek Gupta, Julian Ibarz, and Sergey Levine.
\newblock Diversity is all you need: Learning skills without a reward function.
\newblock \emph{ArXiv}, abs/1802.06070, 2019.

\bibitem[Fakoor et~al.(2019)Fakoor, Chaudhari, Soatto, and
  Smola]{fakoor2019meta}
Rasool Fakoor, Pratik Chaudhari, Stefano Soatto, and Alexander~J Smola.
\newblock Meta-q-learning.
\newblock \emph{arXiv preprint arXiv:1910.00125}, 2019.

\bibitem[Fallah et~al.(2021)Fallah, Mokhtari, and
  Ozdaglar]{fallah2021generalization}
Alireza Fallah, Aryan Mokhtari, and Asuman Ozdaglar.
\newblock Generalization of model-agnostic meta-learning algorithms: Recurring
  and unseen tasks.
\newblock \emph{Advances in Neural Information Processing Systems},
  34:\penalty0 5469--5480, 2021.

\bibitem[Fan et~al.(2022)Fan, Wang, Jiang, Mandlekar, Yang, Zhu, Tang, Huang,
  Zhu, and Anandkumar]{fan2022minedojo}
Linxi Fan, Guanzhi Wang, Yunfan Jiang, Ajay Mandlekar, Yuncong Yang, Haoyi Zhu,
  Andrew Tang, De-An Huang, Yuke Zhu, and Anima Anandkumar.
\newblock Minedojo: Building open-ended embodied agents with internet-scale
  knowledge.
\newblock \emph{arXiv preprint arXiv:2206.08853}, 2022.

\bibitem[Finlayson(2020)]{finlayson2020learning}
Samuel~G Finlayson.
\newblock \emph{Learning Inductive Representations of Biomedical Data}.
\newblock PhD thesis, Harvard University, 2020.

\bibitem[Finn and Levine(2017)]{finn2017deep}
Chelsea Finn and Sergey Levine.
\newblock Deep visual foresight for planning robot motion.
\newblock In \emph{2017 IEEE International Conference on Robotics and
  Automation (ICRA)}, pages 2786--2793. IEEE, 2017.

\bibitem[Finn et~al.(2016)Finn, Levine, and Abbeel]{Finn2016GuidedCL}
Chelsea Finn, Sergey Levine, and P.~Abbeel.
\newblock Guided cost learning: Deep inverse optimal control via policy
  optimization.
\newblock \emph{ArXiv}, abs/1603.00448, 2016.

\bibitem[Finn et~al.(2017)Finn, Abbeel, and Levine]{finn2017model}
Chelsea Finn, Pieter Abbeel, and Sergey Levine.
\newblock Model-agnostic meta-learning for fast adaptation of deep networks.
\newblock In \emph{International Conference on Machine Learning}, pages
  1126--1135. PMLR, 2017.

\bibitem[Florensa et~al.(2017)Florensa, Held, Wulfmeier, Zhang, and
  Abbeel]{florensa2017reverse}
Carlos Florensa, David Held, Markus Wulfmeier, Michael Zhang, and Pieter
  Abbeel.
\newblock Reverse curriculum generation for reinforcement learning.
\newblock In \emph{Conference on robot learning}, pages 482--495. PMLR, 2017.

\bibitem[Fox et~al.(1996)Fox, Hershberger, and Bouchard]{Fox1996-po}
P~W Fox, S~L Hershberger, and T~J Bouchard, Jr.
\newblock Genetic and environmental contributions to the acquisition of a motor
  skill.
\newblock \emph{Nature}, 384\penalty0 (6607):\penalty0 356--358, November 1996.

\bibitem[Fu et~al.(2016)Fu, Levine, and Abbeel]{fu2016one}
Justin Fu, Sergey Levine, and Pieter Abbeel.
\newblock One-shot learning of manipulation skills with online dynamics
  adaptation and neural network priors.
\newblock In \emph{2016 IEEE/RSJ International Conference on Intelligent Robots
  and Systems (IROS)}, pages 4019--4026. IEEE, 2016.

\bibitem[Fu et~al.(2017)Fu, Luo, and Levine]{fu2017learning}
Justin Fu, Katie Luo, and Sergey Levine.
\newblock Learning robust rewards with adversarial inverse reinforcement
  learning.
\newblock \emph{arXiv preprint arXiv:1710.11248}, 2017.

\bibitem[Fu et~al.(2019)Fu, Balan, Levine, and Guadarrama]{Fu2019FromLT}
Justin Fu, Anoop~Korattikara Balan, Sergey Levine, and Sergio Guadarrama.
\newblock From language to goals: Inverse reinforcement learning for
  vision-based instruction following.
\newblock \emph{ArXiv}, abs/1902.07742, 2019.

\bibitem[Fu et~al.(2020)Fu, Kumar, Nachum, Tucker, and Levine]{Fu_2020}
Justin Fu, Aviral Kumar, Ofir Nachum, George Tucker, and Sergey Levine.
\newblock D4rl: Datasets for deep data-driven reinforcement learning.
\newblock \emph{arXiv preprint arXiv:2004.07219}, 2020.

\bibitem[Fujimoto et~al.(2018)Fujimoto, Meger, and
  Precup]{fujimoto2018offpolicy}
Scott Fujimoto, David Meger, and Doina Precup.
\newblock Off-policy deep reinforcement learning without exploration, 2018.

\bibitem[Galashov et~al.(2019{\natexlab{a}})Galashov, Jayakumar, Hasenclever,
  Tirumala, Schwarz, Desjardins, Czarnecki, Teh, Pascanu, and
  Heess]{galashov2019information}
Alexandre Galashov, Siddhant~M Jayakumar, Leonard Hasenclever, Dhruva Tirumala,
  Jonathan Schwarz, Guillaume Desjardins, Wojciech~M Czarnecki, Yee~Whye Teh,
  Razvan Pascanu, and Nicolas Heess.
\newblock Information asymmetry in kl-regularized rl.
\newblock \emph{arXiv preprint arXiv:1905.01240}, 2019{\natexlab{a}}.

\bibitem[Galashov et~al.(2019{\natexlab{b}})Galashov, Schwarz, Kim, Garnelo,
  Saxton, Kohli, Eslami, and Teh]{galashov2019meta}
Alexandre Galashov, Jonathan Schwarz, Hyunjik Kim, Marta Garnelo, David Saxton,
  Pushmeet Kohli, SM~Eslami, and Yee~Whye Teh.
\newblock Meta-learning surrogate models for sequential decision making.
\newblock \emph{arXiv preprint arXiv:1903.11907}, 2019{\natexlab{b}}.

\bibitem[Galashov et~al.(2020)Galashov, Sygnowski, Desjardins, Humplik,
  Hasenclever, Jeong, Teh, and Heess]{galashov2020importance}
Alexandre Galashov, Jakub Sygnowski, Guillaume Desjardins, Jan Humplik, Leonard
  Hasenclever, Rae Jeong, Yee~Whye Teh, and Nicolas Heess.
\newblock Importance weighted policy learning and adaption.
\newblock \emph{arXiv preprint arXiv:2009.04875}, 2020.

\bibitem[Gao et~al.(2020)Gao, Biderman, Black, Golding, Hoppe, Foster, Phang,
  He, Thite, Nabeshima, Presser, and Leahy]{Gao2020ThePA}
Leo Gao, Stella~Rose Biderman, Sid Black, Laurence Golding, Travis Hoppe,
  Charles Foster, Jason Phang, Horace He, Anish Thite, Noa Nabeshima, Shawn
  Presser, and Connor Leahy.
\newblock The pile: An 800gb dataset of diverse text for language modeling.
\newblock \emph{ArXiv}, abs/2101.00027, 2020.

\bibitem[Garcia et~al.(1989)Garcia, Prett, and Morari]{garcia1989model}
Carlos~E Garcia, David~M Prett, and Manfred Morari.
\newblock Model predictive control: Theory and practice—a survey.
\newblock \emph{Automatica}, 25\penalty0 (3):\penalty0 335--348, 1989.

\bibitem[Garg et~al.(2022)Garg, Tsipras, Liang, and Valiant]{garg2022can}
Shivam Garg, Dimitris Tsipras, Percy~S Liang, and Gregory Valiant.
\newblock What can transformers learn in-context? a case study of simple
  function classes.
\newblock \emph{Advances in Neural Information Processing Systems},
  35:\penalty0 30583--30598, 2022.

\bibitem[Gehring et~al.(2021)Gehring, Synnaeve, Krause, and
  Usunier]{Gehring2021HierarchicalSF}
Jonas Gehring, Gabriel Synnaeve, Andreas Krause, and Nicolas Usunier.
\newblock Hierarchical skills for efficient exploration.
\newblock \emph{ArXiv}, abs/2110.10809, 2021.

\bibitem[Gervet et~al.(2023)Gervet, Chintala, Batra, Malik, and
  Chaplot]{gervet2023navigating}
Theophile Gervet, Soumith Chintala, Dhruv Batra, Jitendra Malik, and
  Devendra~Singh Chaplot.
\newblock Navigating to objects in the real world.
\newblock \emph{Science Robotics}, 8\penalty0 (79):\penalty0 eadf6991, 2023.

\bibitem[Ghosh et~al.(2021)Ghosh, Rahme, Kumar, Zhang, Adams, and
  Levine]{Ghosh2021WhyGI}
Dibya Ghosh, Jad Rahme, Aviral Kumar, Amy Zhang, Ryan~P. Adams, and Sergey
  Levine.
\newblock Why generalization in rl is difficult: Epistemic pomdps and implicit
  partial observability.
\newblock \emph{ArXiv}, abs/2107.06277, 2021.

\bibitem[Glorot and Bengio(2010)]{glorot2010understanding}
Xavier Glorot and Yoshua Bengio.
\newblock Understanding the difficulty of training deep feedforward neural
  networks.
\newblock In \emph{Proceedings of the thirteenth international conference on
  artificial intelligence and statistics}, pages 249--256. JMLR Workshop and
  Conference Proceedings, 2010.

\bibitem[Goodfellow et~al.(2013)Goodfellow, Mirza, Xiao, Courville, and
  Bengio]{goodfellow2013empirical}
Ian~J Goodfellow, Mehdi Mirza, Da~Xiao, Aaron Courville, and Yoshua Bengio.
\newblock An empirical investigation of catastrophic forgetting in
  gradient-based neural networks.
\newblock \emph{arXiv preprint arXiv:1312.6211}, 2013.

\bibitem[Goyal et~al.(2019)Goyal, Sodhani, Binas, Peng, Levine, and
  Bengio]{Goyal2019ReinforcementLW}
Anirudh Goyal, Shagun Sodhani, Jonathan Binas, Xue~Bin Peng, Sergey Levine, and
  Yoshua Bengio.
\newblock Reinforcement learning with competitive ensembles of
  information-constrained primitives.
\newblock \emph{ArXiv}, abs/1906.10667, 2019.

\bibitem[Goyal et~al.(2022)Goyal, Friesen, Banino, Weber, Ke, Badia, Guez,
  Mirza, Humphreys, Konyushova, et~al.]{goyal2022retrieval}
Anirudh Goyal, Abram Friesen, Andrea Banino, Theophane Weber, Nan~Rosemary Ke,
  Adria~Puigdomenech Badia, Arthur Guez, Mehdi Mirza, Peter~C Humphreys, Ksenia
  Konyushova, et~al.
\newblock Retrieval-augmented reinforcement learning.
\newblock In \emph{International Conference on Machine Learning}, pages
  7740--7765. PMLR, 2022.

\bibitem[Gregor et~al.(2017)Gregor, Rezende, and
  Wierstra]{Gregor2017VariationalIC}
Karol Gregor, Danilo~Jimenez Rezende, and Daan Wierstra.
\newblock Variational intrinsic control.
\newblock \emph{ArXiv}, abs/1611.07507, 2017.

\bibitem[Greydanus et~al.(2019)Greydanus, Dzamba, and
  Yosinski]{Greydanus2019HamiltonianNN}
Sam Greydanus, Misko Dzamba, and Jason Yosinski.
\newblock Hamiltonian neural networks.
\newblock \emph{ArXiv}, abs/1906.01563, 2019.

\bibitem[Grigsby and Qi(2020)]{grigsby2020measuring}
Jake Grigsby and Yanjun Qi.
\newblock Measuring visual generalization in continuous control from pixels,
  2020.

\bibitem[Grill et~al.(2020)Grill, Strub, Altch{\'e}, Tallec, Richemond,
  Buchatskaya, Doersch, Avila~Pires, Guo, Gheshlaghi~Azar,
  et~al.]{grill2020bootstrap}
Jean-Bastien Grill, Florian Strub, Florent Altch{\'e}, Corentin Tallec, Pierre
  Richemond, Elena Buchatskaya, Carl Doersch, Bernardo Avila~Pires, Zhaohan
  Guo, Mohammad Gheshlaghi~Azar, et~al.
\newblock Bootstrap your own latent-a new approach to self-supervised learning.
\newblock \emph{Advances in Neural Information Processing Systems},
  33:\penalty0 21271--21284, 2020.

\bibitem[Grimm et~al.(2019)Grimm, Higgins, Barreto, Teplyashin, Wulfmeier,
  Hertweck, Hadsell, and Singh]{grimm2019disentangled}
Christopher Grimm, Irina Higgins, Andre Barreto, Denis Teplyashin, Markus
  Wulfmeier, Tim Hertweck, Raia Hadsell, and Satinder Singh.
\newblock Disentangled cumulants help successor representations transfer to new
  tasks.
\newblock \emph{arXiv preprint arXiv:1911.10866}, 2019.

\bibitem[Groth et~al.(2021)Groth, Wulfmeier, Vezzani, Dasagi, Hertweck, Hafner,
  Heess, and Riedmiller]{Groth2021IsCA}
Oliver Groth, Markus Wulfmeier, Giulia Vezzani, Vibhavari Dasagi, Tim Hertweck,
  Roland Hafner, Nicolas Manfred~Otto Heess, and Martin~A. Riedmiller.
\newblock Is curiosity all you need? on the utility of emergent behaviours from
  curious exploration.
\newblock \emph{ArXiv}, abs/2109.08603, 2021.

\bibitem[Gulcehre et~al.(2020)Gulcehre, Wang, Novikov, Paine, G\'{o}mez, Zolna,
  Agarwal, Merel, Mankowitz, Paduraru, Dulac-Arnold, Li, Norouzi, Hoffman,
  Heess, and de~Freitas]{Gulcehre_2020}
Caglar Gulcehre, Ziyu Wang, Alexander Novikov, Thomas Paine, Sergio G\'{o}mez,
  Konrad Zolna, Rishabh Agarwal, Josh~S Merel, Daniel~J Mankowitz, Cosmin
  Paduraru, Gabriel Dulac-Arnold, Jerry Li, Mohammad Norouzi, Matthew Hoffman,
  Nicolas Heess, and Nando de~Freitas.
\newblock Rl unplugged: A suite of benchmarks for offline reinforcement
  learning.
\newblock In \emph{Advances in Neural Information Processing Systems}, 2020.

\bibitem[Guo et~al.(2020)Guo, Rush, and Kim]{guo2020parameter}
Demi Guo, Alexander~M Rush, and Yoon Kim.
\newblock Parameter-efficient transfer learning with diff pruning.
\newblock \emph{arXiv preprint arXiv:2012.07463}, 2020.

\bibitem[Gupta et~al.(2017)Gupta, Devin, Liu, Abbeel, and
  Levine]{Gupta2017LearningIF}
Abhishek Gupta, Coline Devin, Yuxuan Liu, P.~Abbeel, and Sergey Levine.
\newblock Learning invariant feature spaces to transfer skills with
  reinforcement learning.
\newblock \emph{ArXiv}, abs/1703.02949, 2017.

\bibitem[Gupta et~al.(2018{\natexlab{a}})Gupta, Eysenbach, Finn, and
  Levine]{Gupta2018UnsupervisedMF}
Abhishek Gupta, Benjamin Eysenbach, Chelsea Finn, and Sergey Levine.
\newblock Unsupervised meta-learning for reinforcement learning.
\newblock \emph{ArXiv}, abs/1806.04640, 2018{\natexlab{a}}.

\bibitem[Gupta et~al.(2018{\natexlab{b}})Gupta, Mendonca, Liu, Abbeel, and
  Levine]{Gupta2018MetaReinforcementLO}
Abhishek Gupta, Russell Mendonca, Yuxuan Liu, P.~Abbeel, and Sergey Levine.
\newblock Meta-reinforcement learning of structured exploration strategies.
\newblock In \emph{NeurIPS}, 2018{\natexlab{b}}.

\bibitem[Gupta et~al.(2019)Gupta, Kumar, Lynch, Levine, and
  Hausman]{gupta2019relay}
Abhishek Gupta, Vikash Kumar, Corey Lynch, Sergey Levine, and Karol Hausman.
\newblock Relay policy learning: Solving long-horizon tasks via imitation and
  reinforcement learning.
\newblock \emph{arXiv preprint arXiv:1910.11956}, 2019.

\bibitem[Gupta et~al.(2021)Gupta, Yu, Zhao, Kumar, Rovinsky, Xu, Devlin, and
  Levine]{gupta2021reset}
Abhishek Gupta, Justin Yu, Tony~Z Zhao, Vikash Kumar, Aaron Rovinsky, Kelvin
  Xu, Thomas Devlin, and Sergey Levine.
\newblock Reset-free reinforcement learning via multi-task learning: Learning
  dexterous manipulation behaviors without human intervention.
\newblock \emph{arXiv preprint arXiv:2104.11203}, 2021.

\bibitem[Guss et~al.(2019)Guss, Houghton, Topin, Wang, Codel, Veloso, and
  Salakhutdinov]{Guss2019MineRLAL}
William~H. Guss, Brandon Houghton, Nicholay Topin, Phillip Wang, Cayden Codel,
  Manuela~M. Veloso, and Ruslan Salakhutdinov.
\newblock Minerl: A large-scale dataset of minecraft demonstrations.
\newblock \emph{ArXiv}, abs/1907.13440, 2019.

\bibitem[Guu et~al.(2020)Guu, Lee, Tung, Pasupat, and Chang]{Guu2020REALMRL}
Kelvin Guu, Kenton Lee, Zora Tung, Panupong Pasupat, and Ming-Wei Chang.
\newblock Realm: Retrieval-augmented language model pre-training.
\newblock \emph{ArXiv}, abs/2002.08909, 2020.

\bibitem[Ha and Schmidhuber(2018)]{ha2018world}
David Ha and J{\"u}rgen Schmidhuber.
\newblock World models.
\newblock \emph{arXiv preprint arXiv:1803.10122}, 2018.

\bibitem[Haarnoja et~al.(2017)Haarnoja, Tang, Abbeel, and
  Levine]{haarnoja2017reinforcement}
Tuomas Haarnoja, Haoran Tang, Pieter Abbeel, and Sergey Levine.
\newblock Reinforcement learning with deep energy-based policies.
\newblock In \emph{International Conference on Machine Learning}, pages
  1352--1361. PMLR, 2017.

\bibitem[Haarnoja et~al.(2018{\natexlab{a}})Haarnoja, Hartikainen, Abbeel, and
  Levine]{haarnoja2018latent}
Tuomas Haarnoja, Kristian Hartikainen, Pieter Abbeel, and Sergey Levine.
\newblock Latent space policies for hierarchical reinforcement learning.
\newblock In \emph{International Conference on Machine Learning}, pages
  1851--1860. PMLR, 2018{\natexlab{a}}.

\bibitem[Haarnoja et~al.(2018{\natexlab{b}})Haarnoja, Pong, Zhou, Dalal,
  Abbeel, and Levine]{haarnoja2018composable}
Tuomas Haarnoja, Vitchyr Pong, Aurick Zhou, Murtaza Dalal, Pieter Abbeel, and
  Sergey Levine.
\newblock Composable deep reinforcement learning for robotic manipulation.
\newblock In \emph{2018 IEEE international conference on robotics and
  automation (ICRA)}, pages 6244--6251. IEEE, 2018{\natexlab{b}}.

\bibitem[Haarnoja et~al.(2018{\natexlab{c}})Haarnoja, Zhou, Abbeel, and
  Levine]{haarnoja2018soft}
Tuomas Haarnoja, Aurick Zhou, Pieter Abbeel, and Sergey Levine.
\newblock Soft actor-critic: Off-policy maximum entropy deep reinforcement
  learning with a stochastic actor.
\newblock In \emph{International conference on machine learning}, pages
  1861--1870. PMLR, 2018{\natexlab{c}}.

\bibitem[Haarnoja et~al.(2023)Haarnoja, Moran, Lever, Huang, Tirumala,
  Wulfmeier, Humplik, Tunyasuvunakool, Siegel, Hafner, Bloesch, Hartikainen,
  Byravan, Hasenclever, Tassa, Sadeghi, Batchelor, Casarini, Saliceti, Game,
  Sreendra, Patel, Gwira, Huber, Hurley, Nori, Hadsell, and
  Heess]{Haarnoja2023LearningAS}
Tuomas Haarnoja, Ben Moran, Guy Lever, Sandy~H. Huang, Dhruva Tirumala, Markus
  Wulfmeier, Jan Humplik, Saran Tunyasuvunakool, Noah Siegel, Roland Hafner,
  Michael Bloesch, Kristian Hartikainen, Arunkumar Byravan, Leonard
  Hasenclever, Yuval Tassa, Fereshteh Sadeghi, Nathan Batchelor, Federico
  Casarini, Stefano Saliceti, Charles Game, Neil Sreendra, Kushal Patel, Marlon
  Gwira, Andrea Huber, Nicole Hurley, Francesco Nori, Raia Hadsell, and Nicolas
  Manfred~Otto Heess.
\newblock Learning agile soccer skills for a bipedal robot with deep
  reinforcement learning.
\newblock \emph{ArXiv}, abs/2304.13653, 2023.

\bibitem[Hafner(2021)]{hafner2021benchmarking}
Danijar Hafner.
\newblock Benchmarking the spectrum of agent capabilities.
\newblock \emph{arXiv preprint arXiv:2109.06780}, 2021.

\bibitem[Hafner et~al.(2019)Hafner, Lillicrap, Ba, and
  Norouzi]{hafner2019dream}
Danijar Hafner, Timothy~P. Lillicrap, Jimmy Ba, and Mohammad Norouzi.
\newblock Dream to control: Learning behaviors by latent imagination.
\newblock \emph{CoRR}, abs/1912.01603, 2019.
\newblock URL \url{http://arxiv.org/abs/1912.01603}.

\bibitem[Hansen et~al.(2021{\natexlab{a}})Hansen, Su, and
  Wang]{hansen2021stabilizing}
Nicklas Hansen, Hao Su, and Xiaolong Wang.
\newblock Stabilizing deep q-learning with convnets and vision transformers
  under data augmentation.
\newblock \emph{Advances in neural information processing systems},
  34:\penalty0 3680--3693, 2021{\natexlab{a}}.

\bibitem[Hansen et~al.(2021{\natexlab{b}})Hansen, Sun, Abbeel, Efros, Pinto,
  and Wang]{Hansen2021SelfSupervisedPA}
Nicklas Hansen, Yu~Sun, P.~Abbeel, Alexei~A. Efros, Lerrel Pinto, and Xiaolong
  Wang.
\newblock Self-supervised policy adaptation during deployment.
\newblock \emph{ArXiv}, abs/2007.04309, 2021{\natexlab{b}}.

\bibitem[Hansen et~al.(2019)Hansen, Dabney, Barreto, Van~de Wiele,
  Warde-Farley, and Mnih]{hansen2019fast}
Steven Hansen, Will Dabney, Andre Barreto, Tom Van~de Wiele, David
  Warde-Farley, and Volodymyr Mnih.
\newblock Fast task inference with variational intrinsic successor features.
\newblock \emph{arXiv preprint arXiv:1906.05030}, 2019.

\bibitem[Hasenclever et~al.(2020)Hasenclever, Pardo, Hadsell, Heess, and
  Merel]{Hasenclever2020}
Leonard Hasenclever, Fabio Pardo, Raia Hadsell, Nicolas Heess, and Josh Merel.
\newblock {CoMic: C}omplementary task learning \& mimicry for reusable skills.
\newblock In \emph{Proceedings of the International Conference on Machine
  Learning (ICML)}, 2020.

\bibitem[Hausman et~al.(2018)Hausman, Springenberg, Wang, Heess, and
  Riedmiller]{hausman2018learning}
Karol Hausman, Jost~Tobias Springenberg, Ziyu Wang, Nicolas Heess, and Martin
  Riedmiller.
\newblock Learning an embedding space for transferable robot skills.
\newblock In \emph{International Conference on Learning Representations}, 2018.

\bibitem[He et~al.(2019)He, Girshick, and Doll{\'a}r]{he2019rethinking}
Kaiming He, Ross Girshick, and Piotr Doll{\'a}r.
\newblock Rethinking imagenet pre-training.
\newblock In \emph{Proceedings of the IEEE/CVF International Conference on
  Computer Vision}, pages 4918--4927, 2019.

\bibitem[Heess et~al.(2016)Heess, Wayne, Tassa, Lillicrap, Riedmiller, and
  Silver]{heess2016learning}
Nicolas Heess, Greg Wayne, Yuval Tassa, Timothy Lillicrap, Martin Riedmiller,
  and David Silver.
\newblock Learning and transfer of modulated locomotor controllers.
\newblock \emph{arXiv preprint arXiv:1610.05182}, 2016.

\bibitem[Held et~al.(2017)Held, McCarthy, Zhang, Shentu, and
  Abbeel]{held2017probabilistically}
David Held, Zoe McCarthy, Michael Zhang, Fred Shentu, and Pieter Abbeel.
\newblock Probabilistically safe policy transfer.
\newblock In \emph{2017 IEEE International Conference on Robotics and
  Automation (ICRA)}, pages 5798--5805. IEEE, 2017.

\bibitem[Hernandez et~al.(2021)Hernandez, Kaplan, Henighan, and
  McCandlish]{Hernandez2021ScalingLF}
Danny Hernandez, Jared Kaplan, T.~J. Henighan, and Sam McCandlish.
\newblock Scaling laws for transfer.
\newblock \emph{ArXiv}, abs/2102.01293, 2021.

\bibitem[Hester et~al.(2018)Hester, Vecerik, Pietquin, Lanctot, Schaul, Piot,
  Horgan, Quan, Sendonaris, Osband, et~al.]{hester2018deep}
Todd Hester, Matej Vecerik, Olivier Pietquin, Marc Lanctot, Tom Schaul, Bilal
  Piot, Dan Horgan, John Quan, Andrew Sendonaris, Ian Osband, et~al.
\newblock Deep q-learning from demonstrations.
\newblock In \emph{Proceedings of the AAAI Conference on Artificial
  Intelligence}, volume~32, 2018.

\bibitem[Hill et~al.(2020)Hill, Mokra, Wong, and Harley]{Hill2020HumanIW}
Felix Hill, Sona Mokra, Nathaniel Wong, and Tim Harley.
\newblock Human instruction-following with deep reinforcement learning via
  transfer-learning from text.
\newblock \emph{ArXiv}, abs/2005.09382, 2020.

\bibitem[Hinton and Salakhutdinov(2006)]{hinton2006reducing}
Geoffrey~E Hinton and Ruslan~R Salakhutdinov.
\newblock Reducing the dimensionality of data with neural networks.
\newblock \emph{science}, 313\penalty0 (5786):\penalty0 504--507, 2006.

\bibitem[Ho and Ermon(2016)]{ho2016generative}
Jonathan Ho and Stefano Ermon.
\newblock Generative adversarial imitation learning.
\newblock \emph{Advances in neural information processing systems},
  29:\penalty0 4565--4573, 2016.

\bibitem[Hofer et~al.(2020)Hofer, Bekris, Handa, Gamboa, Golemo, Mozifian,
  Atkeson, Fox, Goldberg, Leonard, Liu, Peters, Song, Welinder, and
  White]{Hofer2020PerspectivesOS}
Sebastian Hofer, Kostas~E. Bekris, Ankur Handa, Juan~Camilo Gamboa, Florian
  Golemo, Melissa Mozifian, C.~Atkeson, D.~Fox, Ken Goldberg, John Leonard,
  C.~Liu, Jan Peters, Shuran Song, P.~Welinder, and M.~White.
\newblock Perspectives on sim2real transfer for robotics: A summary of the r:
  Ss 2020 workshop.
\newblock \emph{ArXiv}, abs/2012.03806, 2020.

\bibitem[Hoque et~al.(2020)Hoque, Seita, Balakrishna, Ganapathi, Tanwani,
  Jamali, Yamane, Iba, and Goldberg]{hoque2020visuospatial}
Ryan Hoque, Daniel Seita, Ashwin Balakrishna, Aditya Ganapathi, Ajay~Kumar
  Tanwani, Nawid Jamali, Katsu Yamane, Soshi Iba, and Ken Goldberg.
\newblock Visuospatial foresight for multi-step, multi-task fabric
  manipulation.
\newblock \emph{arXiv preprint arXiv:2003.09044}, 2020.

\bibitem[Hoque et~al.(2021)Hoque, Seita, Balakrishna, Ganapathi, Tanwani,
  Jamali, Yamane, Iba, and Goldberg]{hoque2021visuospatial}
Ryan Hoque, Daniel Seita, Ashwin Balakrishna, Aditya Ganapathi, Ajay~Kumar
  Tanwani, Nawid Jamali, Katsu Yamane, Soshi Iba, and Ken Goldberg.
\newblock Visuospatial foresight for physical sequential fabric manipulation.
\newblock \emph{arXiv preprint arXiv:2102.09754}, 2021.

\bibitem[Houlsby et~al.(2019)Houlsby, Giurgiu, Jastrzebski, Morrone,
  De~Laroussilhe, Gesmundo, Attariyan, and Gelly]{houlsby2019parameter}
Neil Houlsby, Andrei Giurgiu, Stanislaw Jastrzebski, Bruna Morrone, Quentin
  De~Laroussilhe, Andrea Gesmundo, Mona Attariyan, and Sylvain Gelly.
\newblock Parameter-efficient transfer learning for nlp.
\newblock In \emph{International Conference on Machine Learning}, pages
  2790--2799. PMLR, 2019.

\bibitem[Houthooft et~al.(2018)Houthooft, Chen, Isola, Stadie, Wolski,
  Jonathan~Ho, and Abbeel]{houthooft2018evolved}
Rein Houthooft, Yuhua Chen, Phillip Isola, Bradly Stadie, Filip Wolski, OpenAI
  Jonathan~Ho, and Pieter Abbeel.
\newblock Evolved policy gradients.
\newblock \emph{Advances in Neural Information Processing Systems}, 31, 2018.

\bibitem[Hu et~al.(2021)Hu, Shen, Wallis, Allen-Zhu, Li, Wang, and
  Chen]{Hu2021LoRALA}
J.~Edward Hu, Yelong Shen, Phillip Wallis, Zeyuan Allen-Zhu, Yuanzhi Li, Shean
  Wang, and Weizhu Chen.
\newblock Lora: Low-rank adaptation of large language models.
\newblock \emph{ArXiv}, abs/2106.09685, 2021.

\bibitem[Huang et~al.(2022{\natexlab{a}})Huang, Mees, Zeng, and
  Burgard]{huang2022visual}
Chenguang Huang, Oier Mees, Andy Zeng, and Wolfram Burgard.
\newblock Visual language maps for robot navigation.
\newblock \emph{arXiv preprint arXiv:2210.05714}, 2022{\natexlab{a}}.

\bibitem[Huang et~al.(2022{\natexlab{b}})Huang, Abbeel, Pathak, and
  Mordatch]{huang2022language}
Wenlong Huang, Pieter Abbeel, Deepak Pathak, and Igor Mordatch.
\newblock Language models as zero-shot planners: Extracting actionable
  knowledge for embodied agents.
\newblock \emph{arXiv preprint arXiv:2201.07207}, 2022{\natexlab{b}}.

\bibitem[Huang et~al.(2022{\natexlab{c}})Huang, Xia, Xiao, Chan, Liang,
  Florence, Zeng, Tompson, Mordatch, Chebotar, et~al.]{huang2022inner}
Wenlong Huang, Fei Xia, Ted Xiao, Harris Chan, Jacky Liang, Pete Florence, Andy
  Zeng, Jonathan Tompson, Igor Mordatch, Yevgen Chebotar, et~al.
\newblock Inner monologue: Embodied reasoning through planning with language
  models.
\newblock \emph{arXiv preprint arXiv:2207.05608}, 2022{\natexlab{c}}.

\bibitem[HuggingFace(2022)]{downloaddt}
HuggingFace.
\newblock Decision transformer, 2022.
\newblock URL
  \url{https://huggingface.co/docs/transformers/model_doc/decision_transformer}.

\bibitem[Huh et~al.(2016)Huh, Agrawal, and Efros]{huh2016makes}
Minyoung Huh, Pulkit Agrawal, and Alexei~A Efros.
\newblock What makes imagenet good for transfer learning?
\newblock \emph{arXiv preprint arXiv:1608.08614}, 2016.

\bibitem[Humphreys et~al.(2022)Humphreys, Guez, Tieleman, Sifre, Weber, and
  Lillicrap]{humphreys2022large}
Peter Humphreys, Arthur Guez, Olivier Tieleman, Laurent Sifre, Th{\'e}ophane
  Weber, and Timothy Lillicrap.
\newblock Large-scale retrieval for reinforcement learning.
\newblock \emph{Advances in Neural Information Processing Systems},
  35:\penalty0 20092--20104, 2022.

\bibitem[Humplik et~al.(2019)Humplik, Galashov, Hasenclever, Ortega, Teh, and
  Heess]{Humplik2019MetaRL}
Jan Humplik, Alexandre Galashov, Leonard Hasenclever, Pedro~A. Ortega, Yee~Whye
  Teh, and Nicolas Manfred~Otto Heess.
\newblock Meta reinforcement learning as task inference.
\newblock \emph{ArXiv}, abs/1905.06424, 2019.

\bibitem[Hunt et~al.(2019)Hunt, Barreto, Lillicrap, and
  Heess]{hunt2019composing}
Jonathan Hunt, Andre Barreto, Timothy Lillicrap, and Nicolas Heess.
\newblock Composing entropic policies using divergence correction.
\newblock In \emph{International Conference on Machine Learning}, pages
  2911--2920. PMLR, 2019.

\bibitem[Hwangbo et~al.(2019)Hwangbo, Lee, Dosovitskiy, Bellicoso, Tsounis,
  Koltun, and Hutter]{hwangbo2019learning}
Jemin Hwangbo, Joonho Lee, Alexey Dosovitskiy, Dario Bellicoso, Vassilios
  Tsounis, Vladlen Koltun, and Marco Hutter.
\newblock Learning agile and dynamic motor skills for legged robots.
\newblock \emph{Science Robotics}, 4\penalty0 (26), 2019.
\newblock \doi{10.1126/scirobotics.aau5872}.
\newblock URL \url{https://robotics.sciencemag.org/content/4/26/eaau5872}.

\bibitem[Igl et~al.(2020)Igl, Farquhar, Luketina, B{\"o}hmer, and
  Whiteson]{igl2020transient}
Maximilian Igl, Gregory Farquhar, Jelena Luketina, Wendelin B{\"o}hmer, and
  Shimon Whiteson.
\newblock Transient non-stationarity and generalisation in deep reinforcement
  learning.
\newblock \emph{arXiv preprint arXiv:2006.05826}, 2020.

\bibitem[Jaderberg et~al.(2017)Jaderberg, Mnih, Czarnecki, Schaul, Leibo,
  Silver, and Kavukcuoglu]{Jaderberg2017ReinforcementLW}
Max Jaderberg, Volodymyr Mnih, Wojciech~M. Czarnecki, Tom Schaul, Joel~Z.
  Leibo, David Silver, and Koray Kavukcuoglu.
\newblock Reinforcement learning with unsupervised auxiliary tasks.
\newblock \emph{ArXiv}, abs/1611.05397, 2017.

\bibitem[James et~al.(2020)James, Ma, Rovick~Arrojo, and
  Davison]{james2019rlbench}
Stephen James, Zicong Ma, David Rovick~Arrojo, and Andrew~J. Davison.
\newblock Rlbench: The robot learning benchmark \& learning environment.
\newblock \emph{IEEE Robotics and Automation Letters}, 2020.

\bibitem[Janner et~al.(2020)Janner, Mordatch, and Levine]{janner2020generative}
Michael Janner, Igor Mordatch, and Sergey Levine.
\newblock Generative temporal difference learning for infinite-horizon
  prediction.
\newblock \emph{arXiv preprint arXiv:2010.14496}, 2020.

\bibitem[Jeong et~al.(2020)Jeong, Springenberg, Kay, Zheng, Zhou, Galashov,
  Heess, and Nori]{jeong2020learning}
Rae Jeong, Jost~Tobias Springenberg, Jackie Kay, Daniel Zheng, Yuxiang Zhou,
  Alexandre Galashov, Nicolas Heess, and Francesco Nori.
\newblock Learning dexterous manipulation from suboptimal experts.
\newblock \emph{CoRR}, abs/2010.08587, 2020.
\newblock URL \url{https://arxiv.org/abs/2010.08587}.

\bibitem[Jia et~al.(2021)Jia, Yang, Xia, Chen, Parekh, Pham, Le, Sung, Li, and
  Duerig]{jia2021scaling}
Chao Jia, Yinfei Yang, Ye~Xia, Yi-Ting Chen, Zarana Parekh, Hieu Pham, Quoc Le,
  Yun-Hsuan Sung, Zhen Li, and Tom Duerig.
\newblock Scaling up visual and vision-language representation learning with
  noisy text supervision.
\newblock In \emph{International Conference on Machine Learning}, pages
  4904--4916. PMLR, 2021.

\bibitem[Jiang et~al.(2019)Jiang, Gu, Murphy, and Finn]{jiang2019language}
Yiding Jiang, Shixiang~Shane Gu, Kevin~P Murphy, and Chelsea Finn.
\newblock Language as an abstraction for hierarchical deep reinforcement
  learning.
\newblock \emph{Advances in Neural Information Processing Systems}, 32, 2019.

\bibitem[Johannink et~al.(2019)Johannink, Bahl, Nair, Luo, Kumar, Loskyll,
  Ojea, Solowjow, and Levine]{Johannink2019ResidualRL}
Tobias Johannink, Shikhar Bahl, Ashvin Nair, Jianlan Luo, Avinash Kumar,
  Matthias Loskyll, Juan~Aparicio Ojea, Eugen Solowjow, and Sergey Levine.
\newblock Residual reinforcement learning for robot control.
\newblock \emph{2019 International Conference on Robotics and Automation
  (ICRA)}, pages 6023--6029, 2019.

\bibitem[Julian et~al.(2020)Julian, Swanson, Sukhatme, Levine, Finn, and
  Hausman]{julian2020never}
Ryan Julian, Benjamin Swanson, Gaurav~S Sukhatme, Sergey Levine, Chelsea Finn,
  and Karol Hausman.
\newblock Never stop learning: The effectiveness of fine-tuning in robotic
  reinforcement learning.
\newblock \emph{arXiv preprint arXiv:2004.10190}, 2020.

\bibitem[Kaelbling(1993)]{kaelbling1993learning}
Leslie~Pack Kaelbling.
\newblock Learning to achieve goals.
\newblock In \emph{IJCAI}, volume~2, pages 1094--8. Citeseer, 1993.

\bibitem[Kalashnikov et~al.(2021)Kalashnikov, Varley, Chebotar, Swanson,
  Jonschkowski, Finn, Levine, and Hausman]{kalashnikov2021mt}
Dmitry Kalashnikov, Jacob Varley, Yevgen Chebotar, Benjamin Swanson, Rico
  Jonschkowski, Chelsea Finn, Sergey Levine, and Karol Hausman.
\newblock Mt-opt: Continuous multi-task robotic reinforcement learning at
  scale.
\newblock \emph{arXiv preprint arXiv:2104.08212}, 2021.

\bibitem[Kang et~al.(2021)Kang, Kahn, and Levine]{kang2021hierarchically}
Katie Kang, Gregory Kahn, and Sergey Levine.
\newblock Hierarchically integrated models: Learning to navigate from
  heterogeneous robots.
\newblock In \emph{5th Annual Conference on Robot Learning}, 2021.

\bibitem[Kannan et~al.(2021)Kannan, Hafner, Finn, and
  Erhan]{kannan2021robodesk}
Harini Kannan, Danijar Hafner, Chelsea Finn, and Dumitru Erhan.
\newblock Robodesk: A multi-task reinforcement learning benchmark.
\newblock \url{https://github.com/google-research/robodesk}, 2021.

\bibitem[Kaufmann et~al.(2023)Kaufmann, Bauersfeld, Loquercio, M{\"u}ller,
  Koltun, and Scaramuzza]{kaufmann2023champion}
Elia Kaufmann, Leonard Bauersfeld, Antonio Loquercio, Matthias M{\"u}ller,
  Vladlen Koltun, and Davide Scaramuzza.
\newblock Champion-level drone racing using deep reinforcement learning.
\newblock \emph{Nature}, 620\penalty0 (7976):\penalty0 982--987, 2023.

\bibitem[Kaushik et~al.(2020)Kaushik, Anne, and Mouret]{kaushik2020fast}
Rituraj Kaushik, Timoth{\'e}e Anne, and Jean-Baptiste Mouret.
\newblock Fast online adaptation in robotics through meta-learning embeddings
  of simulated priors.
\newblock In \emph{2020 IEEE/RSJ International Conference on Intelligent Robots
  and Systems (IROS)}, pages 5269--5276. IEEE, 2020.

\bibitem[Khandelwal et~al.(2022)Khandelwal, Weihs, Mottaghi, and
  Kembhavi]{khandelwal2022simple}
Apoorv Khandelwal, Luca Weihs, Roozbeh Mottaghi, and Aniruddha Kembhavi.
\newblock Simple but effective: Clip embeddings for embodied ai.
\newblock In \emph{Proceedings of the IEEE/CVF Conference on Computer Vision
  and Pattern Recognition}, pages 14829--14838, 2022.

\bibitem[Killian et~al.(2017)Killian, Daulton, Konidaris, and
  Doshi-Velez]{killian2017robust}
Taylor~W Killian, Samuel Daulton, George Konidaris, and Finale Doshi-Velez.
\newblock Robust and efficient transfer learning with hidden parameter markov
  decision processes.
\newblock \emph{Advances in neural information processing systems}, 30, 2017.

\bibitem[Kim and Park(2018)]{Kim2018ImitationLV}
Kee-Eung Kim and Hyun~Soo Park.
\newblock Imitation learning via kernel mean embedding.
\newblock In \emph{AAAI}, 2018.

\bibitem[Kirk et~al.(2021)Kirk, Zhang, Grefenstette, and
  Rockt{\"a}schel]{kirk2021survey}
Robert Kirk, Amy Zhang, Edward Grefenstette, and Tim Rockt{\"a}schel.
\newblock A survey of generalisation in deep reinforcement learning.
\newblock \emph{arXiv preprint arXiv:2111.09794}, 2021.

\bibitem[Kirkpatrick et~al.(2017)Kirkpatrick, Pascanu, Rabinowitz, Veness,
  Desjardins, Rusu, Milan, Quan, Ramalho, Grabska-Barwinska,
  et~al.]{kirkpatrick2017overcoming}
James Kirkpatrick, Razvan Pascanu, Neil Rabinowitz, Joel Veness, Guillaume
  Desjardins, Andrei~A Rusu, Kieran Milan, John Quan, Tiago Ramalho, Agnieszka
  Grabska-Barwinska, et~al.
\newblock Overcoming catastrophic forgetting in neural networks.
\newblock \emph{Proceedings of the national academy of sciences}, 114\penalty0
  (13):\penalty0 3521--3526, 2017.

\bibitem[Kojima et~al.(2022)Kojima, Gu, Reid, Matsuo, and
  Iwasawa]{Kojima2022LargeLM}
Takeshi Kojima, Shixiang~Shane Gu, Machel Reid, Yutaka Matsuo, and Yusuke
  Iwasawa.
\newblock Large language models are zero-shot reasoners.
\newblock \emph{ArXiv}, abs/2205.11916, 2022.

\bibitem[Konidaris and Barto(2006)]{Konidaris2006AutonomousSK}
G.~Konidaris and A.~Barto.
\newblock Autonomous shaping: knowledge transfer in reinforcement learning.
\newblock \emph{Proceedings of the 23rd international conference on Machine
  learning}, 2006.

\bibitem[Konidaris and Barto(2007)]{konidaris2007}
George Konidaris and Andrew Barto.
\newblock Building portable options: Skill transfer in reinforcement learning.
\newblock In \emph{Proceedings of the 20th International Joint Conference on
  Artifical Intelligence}, IJCAI'07, page 895–900, San Francisco, CA, USA,
  2007. Morgan Kaufmann Publishers Inc.

\bibitem[Kornblith et~al.(2019)Kornblith, Shlens, and Le]{Kornblith2019DoBI}
Simon Kornblith, Jonathon Shlens, and Quoc~V. Le.
\newblock Do better imagenet models transfer better?
\newblock \emph{2019 IEEE/CVF Conference on Computer Vision and Pattern
  Recognition (CVPR)}, pages 2656--2666, 2019.

\bibitem[Kostrikov et~al.(2019)Kostrikov, Nachum, and
  Tompson]{kostrikov2019imitation}
Ilya Kostrikov, Ofir Nachum, and Jonathan Tompson.
\newblock Imitation learning via off-policy distribution matching.
\newblock \emph{arXiv preprint arXiv:1912.05032}, 2019.

\bibitem[Kostrikov et~al.(2021)Kostrikov, Nair, and
  Levine]{kostrikov2021offline}
Ilya Kostrikov, Ashvin Nair, and Sergey Levine.
\newblock Offline reinforcement learning with implicit q-learning.
\newblock \emph{arXiv preprint arXiv:2110.06169}, 2021.

\bibitem[Kulkarni et~al.(2016)Kulkarni, Saeedi, Gautam, and
  Gershman]{kulkarni2016deep}
Tejas~D Kulkarni, Ardavan Saeedi, Simanta Gautam, and Samuel~J Gershman.
\newblock Deep successor reinforcement learning.
\newblock \emph{arXiv preprint arXiv:1606.02396}, 2016.

\bibitem[Kumar et~al.(2022{\natexlab{a}})Kumar, Raghunathan, Jones, Ma, and
  Liang]{Kumar2022FineTuningCD}
Ananya Kumar, Aditi Raghunathan, Robbie Jones, Tengyu Ma, and Percy Liang.
\newblock Fine-tuning can distort pretrained features and underperform
  out-of-distribution.
\newblock \emph{arXiv preprint arXiv:2202.10054}, 2022{\natexlab{a}}.

\bibitem[Kumar et~al.(2021)Kumar, Fu, Pathak, and Malik]{Kumar2021RMARM}
Ashish Kumar, Zipeng Fu, Deepak Pathak, and Jitendra Malik.
\newblock Rma: Rapid motor adaptation for legged robots.
\newblock \emph{ArXiv}, abs/2107.04034, 2021.

\bibitem[Kumar et~al.(2020{\natexlab{a}})Kumar, Zhou, Tucker, and
  Levine]{kumar2020conservative}
Aviral Kumar, Aurick Zhou, George Tucker, and Sergey Levine.
\newblock Conservative q-learning for offline reinforcement learning.
\newblock \emph{arXiv preprint arXiv:2006.04779}, 2020{\natexlab{a}}.

\bibitem[Kumar et~al.(2022{\natexlab{b}})Kumar, Singh, Ebert, Yang, Finn, and
  Levine]{kumar2022pre}
Aviral Kumar, Anikait Singh, Frederik Ebert, Yanlai Yang, Chelsea Finn, and
  Sergey Levine.
\newblock Pre-training for robots: Offline rl enables learning new tasks from a
  handful of trials.
\newblock \emph{arXiv preprint arXiv:2210.05178}, 2022{\natexlab{b}}.

\bibitem[Kumar et~al.(2020{\natexlab{b}})Kumar, Kumar, Levine, and
  Finn]{kumar2020one}
Saurabh Kumar, Aviral Kumar, Sergey Levine, and Chelsea Finn.
\newblock One solution is not all you need: Few-shot extrapolation via
  structured maxent rl.
\newblock \emph{Advances in Neural Information Processing Systems},
  33:\penalty0 8198--8210, 2020{\natexlab{b}}.

\bibitem[Kupcsik et~al.(2017)Kupcsik, Deisenroth, Peters, Loh, Vadakkepat, and
  Neumann]{kupcsik2017model}
Andras Kupcsik, Marc~Peter Deisenroth, Jan Peters, Ai~Poh Loh, Prahlad
  Vadakkepat, and Gerhard Neumann.
\newblock Model-based contextual policy search for data-efficient
  generalization of robot skills.
\newblock \emph{Artificial Intelligence}, 247:\penalty0 415--439, 2017.

\bibitem[Kurenkov et~al.(2019)Kurenkov, Mandlekar, Martin, Savarese, and
  Garg]{Kurenkov2019ACTeachAB}
Andrey Kurenkov, Ajay Mandlekar, Roberto~Martin Martin, Silvio Savarese, and
  Animesh Garg.
\newblock Ac-teach: A bayesian actor-critic method for policy learning with an
  ensemble of suboptimal teachers.
\newblock \emph{ArXiv}, abs/1909.04121, 2019.

\bibitem[Lambert et~al.(2022)Lambert, Wulfmeier, Whitney, Byravan, Bloesch,
  Dasagi, Hertweck, and Riedmiller]{Lambert2022TheCO}
Nathan Lambert, Markus Wulfmeier, William~F. Whitney, Arunkumar Byravan,
  Michael Bloesch, Vibhavari Dasagi, Tim Hertweck, and Martin~A. Riedmiller.
\newblock The challenges of exploration for offline reinforcement learning.
\newblock \emph{ArXiv}, abs/2201.11861, 2022.

\bibitem[Lange and Riedmiller(2010)]{lange2010deep}
Sascha Lange and Martin Riedmiller.
\newblock Deep auto-encoder neural networks in reinforcement learning.
\newblock In \emph{The 2010 International Joint Conference on Neural Networks
  (IJCNN)}, pages 1--8. IEEE, 2010.

\bibitem[Laroche and Barlier(2017)]{laroche2017transfer}
Romain Laroche and Merwan Barlier.
\newblock Transfer reinforcement learning with shared dynamics.
\newblock In \emph{Proceedings of the AAAI Conference on Artificial
  Intelligence}, volume~31, 2017.

\bibitem[Laskin et~al.(2022)Laskin, Wang, Oh, Parisotto, Spencer, Steigerwald,
  Strouse, Hansen, Filos, Brooks, et~al.]{laskin2022context}
Michael Laskin, Luyu Wang, Junhyuk Oh, Emilio Parisotto, Stephen Spencer,
  Richie Steigerwald, DJ~Strouse, Steven Hansen, Angelos Filos, Ethan Brooks,
  et~al.
\newblock In-context reinforcement learning with algorithm distillation.
\newblock \emph{arXiv preprint arXiv:2210.14215}, 2022.

\bibitem[Lawson and Qureshi(2023)]{Lawson2023MergingDT}
Daniel Lawson and Ahmed~Hussain Qureshi.
\newblock Merging decision transformers: Weight averaging for forming
  multi-task policies.
\newblock \emph{ArXiv}, abs/2303.07551, 2023.

\bibitem[Lazaric(2012)]{Lazaric2012-mv}
Alessandro Lazaric.
\newblock Transfer in reinforcement learning: A framework and a survey.
\newblock In Marco Wiering and Martijn van Otterlo, editors,
  \emph{Reinforcement Learning: {State-of-the-Art}}, pages 143--173. Springer
  Berlin Heidelberg, Berlin, Heidelberg, 2012.

\bibitem[Lee et~al.(2021{\natexlab{a}})Lee, Devin, Zhou, Lampe, Bousmalis,
  Springenberg, Byravan, Abdolmaleki, Gileadi, Khosid, et~al.]{lee2021beyond}
Alex~X Lee, Coline~Manon Devin, Yuxiang Zhou, Thomas Lampe, Konstantinos
  Bousmalis, Jost~Tobias Springenberg, Arunkumar Byravan, Abbas Abdolmaleki,
  Nimrod Gileadi, David Khosid, et~al.
\newblock Beyond pick-and-place: Tackling robotic stacking of diverse shapes.
\newblock In \emph{5th Annual Conference on Robot Learning},
  2021{\natexlab{a}}.

\bibitem[Lee et~al.(2020{\natexlab{a}})Lee, Hwangbo, Wellhausen, Koltun, and
  Hutter]{lee2020learning}
Joonho Lee, Jemin Hwangbo, Lorenz Wellhausen, Vladlen Koltun, and Marco Hutter.
\newblock Learning quadrupedal locomotion over challenging terrain.
\newblock \emph{Science robotics}, 5\penalty0 (47):\penalty0 eabc5986,
  2020{\natexlab{a}}.

\bibitem[Lee et~al.(2020{\natexlab{b}})Lee, Seo, Lee, Lee, and
  Shin]{lee2020context}
Kimin Lee, Younggyo Seo, Seunghyun Lee, Honglak Lee, and Jinwoo Shin.
\newblock Context-aware dynamics model for generalization in model-based
  reinforcement learning.
\newblock In \emph{International Conference on Machine Learning}, pages
  5757--5766. PMLR, 2020{\natexlab{b}}.

\bibitem[Lee et~al.(2021{\natexlab{b}})Lee, Smith, and Abbeel]{lee2021pebble}
Kimin Lee, Laura Smith, and Pieter Abbeel.
\newblock Pebble: Feedback-efficient interactive reinforcement learning via
  relabeling experience and unsupervised pre-training.
\newblock \emph{arXiv preprint arXiv:2106.05091}, 2021{\natexlab{b}}.

\bibitem[Lee et~al.(2022{\natexlab{a}})Lee, Nachum, Yang, Lee, Freeman, Xu,
  Guadarrama, Fischer, Jang, Michalewski, et~al.]{lee2022multi}
Kuang-Huei Lee, Ofir Nachum, Mengjiao Yang, Lisa Lee, Daniel Freeman, Winnie
  Xu, Sergio Guadarrama, Ian Fischer, Eric Jang, Henryk Michalewski, et~al.
\newblock Multi-game decision transformers.
\newblock \emph{arXiv preprint arXiv:2205.15241}, 2022{\natexlab{a}}.

\bibitem[Lee et~al.(2019)Lee, Eysenbach, Parisotto, Xing, Levine, and
  Salakhutdinov]{Lee2019EfficientEV}
Lisa Lee, Benjamin Eysenbach, Emilio Parisotto, Eric~P. Xing, Sergey Levine,
  and Ruslan Salakhutdinov.
\newblock Efficient exploration via state marginal matching.
\newblock \emph{ArXiv}, abs/1906.05274, 2019.

\bibitem[Lee et~al.(2022{\natexlab{b}})Lee, Seo, Lee, Abbeel, and
  Shin]{lee2022offline}
Seunghyun Lee, Younggyo Seo, Kimin Lee, Pieter Abbeel, and Jinwoo Shin.
\newblock Offline-to-online reinforcement learning via balanced replay and
  pessimistic q-ensemble.
\newblock In \emph{Conference on Robot Learning}, pages 1702--1712. PMLR,
  2022{\natexlab{b}}.

\bibitem[Lee and Chung(2021)]{lee2021improving}
Suyoung Lee and Sae-Young Chung.
\newblock Improving generalization in meta-rl with imaginary tasks from latent
  dynamics mixture.
\newblock \emph{Advances in Neural Information Processing Systems},
  34:\penalty0 27222--27235, 2021.

\bibitem[Lee et~al.(2018)Lee, Sun, Somasundaram, Hu, and
  Lim]{Lee2018ComposingCS}
Youngwoon Lee, Shao-Hua Sun, Sriram Somasundaram, Edward~S. Hu, and Joseph~J.
  Lim.
\newblock Composing complex skills by learning transition policies.
\newblock In \emph{International Conference on Learning Representations}, 2018.

\bibitem[Lee et~al.(2021{\natexlab{c}})Lee, Lim, Anandkumar, and
  Zhu]{Lee2021AdversarialSC}
Youngwoon Lee, Joseph~J. Lim, Anima Anandkumar, and Yuke Zhu.
\newblock Adversarial skill chaining for long-horizon robot manipulation via
  terminal state regularization.
\newblock In \emph{Conference on Robot Learning}, 2021{\natexlab{c}}.

\bibitem[Lengyel and Dayan(2007)]{Lengyel2007HippocampalCT}
M{\'a}t{\'e} Lengyel and Peter Dayan.
\newblock Hippocampal contributions to control: The third way.
\newblock In \emph{NIPS}, 2007.

\bibitem[Lenz et~al.(2015)Lenz, Knepper, and Saxena]{lenz2015deepmpc}
Ian Lenz, Ross~A Knepper, and Ashutosh Saxena.
\newblock Deepmpc: Learning deep latent features for model predictive control.
\newblock In \emph{Robotics: Science and Systems}. Rome, Italy, 2015.

\bibitem[Lester et~al.(2021)Lester, Al-Rfou, and Constant]{Lester2021ThePO}
Brian Lester, Rami Al-Rfou, and Noah Constant.
\newblock The power of scale for parameter-efficient prompt tuning.
\newblock In \emph{Conference on Empirical Methods in Natural Language
  Processing}, 2021.
\newblock URL \url{https://api.semanticscholar.org/CorpusID:233296808}.

\bibitem[Levine(2021)]{Levine2021UnderstandingTW}
Sergey Levine.
\newblock Understanding the world through action.
\newblock \emph{ArXiv}, abs/2110.12543, 2021.

\bibitem[Levine et~al.(2016)Levine, Finn, Darrell, and Abbeel]{levine2016end}
Sergey Levine, Chelsea Finn, Trevor Darrell, and Pieter Abbeel.
\newblock End-to-end training of deep visuomotor policies.
\newblock \emph{The Journal of Machine Learning Research}, 17\penalty0
  (1):\penalty0 1334--1373, 2016.

\bibitem[Levine et~al.(2020)Levine, Kumar, Tucker, and Fu]{levine2020offline}
Sergey Levine, Aviral Kumar, George Tucker, and Justin Fu.
\newblock Offline reinforcement learning: Tutorial, review, and perspectives on
  open problems.
\newblock \emph{arXiv preprint arXiv:2005.01643}, 2020.

\bibitem[Levy et~al.(2017{\natexlab{a}})Levy, Jr., and
  Saenko]{levy2018hierarchical}
Andrew Levy, Robert~Platt Jr., and Kate Saenko.
\newblock Hierarchical actor-critic.
\newblock \emph{CoRR}, abs/1712.00948, 2017{\natexlab{a}}.
\newblock URL \url{http://arxiv.org/abs/1712.00948}.

\bibitem[Levy et~al.(2017{\natexlab{b}})Levy, Konidaris, Platt, and
  Saenko]{levy2017learning}
Andrew Levy, George Konidaris, Robert Platt, and Kate Saenko.
\newblock Learning multi-level hierarchies with hindsight.
\newblock \emph{arXiv preprint arXiv:1712.00948}, 2017{\natexlab{b}}.

\bibitem[Li et~al.(2020{\natexlab{a}})Li, Florensa, Clavera, and
  Abbeel]{Li2020SubpolicyAF}
Alexander Li, Carlos Florensa, Ignasi Clavera, and P.~Abbeel.
\newblock Sub-policy adaptation for hierarchical reinforcement learning.
\newblock \emph{ArXiv}, abs/1906.05862, 2020{\natexlab{a}}.

\bibitem[Li et~al.(2020{\natexlab{b}})Li, Pinto, and
  Abbeel]{NEURIPS2020_57e5cb96}
Alexander Li, Lerrel Pinto, and Pieter Abbeel.
\newblock Generalized hindsight for reinforcement learning.
\newblock In H.~Larochelle, M.~Ranzato, R.~Hadsell, M.F. Balcan, and H.~Lin,
  editors, \emph{Advances in Neural Information Processing Systems}, volume~33,
  pages 7754--7767. Curran Associates, Inc., 2020{\natexlab{b}}.
\newblock URL
  \url{https://proceedings.neurips.cc/paper/2020/file/57e5cb96e22546001f1d6520ff11d9ba-Paper.pdf}.

\bibitem[Li et~al.(2021)Li, Gupta, Reddy, Pong, Zhou, Yu, and
  Levine]{Li2021MURALMU}
Kevin Li, Abhishek Gupta, Ashwin Reddy, Vitchyr~H. Pong, Aurick Zhou, Justin
  Yu, and Sergey Levine.
\newblock Mural: Meta-learning uncertainty-aware rewards for outcome-driven
  reinforcement learning.
\newblock \emph{ArXiv}, abs/2107.07184, 2021.

\bibitem[Li and Zhang(2018)]{Li2018AnOO}
Siyuan Li and Chongjie Zhang.
\newblock An optimal online method of selecting source policies for
  reinforcement learning.
\newblock In \emph{AAAI}, 2018.

\bibitem[Li and Liang(2021)]{li2021prefix}
Xiang~Lisa Li and Percy Liang.
\newblock Prefix-tuning: Optimizing continuous prompts for generation.
\newblock \emph{arXiv preprint arXiv:2101.00190}, 2021.

\bibitem[Lin et~al.(2014)Lin, Maire, Belongie, Hays, Perona, Ramanan,
  Doll{\'a}r, and Zitnick]{Lin2014MicrosoftCC}
Tsung-Yi Lin, Michael Maire, Serge~J. Belongie, James Hays, Pietro Perona, Deva
  Ramanan, Piotr Doll{\'a}r, and C.~Lawrence Zitnick.
\newblock Microsoft coco: Common objects in context.
\newblock \emph{ArXiv}, abs/1405.0312, 2014.

\bibitem[Liu et~al.(2022)Liu, Lever, Wang, Merel, Eslami, Hennes, Czarnecki,
  Tassa, Omidshafiei, Abdolmaleki, et~al.]{liu2021motor}
Siqi Liu, Guy Lever, Zhe Wang, Josh Merel, SM~Ali Eslami, Daniel Hennes,
  Wojciech~M Czarnecki, Yuval Tassa, Shayegan Omidshafiei, Abbas Abdolmaleki,
  et~al.
\newblock From motor control to team play in simulated humanoid football.
\newblock \emph{Science Robotics}, 7\penalty0 (69):\penalty0 eabo0235, 2022.

\bibitem[Liu et~al.(2021)Liu, Zheng, Du, Ding, Qian, Yang, and
  Tang]{Liu2021GPTUT}
Xiao Liu, Yanan Zheng, Zhengxiao Du, Ming Ding, Yujie Qian, Zhilin Yang, and
  Jie Tang.
\newblock Gpt understands, too.
\newblock \emph{ArXiv}, abs/2103.10385, 2021.
\newblock URL \url{https://api.semanticscholar.org/CorpusID:232269696}.

\bibitem[Ljung(1998)]{ljung1998system}
Lennart Ljung.
\newblock System identification.
\newblock In \emph{Signal analysis and prediction}, pages 163--173. Springer,
  1998.

\bibitem[Lu et~al.(2022)Lu, Hausman, Chebotar, Yan, Jang, Herzog, Xiao, Irpan,
  Khansari, Kalashnikov, et~al.]{lu2022aw}
Yao Lu, Karol Hausman, Yevgen Chebotar, Mengyuan Yan, Eric Jang, Alexander
  Herzog, Ted Xiao, Alex Irpan, Mohi Khansari, Dmitry Kalashnikov, et~al.
\newblock Aw-opt: Learning robotic skills with imitation and reinforcement at
  scale.
\newblock In \emph{Conference on Robot Learning}, pages 1078--1088. PMLR, 2022.

\bibitem[Luketina et~al.(2019)Luketina, Nardelli, Farquhar, Foerster, Andreas,
  Grefenstette, Whiteson, and Rockt{\"a}schel]{luketina2019survey}
Jelena Luketina, Nantas Nardelli, Gregory Farquhar, Jakob Foerster, Jacob
  Andreas, Edward Grefenstette, Shimon Whiteson, and Tim Rockt{\"a}schel.
\newblock A survey of reinforcement learning informed by natural language.
\newblock \emph{arXiv preprint arXiv:1906.03926}, 2019.

\bibitem[Lynch and Sermanet(2020)]{Lynch2020LanguageCI}
Corey Lynch and Pierre Sermanet.
\newblock Language conditioned imitation learning over unstructured data.
\newblock \emph{Robotics: Science and Systems XVII}, 2020.

\bibitem[Lynch et~al.(2019)Lynch, Khansari, Xiao, Kumar, Tompson, Levine, and
  Sermanet]{Lynch2019LearningLP}
Corey Lynch, Mohi Khansari, Ted Xiao, Vikash Kumar, Jonathan Tompson, Sergey
  Levine, and Pierre Sermanet.
\newblock Learning latent plans from play.
\newblock In \emph{CoRL}, 2019.

\bibitem[Lynch et~al.(2022)Lynch, Wahid, Tompson, Ding, Betker, Baruch,
  Armstrong, and Florence]{lynch2022interactive}
Corey Lynch, Ayzaan Wahid, Jonathan Tompson, Tianli Ding, James Betker, Robert
  Baruch, Travis Armstrong, and Pete Florence.
\newblock Interactive language: Talking to robots in real time.
\newblock \emph{arXiv preprint arXiv:2210.06407}, 2022.

\bibitem[Ma et~al.(2022)Ma, Sodhani, Jayaraman, Bastani, Kumar, and
  Zhang]{ma2022vip}
Yecheng~Jason Ma, Shagun Sodhani, Dinesh Jayaraman, Osbert Bastani, Vikash
  Kumar, and Amy Zhang.
\newblock Vip: Towards universal visual reward and representation via
  value-implicit pre-training.
\newblock \emph{arXiv preprint arXiv:2210.00030}, 2022.

\bibitem[Ma et~al.(2023)Ma, Liang, Som, Kumar, Zhang, Bastani, and
  Jayaraman]{ma2023liv}
Yecheng~Jason Ma, William Liang, Vaidehi Som, Vikash Kumar, Amy Zhang, Osbert
  Bastani, and Dinesh Jayaraman.
\newblock Liv: Language-image representations and rewards for robotic control.
\newblock \emph{arXiv preprint arXiv:2306.00958}, 2023.

\bibitem[Machado et~al.(2020)Machado, Bellemare, and Bowling]{machado2020count}
Marlos~C Machado, Marc~G Bellemare, and Michael Bowling.
\newblock Count-based exploration with the successor representation.
\newblock In \emph{Proceedings of the AAAI Conference on Artificial
  Intelligence}, volume~34, pages 5125--5133, 2020.

\bibitem[Mahadevan(1996)]{mahadevan1996average}
Sridhar Mahadevan.
\newblock Average reward reinforcement learning: Foundations, algorithms, and
  empirical results.
\newblock \emph{Machine learning}, 22:\penalty0 159--195, 1996.

\bibitem[Mahmoudieh et~al.(2022)Mahmoudieh, Pathak, and
  Darrell]{mahmoudieh2022zero}
Parsa Mahmoudieh, Deepak Pathak, and Trevor Darrell.
\newblock Zero-shot reward specification via grounded natural language.
\newblock In \emph{International Conference on Machine Learning}, pages
  14743--14752. PMLR, 2022.

\bibitem[{Manolis Savva} et~al.(2019){Manolis Savva}, {Abhishek Kadian},
  {Oleksandr Maksymets}, Zhao, Wijmans, Jain, Straub, Liu, Koltun, Malik,
  Parikh, and Batra]{habitat19iccv}
{Manolis Savva}, {Abhishek Kadian}, {Oleksandr Maksymets}, Yili Zhao, Erik
  Wijmans, Bhavana Jain, Julian Straub, Jia Liu, Vladlen Koltun, Jitendra
  Malik, Devi Parikh, and Dhruv Batra.
\newblock Habitat: {A} {P}latform for {E}mbodied {AI} {R}esearch.
\newblock In \emph{Proceedings of the IEEE/CVF International Conference on
  Computer Vision (ICCV)}, 2019.

\bibitem[Matena and Raffel(2021)]{Matena2021MergingMW}
Michael Matena and Colin Raffel.
\newblock Merging models with fisher-weighted averaging.
\newblock \emph{ArXiv}, abs/2111.09832, 2021.

\bibitem[Matthews et~al.(2022)Matthews, Samvelyan, Parker-Holder, Grefenstette,
  and Rockt{\"a}schel]{matthews2022skillhack}
Michael Matthews, Mikayel Samvelyan, Jack Parker-Holder, Edward Grefenstette,
  and Tim Rockt{\"a}schel.
\newblock Skillhack: A benchmark for skill transfer in open-ended reinforcement
  learning.
\newblock In \emph{ICLR Workshop on Agent Learning in Open-Endedness}, 2022.

\bibitem[Mayne(2014)]{mayne2014model}
David~Q Mayne.
\newblock Model predictive control: Recent developments and future promise.
\newblock \emph{Automatica}, 50\penalty0 (12):\penalty0 2967--2986, 2014.

\bibitem[Mendonca et~al.(2019)Mendonca, Gupta, Kralev, Abbeel, Levine, and
  Finn]{mendonca2019guided}
Russell Mendonca, Abhishek Gupta, Rosen Kralev, Pieter Abbeel, Sergey Levine,
  and Chelsea Finn.
\newblock Guided meta-policy search.
\newblock \emph{arXiv preprint arXiv:1904.00956}, 2019.

\bibitem[Merel et~al.(2017)Merel, Tassa, TB, Srinivasan, Lemmon, Wang, Wayne,
  and Heess]{merel2017learning}
Josh Merel, Yuval Tassa, Dhruva TB, Sriram Srinivasan, Jay Lemmon, Ziyu Wang,
  Greg Wayne, and Nicolas Heess.
\newblock Learning human behaviors from motion capture by adversarial
  imitation.
\newblock \emph{arXiv preprint arXiv:1707.02201}, 2017.

\bibitem[Merel et~al.(2018)Merel, Hasenclever, Galashov, Ahuja, Pham, Wayne,
  Teh, and Heess]{merel2018neural}
Josh Merel, Leonard Hasenclever, Alexandre Galashov, Arun Ahuja, Vu~Pham, Greg
  Wayne, Yee~Whye Teh, and Nicolas Heess.
\newblock Neural probabilistic motor primitives for humanoid control.
\newblock \emph{arXiv preprint arXiv:1811.11711}, 2018.

\bibitem[Merity et~al.(2016)Merity, Xiong, Bradbury, and
  Socher]{merity2016pointer}
Stephen Merity, Caiming Xiong, James Bradbury, and Richard Socher.
\newblock Pointer sentinel mixture models, 2016.

\bibitem[Mesnil et~al.(2012)Mesnil, Dauphin, Glorot, Rifai, Bengio, Goodfellow,
  Lavoie, Muller, Desjardins, Warde-Farley, et~al.]{mesnil2012unsupervised}
Gr{\'e}goire Mesnil, Yann Dauphin, Xavier Glorot, Salah Rifai, Yoshua Bengio,
  Ian Goodfellow, Erick Lavoie, Xavier Muller, Guillaume Desjardins, David
  Warde-Farley, et~al.
\newblock Unsupervised and transfer learning challenge: a deep learning
  approach.
\newblock In \emph{Proceedings of ICML Workshop on Unsupervised and Transfer
  Learning}, pages 97--110. JMLR Workshop and Conference Proceedings, 2012.

\bibitem[Micheli et~al.(2022)Micheli, Alonso, and
  Fleuret]{Micheli2022TransformersAS}
Vincent Micheli, Eloi Alonso, and Franccois Fleuret.
\newblock Transformers are sample efficient world models.
\newblock \emph{ArXiv}, abs/2209.00588, 2022.

\bibitem[Michie and Chambers(1968)]{michie1968boxes}
Donald Michie and Roger~A Chambers.
\newblock Boxes: An experiment in adaptive control.
\newblock \emph{Machine intelligence}, 2\penalty0 (2):\penalty0 137--152, 1968.

\bibitem[Michod(2000)]{michod2000darwinian}
Richard~E Michod.
\newblock \emph{Darwinian dynamics: evolutionary transitions in fitness and
  individuality}.
\newblock Princeton University Press, 2000.

\bibitem[Min et~al.(2021)Min, Lewis, Zettlemoyer, and
  Hajishirzi]{min2021metaicl}
Sewon Min, Mike Lewis, Luke Zettlemoyer, and Hannaneh Hajishirzi.
\newblock Metaicl: Learning to learn in context.
\newblock \emph{arXiv preprint arXiv:2110.15943}, 2021.

\bibitem[Mirhoseini et~al.(2020)Mirhoseini, Goldie, Yazgan, Jiang, Songhori,
  Wang, Lee, Johnson, Pathak, Bae, Nazi, Pak, Tong, Srinivasa, Hang, Tuncer,
  Babu, Le, Laudon, Ho, Carpenter, and Dean]{Mirhoseini2020ChipPW}
Azalia Mirhoseini, Anna Goldie, M.~Yazgan, J.~Jiang, Ebrahim~M. Songhori,
  S.~Wang, Young-Joon Lee, Eric Johnson, Omkar Pathak, Sungmin Bae, Azade Nazi,
  Jiwoo Pak, Andy Tong, Kavya Srinivasa, W.~Hang, Emre Tuncer, Anand Babu,
  Quoc~V. Le, J.~Laudon, Richard Ho, Roger Carpenter, and J.~Dean.
\newblock Chip placement with deep reinforcement learning.
\newblock \emph{ArXiv}, abs/2004.10746, 2020.

\bibitem[Mishra et~al.(2017)Mishra, Rohaninejad, Chen, and
  Abbeel]{mishra2017simple}
Nikhil Mishra, Mostafa Rohaninejad, Xi~Chen, and Pieter Abbeel.
\newblock A simple neural attentive meta-learner.
\newblock \emph{arXiv preprint arXiv:1707.03141}, 2017.

\bibitem[Mitchell et~al.(2021)Mitchell, Rafailov, Peng, Levine, and
  Finn]{mitchell2021offline}
Eric Mitchell, Rafael Rafailov, Xue~Bin Peng, Sergey Levine, and Chelsea Finn.
\newblock Offline meta-reinforcement learning with advantage weighting.
\newblock In \emph{International Conference on Machine Learning}, pages
  7780--7791. PMLR, 2021.

\bibitem[Molchanov et~al.(2019)Molchanov, Chen, H{\"o}nig, Preiss, Ayanian, and
  Sukhatme]{Molchanov2019SimtoMultiRealTO}
A.~Molchanov, Tao Chen, Wolfgang H{\"o}nig, James~A. Preiss, Nora Ayanian, and
  G.~Sukhatme.
\newblock Sim-to-(multi)-real: Transfer of low-level robust control policies to
  multiple quadrotors.
\newblock \emph{2019 IEEE/RSJ International Conference on Intelligent Robots
  and Systems (IROS)}, pages 59--66, 2019.

\bibitem[Morari and Lee(1999)]{morari1999model}
Manfred Morari and Jay~H Lee.
\newblock Model predictive control: past, present and future.
\newblock \emph{Computers \& chemical engineering}, 23\penalty0 (4-5):\penalty0
  667--682, 1999.

\bibitem[Mu et~al.(2021)Mu, Ling, Xiang, Yang, Li, Tao, Huang, Jia, and
  Su]{mu2021maniskill}
Tongzhou Mu, Zhan Ling, Fanbo Xiang, Derek~Cathera Yang, Xuanlin Li, Stone Tao,
  Zhiao Huang, Zhiwei Jia, and Hao Su.
\newblock Maniskill: Generalizable manipulation skill benchmark with
  large-scale demonstrations.
\newblock In \emph{Thirty-fifth Conference on Neural Information Processing
  Systems Datasets and Benchmarks Track (Round 2)}, 2021.

\bibitem[M{\"u}ller et~al.(2018)M{\"u}ller, Dosovitskiy, Ghanem, and
  Koltun]{muller2018driving}
Matthias M{\"u}ller, Alexey Dosovitskiy, Bernard Ghanem, and Vladlen Koltun.
\newblock Driving policy transfer via modularity and abstraction.
\newblock \emph{arXiv preprint arXiv:1804.09364}, 2018.

\bibitem[Muller-Brockhausen et~al.(2021)Muller-Brockhausen, Preuss, and
  Plaat]{muller2021procedural}
Matthias Muller-Brockhausen, Mike Preuss, and Aske Plaat.
\newblock Procedural content generation: Better benchmarks for transfer
  reinforcement learning.
\newblock In \emph{2021 IEEE Conference on games (CoG)}, pages 01--08. IEEE,
  2021.

\bibitem[Mutti et~al.(2020)Mutti, Pratissoli, and Restelli]{Mutti2020APG}
Mirco Mutti, Lorenzo Pratissoli, and Marcello Restelli.
\newblock A policy gradient method for task-agnostic exploration.
\newblock \emph{ArXiv}, abs/2007.04640, 2020.

\bibitem[Nachum et~al.(2018)Nachum, Gu, Lee, and Levine]{nachum2018data}
Ofir Nachum, Shixiang Gu, Honglak Lee, and Sergey Levine.
\newblock Data-efficient hierarchical reinforcement learning.
\newblock \emph{arXiv preprint arXiv:1805.08296}, 2018.

\bibitem[Nachum et~al.(2019{\natexlab{a}})Nachum, Ahn, Ponte, Gu, and
  Kumar]{nachum2019multi}
Ofir Nachum, Michael Ahn, Hugo Ponte, Shixiang Gu, and Vikash Kumar.
\newblock Multi-agent manipulation via locomotion using hierarchical sim2real.
\newblock \emph{arXiv preprint arXiv:1908.05224}, 2019{\natexlab{a}}.

\bibitem[Nachum et~al.(2019{\natexlab{b}})Nachum, Tang, Lu, Gu, Lee, and
  Levine]{nachum2019why}
Ofir Nachum, Haoran Tang, Xingyu Lu, Shixiang Gu, Honglak Lee, and Sergey
  Levine.
\newblock Why does hierarchy (sometimes) work so well in reinforcement
  learning?
\newblock \emph{CoRR}, abs/1909.10618, 2019{\natexlab{b}}.
\newblock URL \url{http://arxiv.org/abs/1909.10618}.

\bibitem[Nagabandi et~al.(2018{\natexlab{a}})Nagabandi, Clavera, Liu, Fearing,
  Abbeel, Levine, and Finn]{nagabandi2018learning}
Anusha Nagabandi, Ignasi Clavera, Simin Liu, Ronald~S Fearing, Pieter Abbeel,
  Sergey Levine, and Chelsea Finn.
\newblock Learning to adapt in dynamic, real-world environments through
  meta-reinforcement learning.
\newblock \emph{arXiv preprint arXiv:1803.11347}, 2018{\natexlab{a}}.

\bibitem[Nagabandi et~al.(2018{\natexlab{b}})Nagabandi, Finn, and
  Levine]{nagabandi2018deep}
Anusha Nagabandi, Chelsea Finn, and Sergey Levine.
\newblock Deep online learning via meta-learning: Continual adaptation for
  model-based rl.
\newblock \emph{arXiv preprint arXiv:1812.07671}, 2018{\natexlab{b}}.

\bibitem[Nagabandi et~al.(2018{\natexlab{c}})Nagabandi, Kahn, Fearing, and
  Levine]{nagabandi2018neural}
Anusha Nagabandi, Gregory Kahn, Ronald~S Fearing, and Sergey Levine.
\newblock Neural network dynamics for model-based deep reinforcement learning
  with model-free fine-tuning.
\newblock In \emph{2018 IEEE International Conference on Robotics and
  Automation (ICRA)}, pages 7559--7566. IEEE, 2018{\natexlab{c}}.

\bibitem[Nair et~al.(2018{\natexlab{a}})Nair, McGrew, Andrychowicz, Zaremba,
  and Abbeel]{nair2018overcoming}
Ashvin Nair, Bob McGrew, Marcin Andrychowicz, Wojciech Zaremba, and Pieter
  Abbeel.
\newblock Overcoming exploration in reinforcement learning with demonstrations.
\newblock In \emph{2018 IEEE international conference on robotics and
  automation (ICRA)}, pages 6292--6299. IEEE, 2018{\natexlab{a}}.

\bibitem[Nair et~al.(2020{\natexlab{a}})Nair, Dalal, Gupta, and
  Levine]{nair2020accelerating}
Ashvin Nair, Murtaza Dalal, Abhishek Gupta, and Sergey Levine.
\newblock Accelerating online reinforcement learning with offline datasets.
\newblock \emph{arXiv preprint arXiv:2006.09359}, 2020{\natexlab{a}}.

\bibitem[Nair et~al.(2018{\natexlab{b}})Nair, Pong, Dalal, Bahl, Lin, and
  Levine]{nair2018visual}
Ashvin~V Nair, Vitchyr Pong, Murtaza Dalal, Shikhar Bahl, Steven Lin, and
  Sergey Levine.
\newblock Visual reinforcement learning with imagined goals.
\newblock \emph{Advances in neural information processing systems}, 31,
  2018{\natexlab{b}}.

\bibitem[Nair and Finn(2019)]{nair2019hierarchical}
Suraj Nair and Chelsea Finn.
\newblock Hierarchical foresight: Self-supervised learning of long-horizon
  tasks via visual subgoal generation.
\newblock \emph{arXiv preprint arXiv:1909.05829}, 2019.

\bibitem[Nair et~al.(2020{\natexlab{b}})Nair, Savarese, and Finn]{nair2020goal}
Suraj Nair, Silvio Savarese, and Chelsea Finn.
\newblock Goal-aware prediction: Learning to model what matters.
\newblock In \emph{International Conference on Machine Learning}, pages
  7207--7219. PMLR, 2020{\natexlab{b}}.

\bibitem[Nair et~al.(2021)Nair, Mitchell, Chen, Ichter, Savarese, and
  Finn]{nair2021learning}
Suraj Nair, Eric Mitchell, Kevin Chen, Brian Ichter, Silvio Savarese, and
  Chelsea Finn.
\newblock Learning language-conditioned robot behavior from offline data and
  crowd-sourced annotation.
\newblock \emph{arXiv preprint arXiv:2109.01115}, 2021.

\bibitem[Nair et~al.(2022)Nair, Rajeswaran, Kumar, Finn, and
  Gupta]{Nair2022R3MAU}
Suraj Nair, Aravind Rajeswaran, Vikash Kumar, Chelsea Finn, and Abhi Gupta.
\newblock R3m: A universal visual representation for robot manipulation.
\newblock \emph{ArXiv}, abs/2203.12601, 2022.

\bibitem[Nangue~Tasse et~al.(2020)Nangue~Tasse, James, and
  Rosman]{nangue2020boolean}
Geraud Nangue~Tasse, Steven James, and Benjamin Rosman.
\newblock A boolean task algebra for reinforcement learning.
\newblock \emph{Advances in Neural Information Processing Systems},
  33:\penalty0 9497--9507, 2020.

\bibitem[Narvekar et~al.(2020)Narvekar, Peng, Leonetti, Sinapov, Taylor, and
  Stone]{narvekar2020curriculum}
Sanmit Narvekar, Bei Peng, Matteo Leonetti, Jivko Sinapov, Matthew~E Taylor,
  and Peter Stone.
\newblock Curriculum learning for reinforcement learning domains: A framework
  and survey.
\newblock \emph{arXiv preprint arXiv:2003.04960}, 2020.

\bibitem[Nelles(2001)]{nelles2001nonlinear}
Oliver Nelles.
\newblock Nonlinear dynamic system identification.
\newblock In \emph{Nonlinear System Identification}, pages 547--577. Springer,
  2001.

\bibitem[Nematollahi et~al.(2020)Nematollahi, Mees, Hermann, and
  Burgard]{nematollahi2020hindsight}
Iman Nematollahi, Oier Mees, Lukas Hermann, and Wolfram Burgard.
\newblock Hindsight for foresight: Unsupervised structured dynamics models from
  physical interaction.
\newblock In \emph{2020 IEEE/RSJ International Conference on Intelligent Robots
  and Systems (IROS)}, pages 5319--5326. IEEE, 2020.

\bibitem[Ng et~al.(1999)Ng, Harada, and Russell]{Ng1999PolicyIU}
A.~Ng, D.~Harada, and Stuart~J. Russell.
\newblock Policy invariance under reward transformations: Theory and
  application to reward shaping.
\newblock In \emph{ICML}, 1999.

\bibitem[Ng and Russell(2000)]{ng2000inverse}
Andrew~Y. Ng and Stuart~J. Russell.
\newblock Algorithms for inverse reinforcement learning.
\newblock In \emph{Proceedings of the Seventeenth International Conference on
  Machine Learning}, ICML '00, page 663–670, San Francisco, CA, USA, 2000.
  Morgan Kaufmann Publishers Inc.
\newblock ISBN 1558607072.

\bibitem[Nguyen and Grover(2022)]{nguyen2022transformer}
Tung Nguyen and Aditya Grover.
\newblock Transformer neural processes: Uncertainty-aware meta learning via
  sequence modeling.
\newblock \emph{arXiv preprint arXiv:2207.04179}, 2022.

\bibitem[Nichol et~al.(2018)Nichol, Achiam, and Schulman]{Nichol2018OnFM}
Alex Nichol, Joshua Achiam, and John Schulman.
\newblock On first-order meta-learning algorithms.
\newblock \emph{ArXiv}, abs/1803.02999, 2018.

\bibitem[Nikishin et~al.(2022)Nikishin, Schwarzer, D'Oro, Bacon, and
  Courville]{Nikishin2022ThePB}
Evgenii Nikishin, Max Schwarzer, Pierluca D'Oro, Pierre-Luc Bacon, and Aaron
  Courville.
\newblock The primacy bias in deep reinforcement learning.
\newblock In \emph{ICML}, 2022.

\bibitem[Nvidia(2023)]{omniverse}
Nvidia, 2023.
\newblock URL \url{https://docs.omniverse.nvidia.com/index.html}.

\bibitem[Oh et~al.(2017)Oh, Singh, Lee, and Kohli]{Oh2017ZeroShotTG}
Junhyuk Oh, Satinder Singh, Honglak Lee, and P.~Kohli.
\newblock Zero-shot task generalization with multi-task deep reinforcement
  learning.
\newblock \emph{ArXiv}, abs/1706.05064, 2017.

\bibitem[OpenAI et~al.(2019{\natexlab{a}})OpenAI, :, Berner, Brockman, Chan,
  Cheung, Dębiak, Dennison, Farhi, Fischer, Hashme, Hesse, Józefowicz, Gray,
  Olsson, Pachocki, Petrov, de~Oliveira~Pinto, Raiman, Salimans, Schlatter,
  Schneider, Sidor, Sutskever, Tang, Wolski, and Zhang]{openai2019dota}
OpenAI, :, Christopher Berner, Greg Brockman, Brooke Chan, Vicki Cheung,
  Przemysław Dębiak, Christy Dennison, David Farhi, Quirin Fischer, Shariq
  Hashme, Chris Hesse, Rafal Józefowicz, Scott Gray, Catherine Olsson, Jakub
  Pachocki, Michael Petrov, Henrique~Pondé de~Oliveira~Pinto, Jonathan Raiman,
  Tim Salimans, Jeremy Schlatter, Jonas Schneider, Szymon Sidor, Ilya
  Sutskever, Jie Tang, Filip Wolski, and Susan Zhang.
\newblock Dota 2 with large scale deep reinforcement learning,
  2019{\natexlab{a}}.

\bibitem[OpenAI et~al.(2019{\natexlab{b}})OpenAI, Akkaya, Andrychowicz,
  Chociej, Litwin, McGrew, Petron, Paino, Plappert, Powell, Ribas, Schneider,
  Tezak, Tworek, Welinder, Weng, Yuan, Zaremba, and Zhang]{openai2019solving}
OpenAI, Ilge Akkaya, Marcin Andrychowicz, Maciek Chociej, Mateusz Litwin, Bob
  McGrew, Arthur Petron, Alex Paino, Matthias Plappert, Glenn Powell, Raphael
  Ribas, Jonas Schneider, Nikolas Tezak, Jerry Tworek, Peter Welinder, Lilian
  Weng, Qiming Yuan, Wojciech Zaremba, and Lei Zhang.
\newblock Solving {R}ubik's cube with a robot hand.
\newblock \emph{CoRR}, abs/1910.07113, 2019{\natexlab{b}}.
\newblock URL \url{http://arxiv.org/abs/1910.07113}.

\bibitem[Packer et~al.(2018)Packer, Gao, Kos, Kr{\"a}henb{\"u}hl, Koltun, and
  Song]{packer2018assessing}
Charles Packer, Katelyn Gao, Jernej Kos, Philipp Kr{\"a}henb{\"u}hl, Vladlen
  Koltun, and Dawn Song.
\newblock Assessing generalization in deep reinforcement learning.
\newblock \emph{arXiv preprint arXiv:1810.12282}, 2018.

\bibitem[Padalkar et~al.(2023)Padalkar, Pooley, Jain, Bewley, Herzog, Irpan,
  Khazatsky, Rai, Singh, Brohan, Raffin, Wahid, Burgess-Limerick, Kim,
  Sch{\"o}lkopf, Ichter, Lu, Xu, Finn, Xu, Chi, Huang, Chan, Pan, Fu, Devin,
  Driess, Pathak, Shah, B{\"u}chler, Kalashnikov, Sadigh, Johns, Ceola, Xia,
  Stulp, Zhou, Sukhatme, Salhotra, Yan, Schiavi, Su, Fang, Shi, Amor,
  Christensen, Furuta, Walke, Fang, Mordatch, Radosavovic, Leal, Liang, Kim,
  Schneider, Hsu, Bohg, Bingham, Wu, Wu, Luo, Gu, Tan, Oh, Malik, Tompson,
  Yang, Lim, Silv{\'e}rio, Han, Rao, Pertsch, Hausman, Go, Gopalakrishnan,
  Goldberg, Byrne, Oslund, Kawaharazuka, Zhang, Majd, Rana, Srinivasan, Chen,
  Pinto, Tan, Ott, Lee, Tomizuka, Du, Ahn, Zhang, Ding, Srirama, Sharma, Kim,
  Kanazawa, Hansen, Heess, Joshi, Suenderhauf, Palo, Shafiullah, Mees, Kroemer,
  Sanketi, Wohlhart, Xu, Sermanet, Sundaresan, Vuong, Rafailov, Tian, Doshi,
  Mendonca, Shah, Hoque, Julian, Bustamante, Kirmani, Levine, Moore, Bahl,
  Dass, Song, Xu, Haldar, Adebola, Guist, Nasiriany, Schaal, Welker, Tian,
  Dasari, Belkhale, Osa, Harada, Matsushima, Xiao, Yu, Ding, Davchev, Zhao,
  Armstrong, Darrell, Jain, Vanhoucke, Zhan, Zhou, Burgard, Chen, Wang, Zhu,
  Li, Lu, Chebotar, Zhou, Zhu, Xu, Wang, Bisk, Cho, Lee, Cui, hua Wu, Tang,
  Zhu, Li, Iwasawa, Matsuo, Xu, and Cui]{Padalkar2023OpenXR}
Abhishek Padalkar, Acorn Pooley, Ajinkya Jain, Alex Bewley, Alex Herzog, Alex
  Irpan, Alexander Khazatsky, Anant Rai, Anika Singh, Anthony Brohan, Antonin
  Raffin, Ayzaan Wahid, Ben Burgess-Limerick, Beomjoon Kim, Bernhard
  Sch{\"o}lkopf, Brian Ichter, Cewu Lu, Charles Xu, Chelsea Finn, Chenfeng Xu,
  Cheng Chi, Chenguang Huang, Christine Chan, Chuer Pan, Chuyuan Fu, Coline
  Devin, Danny Driess, Deepak Pathak, Dhruv Shah, Dieter B{\"u}chler, Dmitry
  Kalashnikov, Dorsa Sadigh, Edward Johns, Federico Ceola, Fei Xia, Freek
  Stulp, Gaoyue Zhou, Gaurav~S. Sukhatme, Gautam Salhotra, Ge~Yan, Giulio
  Schiavi, Hao Su, Haoshu Fang, Haochen Shi, Heni~Ben Amor, Henrik~I
  Christensen, Hiroki Furuta, Homer Walke, Hongjie Fang, Igor Mordatch, Ilija
  Radosavovic, Isabel Leal, Jacky Liang, Jaehyung Kim, Jan Schneider, Jasmine
  Hsu, Jeannette Bohg, Jeff Bingham, Jiajun Wu, Jialin Wu, Jianlan Luo, Jiayuan
  Gu, Jie Tan, Jihoon Oh, Jitendra Malik, Jonathan Tompson, Jonathan Yang,
  Joseph~J. Lim, Jo{\~a}o Silv{\'e}rio, Junhyek Han, Kanishka Rao, Karl
  Pertsch, Karol Hausman, Keegan Go, Keerthana Gopalakrishnan, Ken Goldberg,
  Kendra Byrne, Kenneth Oslund, Kento Kawaharazuka, Kevin Zhang, Keyvan Majd,
  Krishan Rana, Krishna~Parasuram Srinivasan, Lawrence~Yunliang Chen, Lerrel
  Pinto, Liam Tan, Lionel Ott, Lisa Lee, Masayoshi Tomizuka, Maximilian Du,
  Michael Ahn, Mingtong Zhang, Mingyu Ding, Mohan~Kumar Srirama, Mohit Sharma,
  Moo~Jin Kim, Naoaki Kanazawa, Nicklas Hansen, Nicolas Manfred~Otto Heess,
  Nikhil~J. Joshi, Niko Suenderhauf, Norman~Di Palo, Nur Muhammad~Mahi
  Shafiullah, Oier Mees, Oliver Kroemer, Pannag~R. Sanketi, Paul Wohlhart, Peng
  Xu, Pierre Sermanet, Priya Sundaresan, Quan~Ho Vuong, Rafael Rafailov, Ran
  Tian, Ria Doshi, Russell Mendonca, Rutav Shah, Ryan Hoque, Ryan~C. Julian,
  Samuel Bustamante, Sean Kirmani, Sergey Levine, Sherry Moore, Shikhar Bahl,
  Shivin Dass, Shuran Song, Sichun Xu, Siddhant Haldar, S.~O. Adebola, Simon
  Guist, Soroush Nasiriany, Stefan Schaal, Stefan Welker, Stephen Tian, Sudeep
  Dasari, Suneel Belkhale, Takayuki Osa, Tatsuya Harada, Tatsuya Matsushima,
  Ted Xiao, Tianhe Yu, Tianli Ding, Todor Davchev, Tony Zhao, Travis Armstrong,
  Trevor Darrell, Vidhi Jain, Vincent Vanhoucke, Wei Zhan, Wenxuan Zhou,
  Wolfram Burgard, Xi~Chen, Xiaolong Wang, Xinghao Zhu, Xuanlin Li, Yao Lu,
  Yevgen Chebotar, Yifan Zhou, Yifeng Zhu, Ying Xu, Yixuan Wang, Yonatan Bisk,
  Yoonyoung Cho, Youngwoon Lee, Yuchen Cui, Yueh hua Wu, Yujin Tang, Yuke Zhu,
  Yunzhu Li, Yusuke Iwasawa, Yutaka Matsuo, Zhuo Xu, and Zichen~Jeff Cui.
\newblock Open x-embodiment: Robotic learning datasets and rt-x models.
\newblock \emph{ArXiv}, abs/2310.08864, 2023.
\newblock URL \url{https://api.semanticscholar.org/CorpusID:263626099}.

\bibitem[Pari et~al.(2021)Pari, Shafiullah, Arunachalam, and
  Pinto]{Pari2021TheSE}
Jyothish Pari, Nur~Muhammad Shafiullah, Sridhar~Pandian Arunachalam, and Lerrel
  Pinto.
\newblock The surprising effectiveness of representation learning for visual
  imitation.
\newblock \emph{arXiv preprint arXiv:2112.01511}, 2021.

\bibitem[Parisi et~al.(2019)Parisi, Kemker, Part, Kanan, and
  Wermter]{parisi2019continual}
German~I Parisi, Ronald Kemker, Jose~L Part, Christopher Kanan, and Stefan
  Wermter.
\newblock Continual lifelong learning with neural networks: A review.
\newblock \emph{Neural Networks}, 113:\penalty0 54--71, 2019.

\bibitem[Parisotto et~al.(2015)Parisotto, Ba, and
  Salakhutdinov]{parisotto2015actor}
Emilio Parisotto, Jimmy~Lei Ba, and Ruslan Salakhutdinov.
\newblock Actor-mimic: Deep multitask and transfer reinforcement learning.
\newblock \emph{arXiv preprint arXiv:1511.06342}, 2015.

\bibitem[Parisotto et~al.(2020)Parisotto, Song, Rae, Pascanu, Gulcehre,
  Jayakumar, Jaderberg, Kaufman, Clark, Noury,
  et~al.]{parisotto2020stabilizing}
Emilio Parisotto, Francis Song, Jack Rae, Razvan Pascanu, Caglar Gulcehre,
  Siddhant Jayakumar, Max Jaderberg, Raphael~Lopez Kaufman, Aidan Clark, Seb
  Noury, et~al.
\newblock Stabilizing transformers for reinforcement learning.
\newblock In \emph{International conference on machine learning}, pages
  7487--7498. PMLR, 2020.

\bibitem[Paxton et~al.(2019)Paxton, Barnoy, Katyal, Arora, and
  Hager]{paxton2019visual}
Chris Paxton, Yotam Barnoy, Kapil Katyal, Raman Arora, and Gregory~D Hager.
\newblock Visual robot task planning.
\newblock In \emph{2019 international conference on robotics and automation
  (ICRA)}, pages 8832--8838. IEEE, 2019.

\bibitem[Peng et~al.(2018{\natexlab{a}})Peng, Abbeel, Levine, and van~de
  Panne]{peng2018deepmimic}
Xue~Bin Peng, Pieter Abbeel, Sergey Levine, and Michiel van~de Panne.
\newblock {DeepMimic: Example-guided deep reinforcement learning of
  physics-based character skills}.
\newblock \emph{ACM Transactions on Graphics (TOG)}, 37\penalty0 (4):\penalty0
  143, 2018{\natexlab{a}}.

\bibitem[Peng et~al.(2018{\natexlab{b}})Peng, Kanazawa, Malik, Abbeel, and
  Levine]{peng2018sfv}
Xue~Bin Peng, Angjoo Kanazawa, Jitendra Malik, Pieter Abbeel, and Sergey
  Levine.
\newblock Sfv: Reinforcement learning of physical skills from videos.
\newblock \emph{ACM Transactions On Graphics (TOG)}, 37\penalty0 (6):\penalty0
  1--14, 2018{\natexlab{b}}.

\bibitem[Peng et~al.(2019)Peng, Chang, Zhang, Abbeel, and
  Levine]{Peng2019MCPLC}
Xue~Bin Peng, Michael Chang, Grace Zhang, P.~Abbeel, and Sergey Levine.
\newblock Mcp: Learning composable hierarchical control with multiplicative
  compositional policies.
\newblock In \emph{NeurIPS}, 2019.

\bibitem[Peng et~al.(2020)Peng, Coumans, Zhang, Lee, Tan, and
  Levine]{peng2020learning}
Xue~Bin Peng, Erwin Coumans, Tingnan Zhang, Tsang-Wei Lee, Jie Tan, and Sergey
  Levine.
\newblock Learning agile robotic locomotion skills by imitating animals, 2020.

\bibitem[Peng et~al.(2021)Peng, Ma, Abbeel, Levine, and
  Kanazawa]{Peng2021AMPAM}
Xue~Bin Peng, Ze~Ma, P.~Abbeel, Sergey Levine, and Angjoo Kanazawa.
\newblock Amp: Adversarial motion priors for stylized physics-based character
  control.
\newblock \emph{ACM Trans. Graph.}, 40:\penalty0 144:1--144:20, 2021.

\bibitem[Pertsch et~al.(2020)Pertsch, Lee, and Lim]{Pertsch2020AcceleratingRL}
Karl Pertsch, Youngwoo Lee, and Joseph~J. Lim.
\newblock Accelerating reinforcement learning with learned skill priors.
\newblock \emph{ArXiv}, abs/2010.11944, 2020.

\bibitem[Pertsch et~al.(2021)Pertsch, Lee, Wu, and
  Lim]{Pertsch2021DemonstrationGuidedRL}
Karl Pertsch, Youngwoon Lee, Yue Wu, and Joseph~J. Lim.
\newblock Demonstration-guided reinforcement learning with learned skills.
\newblock In \emph{CoRL}, 2021.

\bibitem[Peters and Schaal(2008)]{peters2008natural}
Jan Peters and Stefan Schaal.
\newblock Natural actor-critic.
\newblock \emph{Neurocomputing}, 71\penalty0 (7-9):\penalty0 1180--1190, 2008.

\bibitem[Pitis(2019)]{pitis2019rethinking}
Silviu Pitis.
\newblock Rethinking the discount factor in reinforcement learning: A decision
  theoretic approach.
\newblock In \emph{Proceedings of the AAAI Conference on Artificial
  Intelligence}, volume~33, pages 7949--7956, 2019.

\bibitem[Pomerleau(1988)]{pomerleau1988alvinn}
Dean~A Pomerleau.
\newblock Alvinn: An autonomous land vehicle in a neural network.
\newblock \emph{Advances in neural information processing systems}, 1, 1988.

\bibitem[Pong et~al.(2018)Pong, Gu, Dalal, and Levine]{Pong2018TemporalDM}
Vitchyr~H. Pong, Shixiang~Shane Gu, Murtaza Dalal, and Sergey Levine.
\newblock Temporal difference models: Model-free deep rl for model-based
  control.
\newblock \emph{ArXiv}, abs/1802.09081, 2018.

\bibitem[Pong et~al.(2021)Pong, Nair, Smith, Huang, and
  Levine]{Pong2021OfflineML}
Vitchyr~H. Pong, Ashvin Nair, Laura Smith, Catherine Huang, and Sergey Levine.
\newblock Offline meta-reinforcement learning with online self-supervision.
\newblock In \emph{International Conference on Machine Learning}, 2021.

\bibitem[Pong et~al.(2022)Pong, Nair, Smith, Huang, and
  Levine]{pong2022offline}
Vitchyr~H Pong, Ashvin~V Nair, Laura~M Smith, Catherine Huang, and Sergey
  Levine.
\newblock Offline meta-reinforcement learning with online self-supervision.
\newblock In \emph{International Conference on Machine Learning}, pages
  17811--17829. PMLR, 2022.

\bibitem[Popov et~al.(2017)Popov, Heess, Lillicrap, Hafner, Barth-Maron,
  Vecer{\'i}k, Lampe, Tassa, Erez, and Riedmiller]{Popov2017DataefficientDR}
Ivaylo Popov, Nicolas Manfred~Otto Heess, Timothy~P. Lillicrap, Roland Hafner,
  Gabriel Barth-Maron, Matej Vecer{\'i}k, Thomas Lampe, Yuval Tassa, Tom Erez,
  and Martin~A. Riedmiller.
\newblock Data-efficient deep reinforcement learning for dexterous
  manipulation.
\newblock \emph{ArXiv}, abs/1704.03073, 2017.

\bibitem[Qi et~al.(2022)Qi, Abbeel, and Grover]{Qi2022ImitatingFA}
Carl Qi, P.~Abbeel, and Aditya Grover.
\newblock Imitating, fast and slow: Robust learning from demonstrations via
  decision-time planning.
\newblock \emph{ArXiv}, abs/2204.03597, 2022.

\bibitem[Radford et~al.(2019)Radford, Wu, Child, Luan, Amodei, Sutskever,
  et~al.]{Radford2019LanguageMA}
Alec Radford, Jeffrey Wu, Rewon Child, David Luan, Dario Amodei, Ilya
  Sutskever, et~al.
\newblock Language models are unsupervised multitask learners.
\newblock \emph{OpenAI blog}, 1\penalty0 (8):\penalty0 9, 2019.

\bibitem[Radford et~al.(2021)Radford, Kim, Hallacy, Ramesh, Goh, Agarwal,
  Sastry, Askell, Mishkin, Clark, et~al.]{radford2021learning}
Alec Radford, Jong~Wook Kim, Chris Hallacy, Aditya Ramesh, Gabriel Goh,
  Sandhini Agarwal, Girish Sastry, Amanda Askell, Pamela Mishkin, Jack Clark,
  et~al.
\newblock Learning transferable visual models from natural language
  supervision.
\newblock In \emph{International conference on machine learning}, pages
  8748--8763. PMLR, 2021.

\bibitem[Radosavovic et~al.(2023)Radosavovic, Xiao, James, Abbeel, Malik, and
  Darrell]{pmlr-v205-radosavovic23a}
Ilija Radosavovic, Tete Xiao, Stephen James, Pieter Abbeel, Jitendra Malik, and
  Trevor Darrell.
\newblock Real-world robot learning with masked visual pre-training.
\newblock In Karen Liu, Dana Kulic, and Jeff Ichnowski, editors,
  \emph{Proceedings of The 6th Conference on Robot Learning}, volume 205 of
  \emph{Proceedings of Machine Learning Research}, pages 416--426. PMLR, 14--18
  Dec 2023.
\newblock URL \url{https://proceedings.mlr.press/v205/radosavovic23a.html}.

\bibitem[Raffel(2021)]{raffelcall}
Colin Raffel.
\newblock A call to build models like we build opensource software.
\newblock \emph{blog}, 2021.
\newblock URL
  \url{https://colinraffel.com/blog/a-call-to-build-models-like-we-build-open-source-software.html}.

\bibitem[Raffin(2020)]{rl-zoo3}
Antonin Raffin.
\newblock Rl baselines3 zoo.
\newblock \url{https://github.com/DLR-RM/rl-baselines3-zoo}, 2020.

\bibitem[Rakelly et~al.(2019)Rakelly, Zhou, Finn, Levine, and
  Quillen]{rakelly2019efficient}
Kate Rakelly, Aurick Zhou, Chelsea Finn, Sergey Levine, and Deirdre Quillen.
\newblock Efficient off-policy meta-reinforcement learning via probabilistic
  context variables.
\newblock In \emph{International conference on machine learning}, pages
  5331--5340. PMLR, 2019.

\bibitem[Rana et~al.(2021)Rana, Dasagi, Haviland, Talbot, Milford, and
  Sunderhauf]{Rana2021BayesianCF}
Krishan Rana, Vibhavari Dasagi, Jesse Haviland, Ben Talbot, Michael Milford,
  and N.~Sunderhauf.
\newblock Bayesian controller fusion: Leveraging control priors in deep
  reinforcement learning for robotics.
\newblock \emph{ArXiv}, abs/2107.09822, 2021.

\bibitem[Rao et~al.(2021)Rao, Sadeghi, Hasenclever, Wulfmeier, Zambelli,
  Vezzani, Tirumala, Aytar, Merel, Heess, and Hadsell]{Rao2021LearningTM}
Dushyant Rao, Fereshteh Sadeghi, Leonard Hasenclever, Markus Wulfmeier, Martina
  Zambelli, Giulia Vezzani, Dhruva Tirumala, Yusuf Aytar, Josh Merel, Nicolas
  Manfred~Otto Heess, and Raia Hadsell.
\newblock Learning transferable motor skills with hierarchical latent mixture
  policies.
\newblock \emph{ArXiv}, abs/2112.05062, 2021.

\bibitem[Ratliff et~al.(2006)Ratliff, Bagnell, and
  Zinkevich]{Ratliff2006MaximumMP}
Nathan~D. Ratliff, J.~Andrew Bagnell, and Martin~A. Zinkevich.
\newblock Maximum margin planning.
\newblock \emph{Proceedings of the 23rd international conference on Machine
  learning}, 2006.

\bibitem[Ravichandar et~al.(2020)Ravichandar, Polydoros, Chernova, and
  Billard]{ravichandar2020recent}
Harish Ravichandar, Athanasios~S Polydoros, Sonia Chernova, and Aude Billard.
\newblock Recent advances in robot learning from demonstration.
\newblock \emph{Annual review of control, robotics, and autonomous systems},
  3:\penalty0 297--330, 2020.

\bibitem[Reed et~al.(2022)Reed, Zolna, Parisotto, Colmenarejo, Novikov,
  Barth-Maron, Gimenez, Sulsky, Kay, Springenberg, et~al.]{reed2022generalist}
Scott Reed, Konrad Zolna, Emilio Parisotto, Sergio~Gomez Colmenarejo, Alexander
  Novikov, Gabriel Barth-Maron, Mai Gimenez, Yury Sulsky, Jackie Kay,
  Jost~Tobias Springenberg, et~al.
\newblock A generalist agent.
\newblock \emph{arXiv preprint arXiv:2205.06175}, 2022.

\bibitem[Reid et~al.(2022)Reid, Yamada, and Gu]{Reid2022CanWH}
Machel Reid, Yutaro Yamada, and Shixiang~Shane Gu.
\newblock Can wikipedia help offline reinforcement learning?
\newblock \emph{ArXiv}, abs/2201.12122, 2022.

\bibitem[Richard et~al.(2021)Richard, ARAVECCHIA, Geist, and
  Pradalier]{antoine2021mbrl}
Antoine Richard, Stephanie ARAVECCHIA, Matthieu Geist, and Cédric Pradalier.
\newblock Learning behaviors through physics-driven latent imagination.
\newblock \emph{CoRL}, 2021.

\bibitem[Riedmiller et~al.(2018)Riedmiller, Hafner, Lampe, Neunert, Degrave,
  van~de Wiele, Mnih, Heess, and Springenberg]{riedmiller2018learning}
Martin Riedmiller, Roland Hafner, Thomas Lampe, Michael Neunert, Jonas Degrave,
  Tom van~de Wiele, Vlad Mnih, Nicolas Heess, and Jost~Tobias Springenberg.
\newblock Learning by playing - solving sparse reward tasks from scratch.
\newblock In \emph{Proceedings of the 35th International Conference on Machine
  Learning}, 2018.

\bibitem[Ring et~al.(1994)]{ring1994continual}
Mark~Bishop Ring et~al.
\newblock \emph{Continual learning in reinforcement environments}.
\newblock Citeseer, 1994.

\bibitem[Ross et~al.(2011)Ross, Gordon, and Bagnell]{ross2011reduction}
St{\'e}phane Ross, Geoffrey Gordon, and Drew Bagnell.
\newblock A reduction of imitation learning and structured prediction to
  no-regret online learning.
\newblock In \emph{Proceedings of the fourteenth international conference on
  artificial intelligence and statistics}, pages 627--635. JMLR Workshop and
  Conference Proceedings, 2011.

\bibitem[Russakovsky et~al.(2015)Russakovsky, Deng, Su, Krause, Satheesh, Ma,
  Huang, Karpathy, Khosla, Bernstein, Berg, and
  Fei-Fei]{Russakovsky2015ImageNetLS}
Olga Russakovsky, Jia Deng, Hao Su, Jonathan Krause, Sanjeev Satheesh, Sean Ma,
  Zhiheng Huang, Andrej Karpathy, Aditya Khosla, Michael~S. Bernstein,
  Alexander~C. Berg, and Li~Fei-Fei.
\newblock Imagenet large scale visual recognition challenge.
\newblock \emph{International Journal of Computer Vision}, 115:\penalty0
  211--252, 2015.

\bibitem[Rusu et~al.(2015)Rusu, Colmenarejo, Gulcehre, Desjardins, Kirkpatrick,
  Pascanu, Mnih, Kavukcuoglu, and Hadsell]{rusu2015policy}
Andrei~A Rusu, Sergio~Gomez Colmenarejo, Caglar Gulcehre, Guillaume Desjardins,
  James Kirkpatrick, Razvan Pascanu, Volodymyr Mnih, Koray Kavukcuoglu, and
  Raia Hadsell.
\newblock Policy distillation.
\newblock \emph{arXiv preprint arXiv:1511.06295}, 2015.

\bibitem[Rusu et~al.(2016)Rusu, Rabinowitz, Desjardins, Soyer, Kirkpatrick,
  Kavukcuoglu, Pascanu, and Hadsell]{rusu2016progressive}
Andrei~A Rusu, Neil~C Rabinowitz, Guillaume Desjardins, Hubert Soyer, James
  Kirkpatrick, Koray Kavukcuoglu, Razvan Pascanu, and Raia Hadsell.
\newblock Progressive neural networks.
\newblock \emph{arXiv preprint arXiv:1606.04671}, 2016.

\bibitem[Sadeghi and Levine(2016)]{sadeghi2016cad2rl}
Fereshteh Sadeghi and Sergey Levine.
\newblock Cad2rl: Real single-image flight without a single real image.
\newblock \emph{arXiv preprint arXiv:1611.04201}, 2016.

\bibitem[S{\ae}mundsson et~al.(2018)S{\ae}mundsson, Hofmann, and
  Deisenroth]{saemundsson2018meta}
Steind{\'o}r S{\ae}mundsson, Katja Hofmann, and Marc~Peter Deisenroth.
\newblock Meta reinforcement learning with latent variable gaussian processes.
\newblock \emph{arXiv preprint arXiv:1803.07551}, 2018.

\bibitem[Sala and Gobet(2017)]{SALA201755}
Giovanni Sala and Fernand Gobet.
\newblock When the music's over. does music skill transfer to children's and
  young adolescents' cognitive and academic skills? a meta-analysis.
\newblock \emph{Educational Research Review}, 20:\penalty0 55--67, 2017.
\newblock ISSN 1747-938X.
\newblock \doi{https://doi.org/10.1016/j.edurev.2016.11.005}.
\newblock URL
  \url{https://www.sciencedirect.com/science/article/pii/S1747938X16300641}.

\bibitem[Salter et~al.(2019)Salter, Rao, Wulfmeier, Hadsell, and
  Posner]{salter2019attention}
Sasha Salter, Dushyant Rao, Markus Wulfmeier, Raia Hadsell, and Ingmar Posner.
\newblock Attention-privileged reinforcement learning.
\newblock \emph{arXiv preprint arXiv:1911.08363}, 2019.

\bibitem[Salter et~al.(2022)Salter, Wulfmeier, Tirumala, Heess, Riedmiller,
  Hadsell, and Rao]{Salter2022MO2MO}
Sasha Salter, Markus Wulfmeier, Dhruva Tirumala, Nicolas Manfred~Otto Heess,
  Martin~A. Riedmiller, Raia Hadsell, and Dushyant Rao.
\newblock Mo2: Model-based offline options.
\newblock In \emph{CoLLAs}, 2022.
\newblock URL \url{https://api.semanticscholar.org/CorpusID:252089192}.

\bibitem[Sanchez-Gonzalez et~al.(2018)Sanchez-Gonzalez, Heess, Springenberg,
  Merel, Riedmiller, Hadsell, and Battaglia]{sanchez2018graph}
Alvaro Sanchez-Gonzalez, Nicolas Heess, Jost~Tobias Springenberg, Josh Merel,
  Martin Riedmiller, Raia Hadsell, and Peter Battaglia.
\newblock Graph networks as learnable physics engines for inference and
  control.
\newblock In \emph{International Conference on Machine Learning}, pages
  4470--4479. PMLR, 2018.

\bibitem[Scao et~al.(2022)Scao, Fan, Akiki, Pavlick, Ili'c, Hesslow, Castagn'e,
  and et~al]{Scao2022BLOOMA1}
Teven~Le Scao, Angela Fan, Christopher Akiki, Elizabeth-Jane Pavlick, Suzana
  Ili'c, Daniel Hesslow, Roman Castagn'e, and et~al.
\newblock Bloom: A 176b-parameter open-access multilingual language model.
\newblock \emph{ArXiv}, abs/2211.05100, 2022.

\bibitem[Schaul et~al.(2015)Schaul, Horgan, Gregor, and
  Silver]{schaul2015universal}
Tom Schaul, Daniel Horgan, Karol Gregor, and David Silver.
\newblock Universal value function approximators.
\newblock In \emph{International conference on machine learning}, pages
  1312--1320. PMLR, 2015.

\bibitem[Schaul et~al.(2019)Schaul, Borsa, Ding, Szepesvari, Ostrovski, Dabney,
  and Osindero]{schaul2019adapting}
Tom Schaul, Diana Borsa, David Ding, David Szepesvari, Georg Ostrovski, Will
  Dabney, and Simon Osindero.
\newblock Adapting behaviour for learning progress.
\newblock \emph{arXiv preprint arXiv:1912.06910}, 2019.

\bibitem[Schmeckpeper et~al.(2020)Schmeckpeper, Xie, Rybkin, Tian, Daniilidis,
  Levine, and Finn]{schmeckpeper2020learning}
Karl Schmeckpeper, Annie Xie, Oleh Rybkin, Stephen Tian, Kostas Daniilidis,
  Sergey Levine, and Chelsea Finn.
\newblock Learning predictive models from observation and interaction.
\newblock In \emph{Computer Vision--ECCV 2020: 16th European Conference,
  Glasgow, UK, August 23--28, 2020, Proceedings, Part XX 16}, pages 708--725.
  Springer, 2020.

\bibitem[Schmidhuber(1994)]{Schmidhuber1994OnLH}
Juergen Schmidhuber.
\newblock On learning how to learn learning strategies, 1994.
\newblock URL \url{https://api.semanticscholar.org/CorpusID:59806870}.

\bibitem[Schmidhuber(1987)]{schmidhuber1987srl}
Jurgen Schmidhuber.
\newblock Evolutionary principles in self-referential learning. on learning now
  to learn: The meta-meta-meta...-hook.
\newblock Diploma thesis, Technische Universitat Munchen, Germany, 14 May 1987.
\newblock URL \url{http://www.idsia.ch/~juergen/diploma.html}.

\bibitem[Schmitt et~al.(2018)Schmitt, Hudson, Z{\'i}dek, Osindero, Doersch,
  Czarnecki, Leibo, K{\"u}ttler, Zisserman, Simonyan, and
  Eslami]{Schmitt2018KickstartingDR}
Simon Schmitt, Jonathan~J. Hudson, Augustin Z{\'i}dek, Simon Osindero, Carl
  Doersch, Wojciech~M. Czarnecki, Joel~Z. Leibo, Heinrich K{\"u}ttler, Andrew
  Zisserman, Karen Simonyan, and S.~M.~Ali Eslami.
\newblock Kickstarting deep reinforcement learning.
\newblock \emph{ArXiv}, abs/1803.03835, 2018.

\bibitem[Sch{\"{o}}lkopf(2019)]{schoel2019}
Bernhard Sch{\"{o}}lkopf.
\newblock Causality for machine learning.
\newblock \emph{CoRR}, abs/1911.10500, 2019.
\newblock URL \url{http://arxiv.org/abs/1911.10500}.

\bibitem[Schubert et~al.(2023)Schubert, Zhang, Bruce, Bechtle, Parisotto,
  Riedmiller, Springenberg, Byravan, Hasenclever, and Heess]{Schubert2023AGD}
Ingmar Schubert, Jingwei Zhang, Jake Bruce, Sarah Bechtle, Emilio Parisotto,
  Martin Riedmiller, Jost~Tobias Springenberg, Arunkumar Byravan, Leonard
  Hasenclever, and Nicolas Heess.
\newblock A generalist dynamics model for control.
\newblock \emph{arXiv preprint arXiv:2305.10912}, 2023.

\bibitem[Schwarzer et~al.(2021)Schwarzer, Rajkumar, Noukhovitch, Anand,
  Charlin, Hjelm, Bachman, and Courville]{Schwarzer2021PretrainingRF}
Max Schwarzer, Nitarshan Rajkumar, Michael Noukhovitch, Ankesh Anand, Laurent
  Charlin, Devon Hjelm, Philip Bachman, and Aaron~C. Courville.
\newblock Pretraining representations for data-efficient reinforcement
  learning.
\newblock \emph{ArXiv}, abs/2106.04799, 2021.

\bibitem[Sekar et~al.(2020)Sekar, Rybkin, Daniilidis, Abbeel, Hafner, and
  Pathak]{sekar2020planning}
Ramanan Sekar, Oleh Rybkin, Kostas Daniilidis, Pieter Abbeel, Danijar Hafner,
  and Deepak Pathak.
\newblock Planning to explore via self-supervised world models.
\newblock In \emph{International Conference on Machine Learning}, pages
  8583--8592. PMLR, 2020.

\bibitem[Selfridge et~al.(1985)Selfridge, Sutton, and Barto]{selfridge1985}
Oliver~G. Selfridge, Richard~S. Sutton, and Andrew~G. Barto.
\newblock Training and tracking in robotics.
\newblock In \emph{Proceedings of the 9th International Joint Conference on
  Artificial Intelligence - Volume 1}, IJCAI'85, page 670–672, San Francisco,
  CA, USA, 1985. Morgan Kaufmann Publishers Inc.
\newblock ISBN 0934613028.

\bibitem[Seo et~al.(2020)Seo, Lee, Clavera, Kurutach, Shin, and
  Abbeel]{seo2020trajectory}
Younggyo Seo, Kimin Lee, Ignasi Clavera, Thanard Kurutach, Jinwoo Shin, and
  Pieter Abbeel.
\newblock Trajectory-wise multiple choice learning for dynamics generalization
  in reinforcement learning.
\newblock \emph{arXiv preprint arXiv:2010.13303}, 2020.

\bibitem[Seo et~al.(2022)Seo, Lee, James, and Abbeel]{seo2022reinforcement}
Younggyo Seo, Kimin Lee, Stephen~L James, and Pieter Abbeel.
\newblock Reinforcement learning with action-free pre-training from videos.
\newblock In \emph{International Conference on Machine Learning}, pages
  19561--19579. PMLR, 2022.

\bibitem[Shafiullah et~al.(2022)Shafiullah, Paxton, Pinto, Chintala, and
  Szlam]{shafiullah2022clip}
Nur Muhammad~Mahi Shafiullah, Chris Paxton, Lerrel Pinto, Soumith Chintala, and
  Arthur Szlam.
\newblock Clip-fields: Weakly supervised semantic fields for robotic memory.
\newblock \emph{arXiv preprint arXiv:2210.05663}, 2022.

\bibitem[Sharma et~al.(2021)Sharma, Gupta, Levine, Hausman, and
  Finn]{sharma2021persistent}
Archit Sharma, Abhishek Gupta, Sergey Levine, Karol Hausman, and Chelsea Finn.
\newblock Persistent reinforcement learning via subgoal curricula.
\newblock \emph{arXiv preprint arXiv:2107.12931}, 2021.

\bibitem[Sharma et~al.(2023)Sharma, Fantacci, Zhou, Koppula, Heess, Scholz, and
  Aytar]{sharma2023lossless}
Mohit Sharma, Claudio Fantacci, Yuxiang Zhou, Skanda Koppula, Nicolas Heess,
  Jon Scholz, and Yusuf Aytar.
\newblock Lossless adaptation of pretrained vision models for robotic
  manipulation.
\newblock \emph{arXiv preprint arXiv:2304.06600}, 2023.

\bibitem[Shen et~al.(2021)Shen, Xia, Li, Mart\'in-Mart\'in, Fan, Wang,
  P\'erez-D'Arpino, Buch, Srivastava, Tchapmi, Tchapmi, Vainio, Wong, Fei-Fei,
  and Savarese]{shen2021igibson}
Bokui Shen, Fei Xia, Chengshu Li, Roberto Mart\'in-Mart\'in, Linxi Fan, Guanzhi
  Wang, Claudia P\'erez-D'Arpino, Shyamal Buch, Sanjana Srivastava, Lyne~P.
  Tchapmi, Micael~E. Tchapmi, Kent Vainio, Josiah Wong, Li~Fei-Fei, and Silvio
  Savarese.
\newblock igibson 1.0: a simulation environment for interactive tasks in large
  realistic scenes.
\newblock In \emph{2021 IEEE/RSJ International Conference on Intelligent Robots
  and Systems (IROS)}, page accepted. IEEE, 2021.

\bibitem[Shen et~al.(2020)Shen, Xiong, Xia, and Soatto]{shen2020towards}
Yantao Shen, Yuanjun Xiong, Wei Xia, and Stefano Soatto.
\newblock Towards backward-compatible representation learning.
\newblock In \emph{Proceedings of the IEEE/CVF Conference on Computer Vision
  and Pattern Recognition}, pages 6368--6377, 2020.

\bibitem[Shiarlis et~al.(2018)Shiarlis, Wulfmeier, Salter, Whiteson, and
  Posner]{Shiarlis2018TACOLT}
Kyriacos Shiarlis, Markus Wulfmeier, Sasha Salter, Shimon Whiteson, and Ingmar
  Posner.
\newblock Taco: Learning task decomposition via temporal alignment for control.
\newblock In \emph{ICML}, 2018.

\bibitem[Shridhar et~al.(2022)Shridhar, Manuelli, and Fox]{shridhar2022cliport}
Mohit Shridhar, Lucas Manuelli, and Dieter Fox.
\newblock Cliport: What and where pathways for robotic manipulation.
\newblock In \emph{Conference on Robot Learning}, pages 894--906. PMLR, 2022.

\bibitem[Silver et~al.(2016)Silver, Huang, Maddison, Guez, Sifre, van~den
  Driessche, Schrittwieser, Antonoglou, Panneershelvam, Lanctot, Dieleman,
  Grewe, Nham, Kalchbrenner, Sutskever, Lillicrap, Leach, Kavukcuoglu, Graepel,
  and Hassabis]{SilverAlphaGo}
David Silver, Aja Huang, Chris~J. Maddison, Arthur Guez, Laurent Sifre, George
  van~den Driessche, Julian Schrittwieser, Ioannis Antonoglou, Vedavyas
  Panneershelvam, Marc Lanctot, Sander Dieleman, Dominik Grewe, John Nham, Nal
  Kalchbrenner, Ilya Sutskever, Timothy~P. Lillicrap, Madeleine Leach, Koray
  Kavukcuoglu, Thore Graepel, and Demis Hassabis.
\newblock Mastering the game of go with deep neural networks and tree search.
\newblock \emph{Nature}, 529\penalty0 (7587):\penalty0 484--489, 2016.

\bibitem[Silver et~al.(2018)Silver, Allen, Tenenbaum, and
  Kaelbling]{Silver2018ResidualPL}
Tom Silver, Kelsey~R. Allen, Joshua~B. Tenenbaum, and Leslie~Pack Kaelbling.
\newblock Residual policy learning.
\newblock \emph{ArXiv}, abs/1812.06298, 2018.

\bibitem[Singh et~al.(2019)Singh, Yang, Hartikainen, Finn, and
  Levine]{singh2019end}
Avi Singh, Larry Yang, Kristian Hartikainen, Chelsea Finn, and Sergey Levine.
\newblock End-to-end robotic reinforcement learning without reward engineering.
\newblock \emph{arXiv preprint arXiv:1904.07854}, 2019.

\bibitem[Singh et~al.(2020{\natexlab{a}})Singh, Liu, Zhou, Yu, Rhinehart, and
  Levine]{singh2020parrot}
Avi Singh, Huihan Liu, Gaoyue Zhou, Albert Yu, Nicholas Rhinehart, and Sergey
  Levine.
\newblock Parrot: Data-driven behavioral priors for reinforcement learning.
\newblock \emph{arXiv preprint arXiv:2011.10024}, 2020{\natexlab{a}}.

\bibitem[Singh et~al.(2020{\natexlab{b}})Singh, Yu, Yang, Zhang, Kumar, and
  Levine]{singh2020cog}
Avi Singh, Albert Yu, Jonathan Yang, Jesse Zhang, Aviral Kumar, and Sergey
  Levine.
\newblock Cog: Connecting new skills to past experience with offline
  reinforcement learning.
\newblock \emph{arXiv preprint arXiv:2010.14500}, 2020{\natexlab{b}}.

\bibitem[Singh et~al.(2023)Singh, Blukis, Mousavian, Goyal, Xu, Tremblay, Fox,
  Thomason, and Garg]{singh2023progprompt}
Ishika Singh, Valts Blukis, Arsalan Mousavian, Ankit Goyal, Danfei Xu, Jonathan
  Tremblay, Dieter Fox, Jesse Thomason, and Animesh Garg.
\newblock Progprompt: Generating situated robot task plans using large language
  models.
\newblock In \emph{2023 IEEE International Conference on Robotics and
  Automation (ICRA)}, pages 11523--11530. IEEE, 2023.

\bibitem[Singh et~al.(2009)Singh, Lewis, and Barto]{singh2009rewards}
Satinder Singh, Richard~L Lewis, and Andrew~G Barto.
\newblock Where do rewards come from.
\newblock In \emph{Proceedings of the annual conference of the cognitive
  science society}, pages 2601--2606. Cognitive Science Society, 2009.

\bibitem[Singh(1992)]{singh1992transfer}
Satinder~Pal Singh.
\newblock Transfer of learning by composing solutions of elemental sequential
  tasks.
\newblock \emph{Machine Learning}, 8\penalty0 (3):\penalty0 323--339, 1992.

\bibitem[Sivakumar et~al.(2022)Sivakumar, Shaw, and
  Pathak]{sivakumar2022robotic}
Aravind Sivakumar, Kenneth Shaw, and Deepak Pathak.
\newblock Robotic telekinesis: Learning a robotic hand imitator by watching
  humans on youtube.
\newblock \emph{arXiv preprint arXiv:2202.10448}, 2022.

\bibitem[Skinner(1953)]{skinner1953some}
Burrhus~F Skinner.
\newblock Some contributions of an experimental analysis of behavior to
  psychology as a whole.
\newblock \emph{American Psychologist}, 8\penalty0 (2):\penalty0 69, 1953.

\bibitem[Smith et~al.(2019)Smith, Dhawan, Zhang, Abbeel, and
  Levine]{smith2019avid}
Laura Smith, Nikita Dhawan, Marvin Zhang, Pieter Abbeel, and Sergey Levine.
\newblock Avid: Learning multi-stage tasks via pixel-level translation of human
  videos.
\newblock \emph{arXiv preprint arXiv:1912.04443}, 2019.

\bibitem[Smith et~al.(2023)Smith, Kew, Li, Luu, Peng, Ha, Tan, and
  Levine]{Smith2023LearningAA}
Laura Smith, J.~Chase Kew, Tianyu Li, Linda Luu, Xue~Bin Peng, Sehoon Ha, Jie
  Tan, and Sergey Levine.
\newblock Learning and adapting agile locomotion skills by transferring
  experience.
\newblock \emph{ArXiv}, abs/2304.09834, 2023.

\bibitem[Song et~al.(2022)Song, Zhou, Sekhari, Bagnell, Krishnamurthy, and
  Sun]{song2022hybrid}
Yuda Song, Yifei Zhou, Ayush Sekhari, J~Andrew Bagnell, Akshay Krishnamurthy,
  and Wen Sun.
\newblock Hybrid rl: Using both offline and online data can make rl efficient.
\newblock \emph{arXiv preprint arXiv:2210.06718}, 2022.

\bibitem[Spelke et~al.(1992)Spelke, Breinlinger, Macomber, and
  Jacobson]{spelke1992origins}
Elizabeth~S Spelke, Karen Breinlinger, Janet Macomber, and Kristen Jacobson.
\newblock Origins of knowledge.
\newblock \emph{Psychological review}, 99\penalty0 (4):\penalty0 605, 1992.

\bibitem[Srivastava et~al.(2022)Srivastava, Rastogi, Rao, Shoeb, Abid, Fisch,
  and et~al]{Srivastava2022BeyondTI}
Aarohi Srivastava, Abhinav Rastogi, Abhishek~B Rao, Abu Awal~Md Shoeb, Abubakar
  Abid, Adam Fisch, and et~al.
\newblock Beyond the imitation game: Quantifying and extrapolating the
  capabilities of language models.
\newblock \emph{ArXiv}, abs/2206.04615, 2022.

\bibitem[Stadie et~al.(2017)Stadie, Abbeel, and
  Sutskever]{Stadie2017ThirdPersonIL}
Bradly~C. Stadie, P.~Abbeel, and Ilya Sutskever.
\newblock Third-person imitation learning.
\newblock \emph{ArXiv}, abs/1703.01703, 2017.

\bibitem[Stoica et~al.(2023)Stoica, Bolya, Bjorner, Hearn, and
  Hoffman]{Stoica2023ZipItMM}
George Stoica, Daniel Bolya, Jakob~Bue Bjorner, Taylor~N. Hearn, and Judy
  Hoffman.
\newblock Zipit! merging models from different tasks without training.
\newblock \emph{ArXiv}, abs/2305.03053, 2023.

\bibitem[Stone et~al.(2021)Stone, Ramirez, Konolige, and
  Jonschkowski]{stone2021distracting}
Austin Stone, Oscar Ramirez, Kurt Konolige, and Rico Jonschkowski.
\newblock The distracting control suite -- a challenging benchmark for
  reinforcement learning from pixels.
\newblock \emph{arXiv preprint arXiv:2101.02722}, 2021.

\bibitem[Sun et~al.(2020)Sun, Kretzschmar, Dotiwalla, Chouard, Patnaik, Tsui,
  Guo, Zhou, Chai, Caine, Vasudevan, Han, Ngiam, Zhao, Timofeev, Ettinger,
  Krivokon, Gao, Joshi, Zhang, Shlens, Chen, and Anguelov]{Sun_2020_CVPR}
Pei Sun, Henrik Kretzschmar, Xerxes Dotiwalla, Aurelien Chouard, Vijaysai
  Patnaik, Paul Tsui, James Guo, Yin Zhou, Yuning Chai, Benjamin Caine, Vijay
  Vasudevan, Wei Han, Jiquan Ngiam, Hang Zhao, Aleksei Timofeev, Scott
  Ettinger, Maxim Krivokon, Amy Gao, Aditya Joshi, Yu~Zhang, Jonathon Shlens,
  Zhifeng Chen, and Dragomir Anguelov.
\newblock Scalability in perception for autonomous driving: Waymo open dataset.
\newblock In \emph{Proceedings of the IEEE/CVF Conference on Computer Vision
  and Pattern Recognition (CVPR)}, June 2020.

\bibitem[Sung et~al.(2017)Sung, Zhang, Xiang, Hospedales, and
  Yang]{sung2017learning}
Flood Sung, Li~Zhang, Tao Xiang, Timothy Hospedales, and Yongxin Yang.
\newblock Learning to learn: Meta-critic networks for sample efficient
  learning.
\newblock \emph{arXiv preprint arXiv:1706.09529}, 2017.

\bibitem[Sutton(2019)]{bitter}
Rich Sutton.
\newblock The bitter lesson.
\newblock \emph{Incomplete Ideas (blog)}, 2019.
\newblock URL \url{http://www.incompleteideas.net/IncIdeas/BitterLesson.html}.

\bibitem[Sutton(1991)]{sutton1991dyna}
Richard~S Sutton.
\newblock Dyna, an integrated architecture for learning, planning, and
  reacting.
\newblock \emph{ACM Sigart Bulletin}, 2\penalty0 (4):\penalty0 160--163, 1991.

\bibitem[Sutton and Barto(1990)]{sutton1990time}
Richard~S Sutton and Andrew~G Barto.
\newblock \emph{Time-derivative models of pavlovian reinforcement.}
\newblock The MIT Press, 1990.

\bibitem[Sutton and Barto(2018)]{sutton2018reinforcement}
Richard~S Sutton and Andrew~G Barto.
\newblock \emph{Reinforcement learning: An introduction}.
\newblock MIT press, 2018.

\bibitem[Sutton et~al.(1999)Sutton, Precup, and Singh]{sutton1999between}
Richard~S. Sutton, Doina Precup, and Satinder Singh.
\newblock Between {MDP}s and semi-{MDP}s: A framework for temporal abstraction
  in reinforcement learning.
\newblock \emph{Artif. Intell.}, 112\penalty0 (1–2):\penalty0 181–211,
  August 1999.
\newblock ISSN 0004-3702.
\newblock \doi{10.1016/S0004-3702(99)00052-1}.
\newblock URL \url{https://doi.org/10.1016/S0004-3702(99)00052-1}.

\bibitem[Sutton et~al.(2011)Sutton, Modayil, Delp, Degris, Pilarski, White, and
  Precup]{sutton2011horde}
Richard~S. Sutton, Joseph Modayil, Michael Delp, Thomas Degris, Patrick~M.
  Pilarski, Adam White, and Doina Precup.
\newblock Horde: A scalable real-time architecture for learning knowledge from
  unsupervised sensorimotor interaction.
\newblock In \emph{The 10th International Conference on Autonomous Agents and
  Multiagent Systems - Volume 2}, AAMAS '11, page 761–768. International
  Foundation for Autonomous Agents and Multiagent Systems, 2011.
\newblock ISBN 0982657161.

\bibitem[Szot et~al.(2021)Szot, Clegg, Undersander, Wijmans, Zhao, Turner,
  Maestre, Mukadam, Chaplot, Maksymets, Gokaslan, Vondrus, Dharur, Meier,
  Galuba, Chang, Kira, Koltun, Malik, Savva, and Batra]{szot2021habitat}
Andrew Szot, Alex Clegg, Eric Undersander, Erik Wijmans, Yili Zhao, John
  Turner, Noah Maestre, Mustafa Mukadam, Devendra Chaplot, Oleksandr Maksymets,
  Aaron Gokaslan, Vladimir Vondrus, Sameer Dharur, Franziska Meier, Wojciech
  Galuba, Angel Chang, Zsolt Kira, Vladlen Koltun, Jitendra Malik, Manolis
  Savva, and Dhruv Batra.
\newblock Habitat 2.0: Training home assistants to rearrange their habitat.
\newblock In \emph{Advances in Neural Information Processing Systems
  (NeurIPS)}, 2021.

\bibitem[Tan et~al.(2022)Tan, Senanayake, and Ramos]{tan2022renaissance}
Julia Tan, Ransalu Senanayake, and Fabio Ramos.
\newblock Renaissance robot: Optimal transport policy fusion for learning
  diverse skills.
\newblock In \emph{2022 IEEE/RSJ International Conference on Intelligent Robots
  and Systems (IROS)}, pages 7052--7059. IEEE, 2022.

\bibitem[Tao et~al.(2021)Tao, Genc, Chung, Sun, and Mallya]{tao2021repaint}
Yunzhe Tao, Sahika Genc, Jonathan Chung, Tao Sun, and Sunil Mallya.
\newblock Repaint: Knowledge transfer in deep reinforcement learning.
\newblock In \emph{International Conference on Machine Learning}, pages
  10141--10152. PMLR, 2021.

\bibitem[Tassa et~al.(2018)Tassa, Doron, Muldal, Erez, Li, de~Las~Casas,
  Budden, Abdolmaleki, Merel, Lefrancq, Lillicrap, and
  Riedmiller]{tassa2018control}
Yuval Tassa, Yotam Doron, Alistair Muldal, Tom Erez, Yazhe Li, Diego
  de~Las~Casas, David Budden, Abbas Abdolmaleki, Josh Merel, Andrew Lefrancq,
  Timothy Lillicrap, and Martin Riedmiller.
\newblock Deep{Mind} control suite.
\newblock Technical report, DeepMind, January 2018.
\newblock URL \url{https://arxiv.org/abs/1801.00690}.

\bibitem[Taylor and Stone(2007)]{taylor2007representation}
Matthew~E Taylor and Peter Stone.
\newblock Representation transfer for reinforcement learning.
\newblock In \emph{AAAI Fall Symposium: Computational Approaches to
  Representation Change during Learning and Development}, pages 78--85, 2007.

\bibitem[Taylor and Stone(2009)]{taylor2009transfer}
Matthew~E Taylor and Peter Stone.
\newblock Transfer learning for reinforcement learning domains: A survey.
\newblock \emph{Journal of Machine Learning Research}, 10\penalty0 (7), 2009.

\bibitem[Team et~al.(2021)Team, Abramson, Ahuja, Brussee, Carnevale, Cassin,
  Fischer, Georgiev, Goldin, Harley, et~al.]{team2021creating}
DeepMind Interactive~Agents Team, Josh Abramson, Arun Ahuja, Arthur Brussee,
  Federico Carnevale, Mary Cassin, Felix Fischer, Petko Georgiev, Alex Goldin,
  Tim Harley, et~al.
\newblock Creating multimodal interactive agents with imitation and
  self-supervised learning.
\newblock \emph{arXiv preprint arXiv:2112.03763}, 2021.

\bibitem[Team(2023)]{MosaicML2023Introducing}
MosaicML~NLP Team.
\newblock Introducing mpt-7b: A new standard for open-source, commercially
  usable llms, 2023.
\newblock URL \url{www.mosaicml.com/blog/mpt-7b}.
\newblock Accessed: 2023-05-05.

\bibitem[Teh et~al.(2017)Teh, Bapst, Czarnecki, Quan, Kirkpatrick, Hadsell,
  Heess, and Pascanu]{teh2017distral}
Yee~Whye Teh, Victor Bapst, Wojciech~Marian Czarnecki, John Quan, James
  Kirkpatrick, Raia Hadsell, Nicolas Heess, and Razvan Pascanu.
\newblock Distral: Robust multitask reinforcement learning.
\newblock \emph{arXiv preprint arXiv:1707.04175}, 2017.

\bibitem[Thrun(1998)]{thrun1998lifelong}
Sebastian Thrun.
\newblock Lifelong learning algorithms.
\newblock In \emph{Learning to learn}, pages 181--209. Springer, 1998.

\bibitem[Tirinzoni et~al.(2018{\natexlab{a}})Tirinzoni, Rodriguez~Sanchez, and
  Restelli]{tirinzoni2018transfer}
Andrea Tirinzoni, Rafael Rodriguez~Sanchez, and Marcello Restelli.
\newblock Transfer of value functions via variational methods.
\newblock \emph{Advances in Neural Information Processing Systems},
  31:\penalty0 6179--6189, 2018{\natexlab{a}}.

\bibitem[Tirinzoni et~al.(2018{\natexlab{b}})Tirinzoni, Sessa, Pirotta, and
  Restelli]{Tirinzoni2018ImportanceWT}
Andrea Tirinzoni, Andrea Sessa, Matteo Pirotta, and Marcello Restelli.
\newblock Importance weighted transfer of samples in reinforcement learning.
\newblock \emph{ArXiv}, abs/1805.10886, 2018{\natexlab{b}}.

\bibitem[Tirumala et~al.(2020)Tirumala, Galashov, Noh, Hasenclever, Pascanu,
  Schwarz, Desjardins, Czarnecki, Ahuja, Teh, et~al.]{tirumala2020behavior}
Dhruva Tirumala, Alexandre Galashov, Hyeonwoo Noh, Leonard Hasenclever, Razvan
  Pascanu, Jonathan Schwarz, Guillaume Desjardins, Wojciech~Marian Czarnecki,
  Arun Ahuja, Yee~Whye Teh, et~al.
\newblock Behavior priors for efficient reinforcement learning.
\newblock \emph{arXiv preprint arXiv:2010.14274}, 2020.

\bibitem[Tobin et~al.(2017)Tobin, Fong, Ray, Schneider, Zaremba, and
  Abbeel]{tobin2017domain}
Josh Tobin, Rachel Fong, Alex Ray, Jonas Schneider, Wojciech Zaremba, and
  Pieter Abbeel.
\newblock Domain randomization for transferring deep neural networks from
  simulation to the real world.
\newblock In \emph{2017 IEEE/RSJ international conference on intelligent robots
  and systems (IROS)}, pages 23--30. IEEE, 2017.

\bibitem[Torabi et~al.(2019)Torabi, Warnell, and Stone]{Torabi2019RecentAI}
Faraz Torabi, Garrett Warnell, and Peter Stone.
\newblock Recent advances in imitation learning from observation.
\newblock In \emph{IJCAI}, 2019.

\bibitem[Torrey et~al.(2007)Torrey, Shavlik, Walker, and
  Maclin]{torrey2007relational}
Lisa Torrey, Jude Shavlik, Trevor Walker, and Richard Maclin.
\newblock Relational macros for transfer in reinforcement learning.
\newblock In \emph{International Conference on Inductive Logic Programming},
  pages 254--268. Springer, 2007.

\bibitem[Touvron et~al.(2023)Touvron, Martin, Stone, Albert, Almahairi, Babaei,
  Bashlykov, Batra, Bhargava, Bhosale, Bikel, Blecher, Ferrer, Chen, Cucurull,
  Esiobu, Fernandes, Fu, Fu, Fuller, Gao, Goswami, Goyal, Hartshorn, Hosseini,
  Hou, Inan, Kardas, Kerkez, Khabsa, Kloumann, Korenev, Koura, Lachaux, Lavril,
  Lee, Liskovich, Lu, Mao, Martinet, Mihaylov, Mishra, Molybog, Nie, Poulton,
  Reizenstein, Rungta, Saladi, Schelten, Silva, Smith, Subramanian, Tan, Tang,
  Taylor, Williams, Kuan, Xu, Yan, Zarov, Zhang, Fan, Kambadur, Narang,
  Rodriguez, Stojnic, Edunov, and Scialom]{Touvron2023Llama2O}
Hugo Touvron, Louis Martin, Kevin~R. Stone, Peter Albert, Amjad Almahairi,
  Yasmine Babaei, Nikolay Bashlykov, Soumya Batra, Prajjwal Bhargava, Shruti
  Bhosale, Daniel~M. Bikel, Lukas Blecher, Cristian~Canton Ferrer, Moya Chen,
  Guillem Cucurull, David Esiobu, Jude Fernandes, Jeremy Fu, Wenyin Fu, Brian
  Fuller, Cynthia Gao, Vedanuj Goswami, Naman Goyal, Anthony~S. Hartshorn,
  Saghar Hosseini, Rui Hou, Hakan Inan, Marcin Kardas, Viktor Kerkez, Madian
  Khabsa, Isabel~M. Kloumann, A.~V. Korenev, Punit~Singh Koura, Marie-Anne
  Lachaux, Thibaut Lavril, Jenya Lee, Diana Liskovich, Yinghai Lu, Yuning Mao,
  Xavier Martinet, Todor Mihaylov, Pushkar Mishra, Igor Molybog, Yixin Nie,
  Andrew Poulton, Jeremy Reizenstein, Rashi Rungta, Kalyan Saladi, Alan
  Schelten, Ruan Silva, Eric~Michael Smith, R.~Subramanian, Xia Tan, Binh Tang,
  Ross Taylor, Adina Williams, Jian~Xiang Kuan, Puxin Xu, Zhengxu Yan, Iliyan
  Zarov, Yuchen Zhang, Angela Fan, Melanie Kambadur, Sharan Narang, Aurelien
  Rodriguez, Robert Stojnic, Sergey Edunov, and Thomas Scialom.
\newblock Llama 2: Open foundation and fine-tuned chat models.
\newblock \emph{ArXiv}, abs/2307.09288, 2023.
\newblock URL \url{https://api.semanticscholar.org/CorpusID:259950998}.

\bibitem[Truong et~al.(2021)Truong, Yarats, Li, Meier, Chernova, Batra, and
  Rai]{truong2021learning}
Joanne Truong, Denis Yarats, Tianyu Li, Franziska Meier, Sonia Chernova, Dhruv
  Batra, and Akshara Rai.
\newblock Learning navigation skills for legged robots with learned robot
  embeddings.
\newblock In \emph{2021 IEEE/RSJ International Conference on Intelligent Robots
  and Systems (IROS)}, pages 484--491. IEEE, 2021.

\bibitem[Van~Niekerk et~al.(2019)Van~Niekerk, James, Earle, and
  Rosman]{van2019composing}
Benjamin Van~Niekerk, Steven James, Adam Earle, and Benjamin Rosman.
\newblock Composing value functions in reinforcement learning.
\newblock In \emph{International Conference on Machine Learning}, pages
  6401--6409. PMLR, 2019.

\bibitem[Vaswani et~al.(2017)Vaswani, Shazeer, Parmar, Uszkoreit, Jones, Gomez,
  Kaiser, and Polosukhin]{Vaswani2017AttentionIA}
Ashish Vaswani, Noam~M. Shazeer, Niki Parmar, Jakob Uszkoreit, Llion Jones,
  Aidan~N. Gomez, Lukasz Kaiser, and Illia Polosukhin.
\newblock Attention is all you need.
\newblock In \emph{NIPS}, 2017.

\bibitem[Vecerik et~al.(2017)Vecerik, Hester, Scholz, Wang, Pietquin, Piot,
  Heess, Roth{\"o}rl, Lampe, and Riedmiller]{vecerik2017leveraging}
Mel Vecerik, Todd Hester, Jonathan Scholz, Fumin Wang, Olivier Pietquin, Bilal
  Piot, Nicolas Heess, Thomas Roth{\"o}rl, Thomas Lampe, and Martin Riedmiller.
\newblock Leveraging demonstrations for deep reinforcement learning on robotics
  problems with sparse rewards.
\newblock \emph{arXiv preprint arXiv:1707.08817}, 2017.

\bibitem[Veerapaneni et~al.(2020)Veerapaneni, Co-Reyes, Chang, Janner, Finn,
  Wu, Tenenbaum, and Levine]{veerapaneni2020entity}
Rishi Veerapaneni, John~D Co-Reyes, Michael Chang, Michael Janner, Chelsea
  Finn, Jiajun Wu, Joshua Tenenbaum, and Sergey Levine.
\newblock Entity abstraction in visual model-based reinforcement learning.
\newblock In \emph{Conference on Robot Learning}, pages 1439--1456. PMLR, 2020.

\bibitem[Venkateswara et~al.(2017)Venkateswara, Eusebio, Chakraborty, and
  Panchanathan]{Venkateswara2017DeepHN}
Hemanth Venkateswara, Jose Eusebio, Shayok Chakraborty, and Sethuraman
  Panchanathan.
\newblock Deep hashing network for unsupervised domain adaptation.
\newblock \emph{2017 IEEE Conference on Computer Vision and Pattern Recognition
  (CVPR)}, pages 5385--5394, 2017.

\bibitem[Vezhnevets et~al.(2017)Vezhnevets, Osindero, Schaul, Heess, Jaderberg,
  Silver, and Kavukcuoglu]{vezhnevets2017feudal}
Alexander~Sasha Vezhnevets, Simon Osindero, Tom Schaul, Nicolas Heess, Max
  Jaderberg, David Silver, and Koray Kavukcuoglu.
\newblock Feudal networks for hierarchical reinforcement learning.
\newblock In \emph{International Conference on Machine Learning}, pages
  3540--3549. PMLR, 2017.

\bibitem[Vezzani et~al.(2022)Vezzani, Tirumala, Wulfmeier, Rao, Abdolmaleki,
  Moran, Haarnoja, Humplik, Hafner, Neunert, Fantacci, Hertweck, Lampe,
  Sadeghi, Heess, and Riedmiller]{Vezzani2022SkillSAS}
Giulia Vezzani, Dhruva Tirumala, Markus Wulfmeier, Dushyant Rao, Abbas
  Abdolmaleki, Ben Moran, Tuomas Haarnoja, Jan Humplik, Roland Hafner, Michael
  Neunert, Claudio Fantacci, Tim Hertweck, Thomas Lampe, Fereshteh Sadeghi,
  Nicolas Manfred~Otto Heess, and Martin~A. Riedmiller.
\newblock Skills: Adaptive skill sequencing for efficient temporally-extended
  exploration.
\newblock \emph{ArXiv}, abs/2211.13743, 2022.

\bibitem[Vinyals et~al.(2019)Vinyals, Babuschkin, Czarnecki, Mathieu, Dudzik,
  Chung, Choi, Powell, Ewalds, Georgiev, Oh, Horgan, Kroiss, Danihelka, Huang,
  Sifre, Cai, Agapiou, Jaderberg, Vezhnevets, Leblond, Pohlen, Dalibard,
  Budden, Sulsky, Molloy, Paine, G{\"{u}}l{\c{c}}ehre, Wang, Pfaff, Wu, Ring,
  Yogatama, W{\"{u}}nsch, McKinney, Smith, Schaul, Lillicrap, Kavukcuoglu,
  Hassabis, Apps, and Silver]{VinyalsStarCraft}
Oriol Vinyals, Igor Babuschkin, Wojciech~M. Czarnecki, Micha{\"{e}}l Mathieu,
  Andrew Dudzik, Junyoung Chung, David~H. Choi, Richard Powell, Timo Ewalds,
  Petko Georgiev, Junhyuk Oh, Dan Horgan, Manuel Kroiss, Ivo Danihelka, Aja
  Huang, Laurent Sifre, Trevor Cai, John~P. Agapiou, Max Jaderberg,
  Alexander~Sasha Vezhnevets, R{\'{e}}mi Leblond, Tobias Pohlen, Valentin
  Dalibard, David Budden, Yury Sulsky, James Molloy, Tom~L. Paine, {\c{C}}aglar
  G{\"{u}}l{\c{c}}ehre, Ziyu Wang, Tobias Pfaff, Yuhuai Wu, Roman Ring, Dani
  Yogatama, Dario W{\"{u}}nsch, Katrina McKinney, Oliver Smith, Tom Schaul,
  Timothy~P. Lillicrap, Koray Kavukcuoglu, Demis Hassabis, Chris Apps, and
  David Silver.
\newblock Grandmaster level in {StarCraft} {II} using multi-agent reinforcement
  learning.
\newblock \emph{Nature}, 575\penalty0 (7782):\penalty0 350--354, 2019.
\newblock \doi{10.1038/s41586-019-1724-z}.
\newblock URL \url{https://doi.org/10.1038/s41586-019-1724-z}.

\bibitem[Von~Oswald et~al.(2023)Von~Oswald, Niklasson, Randazzo, Sacramento,
  Mordvintsev, Zhmoginov, and Vladymyrov]{von2023transformers}
Johannes Von~Oswald, Eyvind Niklasson, Ettore Randazzo, Jo{\~a}o Sacramento,
  Alexander Mordvintsev, Andrey Zhmoginov, and Max Vladymyrov.
\newblock Transformers learn in-context by gradient descent.
\newblock In \emph{International Conference on Machine Learning}, pages
  35151--35174. PMLR, 2023.

\bibitem[Walke et~al.(2022)Walke, Yang, Yu, Kumar, Orbik, Singh, and
  Levine]{Walke2022DontSF}
Homer Walke, Jonathan Yang, Albert Yu, Aviral Kumar, Jedrzej Orbik, Avi Singh,
  and Sergey Levine.
\newblock Don't start from scratch: Leveraging prior data to automate robotic
  reinforcement learning.
\newblock In \emph{Conference on Robot Learning}, 2022.

\bibitem[Walker et~al.(2023)Walker, V'ertes, Li, Dulac-Arnold, Anand, Weber,
  and Hamrick]{Walker2023InvestigatingTR}
Jacob Walker, Eszter V'ertes, Yazhe Li, Gabriel Dulac-Arnold, Ankesh Anand,
  Th{\'e}ophane Weber, and Jessica~B. Hamrick.
\newblock Investigating the role of model-based learning in exploration and
  transfer.
\newblock In \emph{International Conference on Machine Learning}, 2023.
\newblock URL \url{https://api.semanticscholar.org/CorpusID:256662382}.

\bibitem[Wang et~al.(2018)Wang, Singh, Michael, Hill, Levy, and
  Bowman]{wang2019glue}
Alex Wang, Amanpreet Singh, Julian Michael, Felix Hill, Omer Levy, and Samuel~R
  Bowman.
\newblock Glue: A multi-task benchmark and analysis platform for natural
  language understanding.
\newblock \emph{arXiv preprint arXiv:1804.07461}, 2018.

\bibitem[Wang et~al.(2021)Wang, King, Porcel, Kurth-Nelson, Zhu, Deck, Choy,
  Cassin, Reynolds, Song, Buttimore, Reichert, Rabinowitz, Matthey, Hassabis,
  Lerchner, and Botvinick]{wang2021alchemy}
Jane Wang, Michael King, Nicolas Porcel, Zeb Kurth-Nelson, Tina Zhu, Charlie
  Deck, Peter Choy, Mary Cassin, Malcolm Reynolds, Francis Song, Gavin
  Buttimore, David Reichert, Neil Rabinowitz, Loic Matthey, Demis Hassabis,
  Alex Lerchner, and Matthew Botvinick.
\newblock Alchemy: A structured task distribution for meta-reinforcement
  learning.
\newblock \emph{arXiv preprint arXiv:2102.02926}, 2021.
\newblock URL \url{https://arxiv.org/abs/2102.02926}.

\bibitem[Wang et~al.(2016)Wang, Kurth-Nelson, Tirumala, Soyer, Leibo, Munos,
  Blundell, Kumaran, and Botvinick]{wang2016learning}
Jane~X Wang, Zeb Kurth-Nelson, Dhruva Tirumala, Hubert Soyer, Joel~Z Leibo,
  Remi Munos, Charles Blundell, Dharshan Kumaran, and Matt Botvinick.
\newblock Learning to reinforcement learn, 2016.

\bibitem[Wang and Van~Hoof(2022)]{wang2022model}
Qi~Wang and Herke Van~Hoof.
\newblock Model-based meta reinforcement learning using graph structured
  surrogate models and amortized policy search.
\newblock In \emph{International Conference on Machine Learning}, pages
  23055--23077. PMLR, 2022.

\bibitem[Wang and Ba(2019)]{wang2019exploring}
Tingwu Wang and Jimmy Ba.
\newblock Exploring model-based planning with policy networks.
\newblock \emph{arXiv preprint arXiv:1906.08649}, 2019.

\bibitem[Wang et~al.(2019)Wang, Bao, Clavera, Hoang, Wen, Langlois, Zhang,
  Zhang, Abbeel, and Ba]{wang2019benchmarking}
Tingwu Wang, Xuchan Bao, Ignasi Clavera, Jerrick Hoang, Yeming Wen, Eric
  Langlois, Shunshi Zhang, Guodong Zhang, Pieter Abbeel, and Jimmy Ba.
\newblock Benchmarking model-based reinforcement learning.
\newblock \emph{arXiv preprint arXiv:1907.02057}, 2019.

\bibitem[Wang et~al.(2020{\natexlab{a}})Wang, Liu, Chen, Ma, and
  Liu]{wang2020target}
Yue Wang, Yuting Liu, Wei Chen, Zhi-Ming Ma, and Tie-Yan Liu.
\newblock Target transfer q-learning and its convergence analysis.
\newblock \emph{Neurocomputing}, 392:\penalty0 11--22, 2020{\natexlab{a}}.

\bibitem[Wang et~al.(2017)Wang, Merel, Reed, Wayne, de~Freitas, and
  Heess]{wang2017robust}
Ziyu Wang, Josh Merel, Scott~E. Reed, Greg Wayne, Nando de~Freitas, and Nicolas
  Heess.
\newblock Robust imitation of diverse behaviors.
\newblock \emph{CoRR}, abs/1707.02747, 2017.

\bibitem[Wang et~al.(2020{\natexlab{b}})Wang, Novikov, Zolna, Springenberg,
  Reed, Shahriari, Siegel, Merel, Gulcehre, Heess, and
  de~Freitas]{wang2020critic}
Ziyu Wang, Alexander Novikov, Konrad Zolna, Jost~Tobias Springenberg, Scott
  Reed, Bobak Shahriari, Noah Siegel, Josh Merel, Caglar Gulcehre, Nicolas
  Heess, and Nando de~Freitas.
\newblock Critic regularized regression, 2020{\natexlab{b}}.

\bibitem[Watkins and Dayan(1992)]{watkins1992q}
Christopher~JCH Watkins and Peter Dayan.
\newblock Q-learning.
\newblock \emph{Machine learning}, 8\penalty0 (3):\penalty0 279--292, 1992.

\bibitem[Weinstein and Botvinick(2017)]{weinstein2017structure}
Ari Weinstein and Matthew~M Botvinick.
\newblock Structure learning in motor control: A deep reinforcement learning
  model.
\newblock \emph{arXiv preprint arXiv:1706.06827}, 2017.

\bibitem[White(2017)]{white2017unifying}
Martha White.
\newblock Unifying task specification in reinforcement learning.
\newblock In \emph{International Conference on Machine Learning}, pages
  3742--3750. PMLR, 2017.

\bibitem[Williams et~al.(2017)Williams, Wagener, Goldfain, Drews, Rehg, Boots,
  and Theodorou]{williams2017information}
Grady Williams, Nolan Wagener, Brian Goldfain, Paul Drews, James~M Rehg, Byron
  Boots, and Evangelos~A Theodorou.
\newblock Information theoretic mpc for model-based reinforcement learning.
\newblock In \emph{2017 IEEE International Conference on Robotics and
  Automation (ICRA)}, pages 1714--1721. IEEE, 2017.

\bibitem[Williams(1992)]{williams1992simple}
Ronald~J Williams.
\newblock Simple statistical gradient-following algorithms for connectionist
  reinforcement learning.
\newblock \emph{Machine learning}, 8\penalty0 (3):\penalty0 229--256, 1992.

\bibitem[Witten(1977)]{WITTEN1977286}
Ian~H. Witten.
\newblock An adaptive optimal controller for discrete-time markov environments.
\newblock \emph{Information and Control}, 34\penalty0 (4):\penalty0 286--295,
  1977.
\newblock ISSN 0019-9958.
\newblock \doi{https://doi.org/10.1016/S0019-9958(77)90354-0}.
\newblock URL
  \url{https://www.sciencedirect.com/science/article/pii/S0019995877903540}.

\bibitem[Wolpert et~al.(2001)Wolpert, Ghahramani, and Flanagan]{Wolpert2001-or}
Daniel~M Wolpert, Zoubin Ghahramani, and J~Randall Flanagan.
\newblock Perspectives and problems in motor learning.
\newblock \emph{Trends Cogn. Sci.}, 5\penalty0 (11):\penalty0 487--494,
  November 2001.

\bibitem[Woodworth and Thorndike(1901)]{woodworth1901influence}
Robert~S Woodworth and Edward~Lee Thorndike.
\newblock The influence of improvement in one mental function upon the
  efficiency of other functions.(i).
\newblock \emph{Psychological review}, 8\penalty0 (3):\penalty0 247, 1901.

\bibitem[Wołczyk et~al.(2022)Wołczyk, Zajkac, Pascanu, Kuci'nski, and
  Milo's]{Woczyk2022DisentanglingTI}
Maciej Wołczyk, Michal Zajkac, Razvan Pascanu, Lukasz Kuci'nski, and Piotr
  Milo's.
\newblock Disentangling transfer in continual reinforcement learning.
\newblock \emph{ArXiv}, abs/2209.13900, 2022.

\bibitem[Wu et~al.(2021)Wu, Nair, Fei-Fei, and Finn]{wu2021example}
Bohan Wu, Suraj Nair, Li~Fei-Fei, and Chelsea Finn.
\newblock Example-driven model-based reinforcement learning for solving
  long-horizon visuomotor tasks.
\newblock \emph{arXiv preprint arXiv:2109.10312}, 2021.

\bibitem[Wu et~al.(2020)Wu, Pan, Chen, Long, Zhang, and
  Philip]{wu2020comprehensive}
Zonghan Wu, Shirui Pan, Fengwen Chen, Guodong Long, Chengqi Zhang, and S~Yu
  Philip.
\newblock A comprehensive survey on graph neural networks.
\newblock \emph{IEEE transactions on neural networks and learning systems},
  32\penalty0 (1):\penalty0 4--24, 2020.

\bibitem[Wulfmeier et~al.(2015)Wulfmeier, Ondruska, and
  Posner]{wulfmeier2015maximum}
Markus Wulfmeier, Peter Ondruska, and Ingmar Posner.
\newblock Maximum entropy deep inverse reinforcement learning.
\newblock \emph{arXiv preprint arXiv:1507.04888}, 2015.

\bibitem[Wulfmeier et~al.(2017{\natexlab{a}})Wulfmeier, Bewley, and
  Posner]{Wulfmeier2017AddressingAC}
Markus Wulfmeier, Alex Bewley, and Ingmar Posner.
\newblock Addressing appearance change in outdoor robotics with adversarial
  domain adaptation.
\newblock \emph{2017 IEEE/RSJ International Conference on Intelligent Robots
  and Systems (IROS)}, pages 1551--1558, 2017{\natexlab{a}}.

\bibitem[Wulfmeier et~al.(2017{\natexlab{b}})Wulfmeier, Posner, and
  Abbeel]{Wulfmeier2017MutualAT}
Markus Wulfmeier, Ingmar Posner, and P.~Abbeel.
\newblock Mutual alignment transfer learning.
\newblock \emph{ArXiv}, abs/1707.07907, 2017{\natexlab{b}}.

\bibitem[Wulfmeier et~al.(2020{\natexlab{a}})Wulfmeier, Abdolmaleki, Hafner,
  Tobias~Springenberg, Neunert, Siegel, Hertweck, Lampe, Heess, and
  Riedmiller]{wulfmeier2020compositional}
Markus Wulfmeier, Abbas Abdolmaleki, Roland Hafner, Jost Tobias~Springenberg,
  Michael Neunert, Noah Siegel, Tim Hertweck, Thomas Lampe, Nicolas Heess, and
  Martin Riedmiller.
\newblock Compositional transfer in hierarchical reinforcement learning.
\newblock \emph{Robotics: Science and Systems XVI}, Jul 2020{\natexlab{a}}.
\newblock \doi{10.15607/rss.2020.xvi.054}.
\newblock URL \url{http://dx.doi.org/10.15607/rss.2020.xvi.054}.

\bibitem[Wulfmeier et~al.(2020{\natexlab{b}})Wulfmeier, Rao, Hafner, Lampe,
  Abdolmaleki, Hertweck, Neunert, Tirumala, Siegel, Heess, and
  Riedmiller]{wulfmeier2020dataefficient}
Markus Wulfmeier, Dushyant Rao, Roland Hafner, Thomas Lampe, Abbas Abdolmaleki,
  Tim Hertweck, Michael Neunert, Dhruva Tirumala, Noah Siegel, Nicolas Heess,
  and Martin Riedmiller.
\newblock Data-efficient hindsight off-policy option learning,
  2020{\natexlab{b}}.

\bibitem[Wulfmeier et~al.(2021)Wulfmeier, Byravan, Hertweck, Higgins, Gupta,
  Kulkarni, Reynolds, Teplyashin, Hafner, Lampe, and
  Riedmiller]{Wulfmeier2021RepresentationMI}
Markus Wulfmeier, Arunkumar Byravan, Tim Hertweck, Irina Higgins, Ankush Gupta,
  Tejas~D. Kulkarni, Malcolm Reynolds, Denis Teplyashin, Roland Hafner, Thomas
  Lampe, and Martin~A. Riedmiller.
\newblock Representation matters: Improving perception and exploration for
  robotics.
\newblock \emph{2021 IEEE International Conference on Robotics and Automation
  (ICRA)}, pages 6512--6519, 2021.

\bibitem[Xiao et~al.(2022{\natexlab{a}})Xiao, Chan, Sermanet, Wahid, Brohan,
  Hausman, Levine, and Tompson]{xiao2022robotic}
Ted Xiao, Harris Chan, Pierre Sermanet, Ayzaan Wahid, Anthony Brohan, Karol
  Hausman, Sergey Levine, and Jonathan Tompson.
\newblock Robotic skill acquisition via instruction augmentation with
  vision-language models.
\newblock \emph{arXiv preprint arXiv:2211.11736}, 2022{\natexlab{a}}.

\bibitem[Xiao et~al.(2022{\natexlab{b}})Xiao, Radosavovic, Darrell, and
  Malik]{Xiao2022MaskedVP}
Tete Xiao, Ilija Radosavovic, Trevor Darrell, and Jitendra Malik.
\newblock Masked visual pre-training for motor control.
\newblock \emph{ArXiv}, abs/2203.06173, 2022{\natexlab{b}}.

\bibitem[Xie et~al.(2019)Xie, Ebert, Levine, and Finn]{xie2019improvisation}
Annie Xie, Frederik Ebert, Sergey Levine, and Chelsea Finn.
\newblock Improvisation through physical understanding: Using novel objects as
  tools with visual foresight.
\newblock \emph{arXiv preprint arXiv:1904.05538}, 2019.

\bibitem[Xie et~al.(2020)Xie, Harrison, and Finn]{xie2020deep}
Annie Xie, James Harrison, and Chelsea Finn.
\newblock Deep reinforcement learning amidst lifelong non-stationarity.
\newblock \emph{arXiv preprint arXiv:2006.10701}, 2020.

\bibitem[Xie et~al.(2018)Xie, Wang, Rosa, Markham, and
  Trigoni]{Xie2018LearningWT}
Linhai Xie, Sen Wang, Stefano Rosa, A.~Markham, and Agathoniki Trigoni.
\newblock Learning with training wheels: Speeding up training with a simple
  controller for deep reinforcement learning.
\newblock \emph{2018 IEEE International Conference on Robotics and Automation
  (ICRA)}, pages 6276--6283, 2018.

\bibitem[Xing et~al.(2021)Xing, Gupta, Powers, and Dean]{xing2021kitchenshift}
Eliot Xing, Abhinav Gupta, Sam Powers, and Victoria Dean.
\newblock Kitchenshift: Evaluating zero-shot generalization of imitation-based
  policy learning under domain shifts.
\newblock In \emph{NeurIPS 2021 Workshop on Distribution Shifts: Connecting
  Methods and Applications}, 2021.
\newblock URL \url{https://openreview.net/forum?id=DdglKo8hBq0}.

\bibitem[Xu et~al.(2018)Xu, Hasselt, and Silver]{Xu2018MetaGradientRL}
Zhongwen Xu, H.~V. Hasselt, and David Silver.
\newblock Meta-gradient reinforcement learning.
\newblock \emph{ArXiv}, abs/1805.09801, 2018.

\bibitem[Xu et~al.(2020)Xu, van Hasselt, Hessel, Oh, Singh, and
  Silver]{xu2020meta}
Zhongwen Xu, Hado van Hasselt, Matteo Hessel, Junhyuk Oh, Satinder Singh, and
  David Silver.
\newblock Meta-gradient reinforcement learning with an objective discovered
  online.
\newblock \emph{arXiv preprint arXiv:2007.08433}, 2020.

\bibitem[Yang et~al.(2020)Yang, Petersen, Zha, and Faissol]{Yang2020SingleEP}
Jiacheng Yang, Brenden~K. Petersen, H.~Zha, and D.~Faissol.
\newblock Single episode policy transfer in reinforcement learning.
\newblock \emph{ArXiv}, abs/1910.07719, 2020.

\bibitem[Yang et~al.(2023)Yang, Nachum, Du, Wei, Abbeel, and
  Schuurmans]{Yang2023FoundationMF}
Sherry Yang, Ofir Nachum, Yilun Du, Jason Wei, P.~Abbeel, and Dale Schuurmans.
\newblock Foundation models for decision making: Problems, methods, and
  opportunities.
\newblock \emph{ArXiv}, abs/2303.04129, 2023.

\bibitem[Yao et~al.(2018)Yao, Killian, Konidaris, and Doshi-Velez]{yao2018}
Jiayu Yao, Taylor Killian, George Konidaris, and Finale Doshi-Velez.
\newblock Direct policy transfer via hidden parameter markov decision
  processes.
\newblock In \emph{LLARLA Workshop, FAIM}, volume 2018, 2018.

\bibitem[Yarats et~al.(2022)Yarats, Fergus, Lazaric, and
  Pinto]{Yarats2022MasteringVC}
Denis Yarats, Rob Fergus, Alessandro Lazaric, and Lerrel Pinto.
\newblock Mastering visual continuous control: Improved data-augmented
  reinforcement learning.
\newblock \emph{ArXiv}, abs/2107.09645, 2022.

\bibitem[Yen-Chen et~al.(2020)Yen-Chen, Bauza, and Isola]{yen2020experience}
Lin Yen-Chen, Maria Bauza, and Phillip Isola.
\newblock Experience-embedded visual foresight.
\newblock In \emph{Conference on Robot Learning}, pages 1015--1024. PMLR, 2020.

\bibitem[Yokoyama et~al.(2023)Yokoyama, Clegg, Undersander, Ha, Batra, and
  Rai]{yokoyama2023adaptive}
Naoki Yokoyama, Alexander~William Clegg, Eric Undersander, Sehoon Ha, Dhruv
  Batra, and Akshara Rai.
\newblock Adaptive skill coordination for robotic mobile manipulation.
\newblock \emph{arXiv preprint arXiv:2304.00410}, 2023.

\bibitem[Yoo and Seo(2022)]{Yoo2022LearningMT}
Se-Wook Yoo and Seung-Woo Seo.
\newblock Learning multi-task transferable rewards via variational inverse
  reinforcement learning.
\newblock \emph{2022 International Conference on Robotics and Automation
  (ICRA)}, pages 434--440, 2022.

\bibitem[Yu et~al.(2020{\natexlab{a}})Yu, Quillen, He, Julian, Hausman, Finn,
  and Levine]{yu2020meta}
Tianhe Yu, Deirdre Quillen, Zhanpeng He, Ryan Julian, Karol Hausman, Chelsea
  Finn, and Sergey Levine.
\newblock Meta-world: A benchmark and evaluation for multi-task and meta
  reinforcement learning.
\newblock In \emph{Conference on robot learning}, pages 1094--1100. PMLR,
  2020{\natexlab{a}}.

\bibitem[Yu et~al.(2021)Yu, Kumar, Chebotar, Hausman, Levine, and
  Finn]{yu2021conservative}
Tianhe Yu, Aviral Kumar, Yevgen Chebotar, Karol Hausman, Sergey Levine, and
  Chelsea Finn.
\newblock Conservative data sharing for multi-task offline reinforcement
  learning.
\newblock \emph{Advances in Neural Information Processing Systems}, 34, 2021.

\bibitem[Yu et~al.(2023)Yu, Xiao, Stone, Tompson, Brohan, Wang, Singh, Tan,
  Peralta, Ichter, et~al.]{yu2023scaling}
Tianhe Yu, Ted Xiao, Austin Stone, Jonathan Tompson, Anthony Brohan, Su~Wang,
  Jaspiar Singh, Clayton Tan, Jodilyn Peralta, Brian Ichter, et~al.
\newblock Scaling robot learning with semantically imagined experience.
\newblock \emph{arXiv preprint arXiv:2302.11550}, 2023.

\bibitem[Yu et~al.(2017)Yu, Tan, Liu, and Turk]{yu2017preparing}
Wenhao Yu, Jie Tan, C~Karen Liu, and Greg Turk.
\newblock Preparing for the unknown: Learning a universal policy with online
  system identification.
\newblock \emph{arXiv preprint arXiv:1702.02453}, 2017.

\bibitem[Yu et~al.(2019{\natexlab{a}})Yu, Kumar, Turk, and Liu]{yu2019sim}
Wenhao Yu, Visak~CV Kumar, Greg Turk, and C~Karen Liu.
\newblock Sim-to-real transfer for biped locomotion.
\newblock \emph{arXiv preprint arXiv:1903.01390}, 2019{\natexlab{a}}.

\bibitem[Yu et~al.(2019{\natexlab{b}})Yu, Liu, and Turk]{Yu2019PolicyTW}
Wenhao Yu, C.~Karen Liu, and Greg Turk.
\newblock Policy transfer with strategy optimization.
\newblock \emph{ArXiv}, abs/1810.05751, 2019{\natexlab{b}}.

\bibitem[Yu et~al.(2020{\natexlab{b}})Yu, Tan, Bai, Coumans, and
  Ha]{yu2020learning}
Wenhao Yu, Jie Tan, Yunfei Bai, Erwin Coumans, and Sehoon Ha.
\newblock Learning fast adaptation with meta strategy optimization.
\newblock \emph{IEEE Robotics and Automation Letters}, 5\penalty0 (2):\penalty0
  2950--2957, 2020{\natexlab{b}}.

\bibitem[Yuan and Lu(2022)]{yuan2022robust}
Haoqi Yuan and Zongqing Lu.
\newblock Robust task representations for offline meta-reinforcement learning
  via contrastive learning.
\newblock In \emph{International Conference on Machine Learning}, pages
  25747--25759. PMLR, 2022.

\bibitem[Zakka et~al.(2022)Zakka, Zeng, Florence, Tompson, Bohg, and
  Dwibedi]{zakka2022xirl}
Kevin Zakka, Andy Zeng, Pete Florence, Jonathan Tompson, Jeannette Bohg, and
  Debidatta Dwibedi.
\newblock Xirl: Cross-embodiment inverse reinforcement learning.
\newblock In \emph{Conference on Robot Learning}, pages 537--546. PMLR, 2022.

\bibitem[Zeng et~al.(2022)Zeng, Wong, Welker, Choromanski, Tombari, Purohit,
  Ryoo, Sindhwani, Lee, Vanhoucke, et~al.]{zeng2022socratic}
Andy Zeng, Adrian Wong, Stefan Welker, Krzysztof Choromanski, Federico Tombari,
  Aveek Purohit, Michael Ryoo, Vikas Sindhwani, Johnny Lee, Vincent Vanhoucke,
  et~al.
\newblock Socratic models: Composing zero-shot multimodal reasoning with
  language.
\newblock \emph{arXiv preprint arXiv:2204.00598}, 2022.

\bibitem[Zhang et~al.(2019)Zhang, Vikram, Smith, Abbeel, Johnson, and
  Levine]{zhang2019solar}
Marvin Zhang, Sharad Vikram, Laura Smith, Pieter Abbeel, Matthew Johnson, and
  Sergey Levine.
\newblock Solar: Deep structured representations for model-based reinforcement
  learning.
\newblock In \emph{International Conference on Machine Learning}, pages
  7444--7453. PMLR, 2019.

\bibitem[Zhang et~al.(2016)Zhang, Kahn, Levine, and Abbeel]{zhang2016learning}
Tianhao Zhang, Gregory Kahn, Sergey Levine, and Pieter Abbeel.
\newblock Learning deep control policies for autonomous aerial vehicles with
  mpc-guided policy search.
\newblock In \emph{2016 IEEE international conference on robotics and
  automation (ICRA)}, pages 528--535. IEEE, 2016.

\bibitem[Zhang et~al.(2020)Zhang, Li, Zhang, and Zhang]{Zhang2020fGAILLF}
Xin Zhang, Yanhua Li, Ziming Zhang, and Zhi-Li Zhang.
\newblock f-gail: Learning f-divergence for generative adversarial imitation
  learning.
\newblock \emph{ArXiv}, abs/2010.01207, 2020.

\bibitem[Zhao et~al.(2019)Zhao, Sigaud, Stulp, and
  Hospedales]{zhao2019investigating}
Chenyang Zhao, Olivier Sigaud, Freek Stulp, and Timothy~M Hospedales.
\newblock Investigating generalisation in continuous deep reinforcement
  learning.
\newblock \emph{arXiv preprint arXiv:1902.07015}, 2019.

\bibitem[Zhao et~al.(2020)Zhao, Queralta, and Westerlund]{zhao2020sim}
Wenshuai Zhao, Jorge~Pe{\~n}a Queralta, and Tomi Westerlund.
\newblock Sim-to-real transfer in deep reinforcement learning for robotics: a
  survey.
\newblock In \emph{2020 IEEE Symposium Series on Computational Intelligence
  (SSCI)}, pages 737--744. IEEE, 2020.

\bibitem[Zheng et~al.(2022)Zheng, Zhang, and Grover]{zheng2022online}
Qinqing Zheng, Amy Zhang, and Aditya Grover.
\newblock Online decision transformer.
\newblock In \emph{international conference on machine learning}, pages
  27042--27059. PMLR, 2022.

\bibitem[Zhou et~al.(2019{\natexlab{a}})Zhou, Jang, Kappler, Herzog, Khansari,
  Wohlhart, Bai, Kalakrishnan, Levine, and Finn]{zhou2019watch}
Allan Zhou, Eric Jang, Daniel Kappler, Alex Herzog, Mohi Khansari, Paul
  Wohlhart, Yunfei Bai, Mrinal Kalakrishnan, Sergey Levine, and Chelsea Finn.
\newblock Watch, try, learn: Meta-learning from demonstrations and reward.
\newblock \emph{arXiv preprint arXiv:1906.03352}, 2019{\natexlab{a}}.

\bibitem[Zhou et~al.(2022{\natexlab{a}})Zhou, Kumar, Finn, and
  Rajeswaran]{Zhou2022mj}
Allan Zhou, Vikash Kumar, Chelsea Finn, and Aravind Rajeswaran.
\newblock Policy architectures for compositional generalization in control.
\newblock \emph{arXiv preprint arXiv:2203.05960}, 2022{\natexlab{a}}.

\bibitem[Zhou et~al.(2020)Zhou, Li, Yang, Wang, and Hospedales]{zhou2020online}
Wei Zhou, Yiying Li, Yongxin Yang, Huaimin Wang, and Timothy Hospedales.
\newblock Online meta-critic learning for off-policy actor-critic methods.
\newblock \emph{Advances in neural information processing systems},
  33:\penalty0 17662--17673, 2020.

\bibitem[Zhou et~al.(2019{\natexlab{b}})Zhou, Pinto, and
  Gupta]{zhou2019environment}
Wenxuan Zhou, Lerrel Pinto, and Abhinav Gupta.
\newblock Environment probing interaction policies.
\newblock \emph{arXiv preprint arXiv:1907.11740}, 2019{\natexlab{b}}.

\bibitem[Zhou et~al.(2022{\natexlab{b}})Zhou, Bohez, Humplik, Heess,
  Abdolmaleki, Rao, Wulfmeier, and Haarnoja]{zhou2022forgetting}
Wenxuan Zhou, Steven Bohez, Jan Humplik, Nicolas Heess, Abbas Abdolmaleki,
  Dushyant Rao, Markus Wulfmeier, and Tuomas Haarnoja.
\newblock Forgetting and imbalance in robot lifelong learning with off-policy
  data.
\newblock In \emph{Conference on Lifelong Learning Agents}, pages 294--309.
  PMLR, 2022{\natexlab{b}}.

\bibitem[Zhu et~al.(2020{\natexlab{a}})Zhu, Yu, Gupta, Shah, Hartikainen,
  Singh, Kumar, and Levine]{zhu2020ingredients}
Henry Zhu, Justin Yu, Abhishek Gupta, Dhruv Shah, Kristian Hartikainen, Avi
  Singh, Vikash Kumar, and Sergey Levine.
\newblock The ingredients of real-world robotic reinforcement learning.
\newblock \emph{arXiv preprint arXiv:2004.12570}, 2020{\natexlab{a}}.

\bibitem[Zhu et~al.(2018)Zhu, Wang, Merel, Rusu, Erez, Cabi, Tunyasuvunakool,
  Kram{\'a}r, Hadsell, de~Freitas, et~al.]{zhu2018reinforcement}
Yuke Zhu, Ziyu Wang, Josh Merel, Andrei Rusu, Tom Erez, Serkan Cabi, Saran
  Tunyasuvunakool, J{\'a}nos Kram{\'a}r, Raia Hadsell, Nando de~Freitas, et~al.
\newblock Reinforcement and imitation learning for diverse visuomotor skills.
\newblock \emph{arXiv preprint arXiv:1802.09564}, 2018.

\bibitem[Zhu et~al.(2020{\natexlab{b}})Zhu, Wong, Mandlekar,
  Mart\'{i}n-Mart\'{i}n, Joshi, Nasiriany, and Zhu]{robosuite2020}
Yuke Zhu, Josiah Wong, Ajay Mandlekar, Roberto Mart\'{i}n-Mart\'{i}n, Abhishek
  Joshi, Soroush Nasiriany, and Yifeng Zhu.
\newblock robosuite: A modular simulation framework and benchmark for robot
  learning.
\newblock In \emph{arXiv preprint arXiv:2009.12293}, 2020{\natexlab{b}}.

\bibitem[Zhu et~al.(2020{\natexlab{c}})Zhu, Lin, and Zhou]{zhu2020transfer}
Zhuangdi Zhu, Kaixiang Lin, and Jiayu Zhou.
\newblock Transfer learning in deep reinforcement learning: A survey.
\newblock \emph{arXiv preprint arXiv:2009.07888}, 2020{\natexlab{c}}.

\bibitem[Ziebart et~al.(2008)Ziebart, Maas, Bagnell, Dey,
  et~al.]{ziebart2008maximum}
Brian~D Ziebart, Andrew~L Maas, J~Andrew Bagnell, Anind~K Dey, et~al.
\newblock Maximum entropy inverse reinforcement learning.
\newblock In \emph{Aaai}, volume~8, pages 1433--1438. Chicago, IL, USA, 2008.

\bibitem[Zintgraf et~al.(2020)Zintgraf, Shiarlis, Igl, Schulze, Gal, Hofmann,
  and Whiteson]{Zintgraf2020VariBADAV}
Luisa~M. Zintgraf, Kyriacos Shiarlis, Maximilian Igl, Sebastian Schulze, Yarin
  Gal, Katja Hofmann, and Shimon Whiteson.
\newblock Varibad: A very good method for bayes-adaptive deep rl via
  meta-learning.
\newblock \emph{ArXiv}, abs/1910.08348, 2020.

\bibitem[Zintgraf et~al.(2021)Zintgraf, Feng, Lu, Igl, Hartikainen, Hofmann,
  and Whiteson]{Zintgraf2021ExplorationIA}
Luisa~M. Zintgraf, Leo Feng, Cong Lu, Maximilian Igl, Kristian Hartikainen,
  Katja Hofmann, and Shimon Whiteson.
\newblock Exploration in approximate hyper-state space for meta reinforcement
  learning.
\newblock In \emph{ICML}, 2021.

\end{thebibliography}
